\documentclass[10pt,twocolumn,letterpaper]{article}

\usepackage{iccv}              

\usepackage{tikz}  
\usepackage{bigfoot}  

\newcommand{\redunderline}[1]{\textcolor{red}{\underline{#1}}}
\newcommand*{\imagenet}{\textsc{ImageNet1K}\xspace}
\newcommand*{\imnet}{\textsc{IN1K}\xspace}

\definecolor{iccvblue}{rgb}{0.21,0.49,0.74}
\usepackage[pagebackref,breaklinks,colorlinks,allcolors=iccvblue]{hyperref}
\usepackage{adjustbox, float, enumitem, verbatim, colortbl}
\usepackage{enumerate}
\usepackage{titletoc}
\usepackage{multirow}

\usepackage{soul}
\usepackage{graphicx}

\makeatletter
\newcommand\level[1]{%
  \ifcase#1\relax\expandafter\chapter\or
    \expandafter\section\or
    \expandafter\subsection\or
    \expandafter\subsubsection\else
    \def\next{\@level{#1}}\expandafter\next
  \fi}
\newcommand{\@level}[1]{%
  \@startsection{level#1}
    {#1}
    {\z@}%
    {-3.25ex\@plus -1ex \@minus -.2ex}%
    {1.5ex \@plus .2ex}%
    {\normalfont\normalsize\bfseries}}

\newcounter{level4}[subsubsection]
\@namedef{thelevel4}{\thesubsubsection.\arabic{level4}}
\@namedef{level4mark}#1{}
\count@=4
\loop\ifnum\count@<100
  \begingroup\edef\x{\endgroup
    \noexpand\newcounter{level\number\numexpr\count@+1\relax}[level\number\count@]
    \noexpand\@namedef{thelevel\number\numexpr\count@+1\relax}{%
      \noexpand\@nameuse{thelevel\number\count@}.\noexpand\arabic{level\number\numexpr\count@+1\relax}}
    \noexpand\@namedef{level\number\numexpr\count@+1\relax mark}####1{}}
  \x
  \advance\count@\@ne
\repeat
\makeatother
\def\paperID{14670} 
\def\confName{ICCV}
\def\confYear{2025}

\title{Evaluating Self-Supervised Learning in Medical Imaging:  A Systematic Investigation of Robustness, Generalizability, and Multi-Domain Transfer}

\author{%
Valay Bundele$^{*}$ \quad
Karahan Sarıtaş$^{*}$ \quad 
Bora Kargi$^{*}$ \quad
Oğuz Ata Çal$^{*}$ \quad \\
Kıvanç Tezören$^{*}$ \quad
Zohreh Ghaderi \quad
Hendrik Lensch \\
University of Tübingen \\
{\small $^*$Equal contribution.}
}

\begin{document}
\maketitle
\begin{abstract}
Self-supervised learning (SSL) has emerged as a promising paradigm in medical imaging, addressing the chronic challenge of limited labeled data in healthcare settings. While SSL has shown impressive results, existing studies in the medical domain are often limited in scope, focusing on specific datasets or modalities, or evaluating only isolated aspects of model performance. This fragmented evaluation approach poses a significant challenge, as models deployed in critical medical settings must not only achieve high accuracy but also demonstrate robust performance and generalizability across diverse datasets and varying conditions. To bridge this gap, we conduct a rigorous investigation into the design space of SSL for medical imaging, evaluating 8 major SSL methods across 11 real-world medical datasets. Our analysis spans three core dimensions: (1) in-domain performance under varying label proportions (1\%, 10\%, and 100\%), (2) cross-dataset generalization, and (3) robustness to out-of-distribution (OOD) samples. Beyond empirical evaluation, we further examine how initialization strategies, model architectures, and multi-domain pre-training contribute to SSL’s success in medical imaging.
Our findings establish a principled understanding of how design choices shape SSL performance in medical imaging, providing a foundation for building more robust, generalizable, and clinically viable self-supervised models. 
\end{abstract}

\section{Introduction}
\label{sec:intro}

Medical image annotation is a resource-intensive task that requires specialized domain knowledge, making it significantly more costly and laborious than annotating natural images \cite{10.1016/j.eswa.2023.121282, kunzmann2022unobtrusivequalitysupervisionapproach, schulz2021labelscarcitybiomedicinedatarich, humbert2022strategies}. The scarcity of labeled medical data, coupled with the complexity of the annotation process, presents a significant challenge for building effective machine learning models in healthcare. Self-supervised learning has emerged as a powerful solution to these limitations, enabling models to learn rich representations from unlabeled data prior to task-specific fine-tuning with minimal labeled samples \cite{Ericsson_2022, doerrich2024rethinkingmodelprototypingmedmnist, kang2023benchmarkingselfsupervisedlearningdiverse, azizi2021bigselfsupervisedmodelsadvance, afzal2024a, xiao2022delvingmaskedautoencodersmultilabel, huang2024systematiccomparisonsemisupervisedselfsupervised}.

While self-supervised learning (SSL) has proven effective in improving classification performance with limited labels \cite{chen2020simclr, grill2020byol, zheng2021resslrelationalselfsupervisedlearning, caron2021dino, chen2021empiricalstudytrainingselfsupervised}, there has been limited work in evaluating the \textit{robustness} and \textit{generalizability} of learned models. For safe deployment in healthcare, models must perform reliably across diverse settings and recognize when to withhold predictions on out-of-distribution (OOD) samples. Although SSL methods have been evaluated for OOD detection in natural image contexts \cite{li2023rethinkingoutofdistributionooddetection, khalid2022roddselfsupervisedapproachrobust, hendrycks2019usingselfsupervisedlearningimprove, Mohseni2020SelfSupervisedLF, guilleescuret2023cadetfullyselfsupervisedoutofdistribution}, their effectiveness in medical imaging remains largely under-researched. Existing studies are limited in scope: either focusing on a single imaging modality using only supervised approaches \cite{berger2021confidencebasedoutofdistributiondetectioncomparative}, examining a few SSL methods in one medical domain \cite{9607789}, or investigating only task-specific SSL techniques rather than general-purpose methods \cite{cai2024medianomalycomparativestudyanomaly}.

Beyond OOD detection, an effective model should generalize across different tasks and modalities—
a crucial yet underexplored aspect of SSL in medical imaging,
where datasets and conditions vary widely. A model that performs well across different datasets or modalities ensures continued diagnostic support even when some imaging modalities are inaccessible, reduces the need for extensive retraining, and ensures robust performance in diverse clinical settings.
\citet{azizi2022re} propose an SSL-based representation learning method and assess its robustness across various medical imaging tasks and domains. However, their study focuses on a single approach, leaving it unclear whether their findings hold across different SSL paradigms.


To address these gaps,  
we conduct a systematic investigation into the SSL design space in medical imaging, evaluating robustness and generalizability across diverse settings.
We assess various SSL methods for OOD detection, comparing the effectiveness of convolutional networks (ResNet-50) with transformer-based architectures (ViT-Small). We further assess the generalizability of learned representations by training linear classifiers on frozen encoders across different datasets, enabling cross-dataset transfer evaluation.  

A widely adopted strategy in practice is to initialize model training with weights pre-trained on large-scale natural image datasets, such as \imagenet \cite{imagenet}, and subsequently fine-tune the model on a medical dataset \cite{matsoukas2022makestransferlearningwork, Morid_2021, Tajbakhsh_2016} — a paradigm known as transfer learning. Supervised \imagenet weights can be used either to initiate subsequent self-supervised training \cite{matsoukas2022makestransferlearningwork, matsoukas2021timereplacecnnstransformers}, or can be directly used without further training \cite{doerrich2024rethinkingmodelprototypingmedmnist}. We adopt the former approach as it better adapts the model to the fine-grained medical domain. Building upon previous research, we investigate how \imagenet initialization affects model generalizability across diverse datasets while also analyzing its impact on OOD detection performance.

Recent studies also suggest that training across multiple domains can further improve a model’s robustness and generalizability, especially in data-limited, OOD-prone scenarios, such as those often encountered in healthcare \cite{zadorozhny2022outofdistributiondetectionmedicalapplications, ZAID2022103882, ozkan2024multidomainimprovesoutofdistributiondatalimited}. 
Multi-domain models leverage diverse data sources, allowing them to utilize complementary information across domains and enhance overall performance \cite{CHEN2023107200, woerner2024comprehensiveeasytousemultidomainmultitask}. 
However, existing studies do not fully evaluate how common SSL methods perform in the context of multi-domain pre-training. To address this, we conduct experiments to examine influence of multi-domain data on the performance of SSL methods, providing a more comprehensive understanding of their generalizability and robustness.

To summarize, we explore the following key questions:

\textbf{Q1.} How is the in-domain classification performance, generalizability, and robustness to OODs affected by: (1) the choice of SSL method, (2) the initialization strategy, and (3) the model architecture?

\textbf{Q2.} Which SSL method and initialization strategy yield the best performance when labeled data is limited?

\textbf{Q3.} Does multi-domain training improve the robustness and generalizability of SSL encoders, enhancing both in-domain and OOD performance?

By addressing these questions, we provide a deeper understanding of the potential of SSL in medical imaging, establishing a principled framework for 
developing robust and generalizable models in real-world healthcare settings.
 \\

\section{Related Work}
\label{sec:formatting}

\paragraph*{Prior self-supervised evaluations}

Several works have assessed self-supervised strategies on natural image datasets \cite{dacosta2022sololearnlibraryselfsupervisedmethods, goyal2019scalingbenchmarkingselfsupervisedvisual, marks2024closerlookbenchmarkingselfsupervised}. However, a gap remains in the literature when it comes to a systematic evaluation of SSL methods in the medical domain using standardized datasets. Recently, \citet{doerrich2024rethinkingmodelprototypingmedmnist} proposed a systematic evaluation using supervised and self-supervised approaches. However, they mainly experiment with \imagenet pre-trained encoders without any fine-tuning on the medical data, which may limit model’s ability to capture domain-specific features. \citet{kang2023benchmarkingselfsupervisedlearningdiverse} demonstrated that self-supervised
pre-training outperforms supervised \imagenet baselines for pathology but did not explore other modalities.
\citet{huang2024systematiccomparisonsemisupervisedselfsupervised} analyzed SSL and semi-supervised methods but limited their study to four medical datasets, without assessing robustness or generalizability across diverse tasks. In contrast, our work offers a comprehensive evaluation across several real-world medical datasets, focusing on multi-domain performance, encoder robustness, generalizability, and adaptability to limited labeled data.

\vspace{-12px}
\paragraph*{Out-of-Distribution Detection}

Robust pre-trained encoders can be employed as OOD detectors to prevent dangerous misclassifications in the medical domain. To this end, several studies have investigated the use of visual recognition systems as OOD detectors \cite{li2023rethinkingoutofdistributionooddetection, berger2021confidencebasedoutofdistributiondetectioncomparative, galil2023frameworkbenchmarkingclassoutofdistributiondetection, Narayanaswamy_2023_ICCV, zhong2022selfsupervisedlearningrobustsupervised}. \citet{hendrycks2019usingselfsupervisedlearningimprove} show that SSL methods outperform fully supervised ones on natural image datasets for OOD detection. \citet{Narayanaswamy_2023_ICCV} address modality shift and novel class detection in the medical domain, but their focus is limited to supervised training of OOD detectors. The SSD framework by \citet{2103.12051} demonstrates that SSL can significantly improve OOD detection, achieving performance comparable to supervised methods.  Additionally, both \citet{li2023rethinkingoutofdistributionooddetection} and \citet{Mohseni2020SelfSupervisedLF} introduce self-supervised approaches that improve OOD detection, though their analyses are limited to natural images.

Despite these advancements, there remains a notable gap: no prior work has systematically examined SSL methods for OOD detection in the medical domain, across diverse architectures and pre-training strategies. Our study aims to fill this gap by conducting a rigorous analysis of OOD detection across diverse medical imaging conditions.

\vspace{-12px}
\paragraph*{Generalizability}

The ability of an encoder to perform effectively on datasets and domains beyond its training distribution is crucial for assessing representation quality. 
\citet{li2020domain} enhance generalizability through variational encoding with a linear-dependency regularization, while \citet{yan2024prompt} propose a domain-generalization framework for medical image classification without domain labels in the supervised setting. Although \citet{fedorov2022tastingcakeevaluatingselfsupervised} investigate the generalization of SSL methods in medical imaging, their analysis is confined to MRI data. 
\citet{azizi2022re} propose an SSL-based representation learning method, evaluating the robustness of their approach across multiple medical imaging tasks and modalities, yet they do not compare different SSL methods. In contrast, we conduct systematic cross-dataset experiments to evaluate the true generalization capabilities of several SSL approaches.



\vspace{-12px}
\paragraph*{Transfer Learning}
In medical domain, network initialization with pre-trained weights from large-scale datasets like \imagenet is a common and effective practice
\cite{azizi2022re, matsoukas2021timereplacecnnstransformers, matsoukas2021timereplacecnnstransformers, taher2021systematicbenchmarkinganalysistransfer, Tajbakhsh_2016}.
\citet{matsoukas2021timereplacecnnstransformers} demonstrated that SSL in medical domain, initialized with supervised \imagenet weights, is effective for both ViTs and CNNs in medical imaging.
To the best of our knowledge, the most comprehensive study on transfer learning in medical domain is conducted by \citet{taher2021systematicbenchmarkinganalysistransfer}. Although they emphasize domain-adapted continual pre-training, they implement it exclusively with supervised models, leaving the impact of continual pre-training on SSL largely unexplored. In our work, we investigate continual self-supervised pre-training, and the potential of self-supervised \imagenet initialization to yield better representations for in-domain performance and OOD detection, an approach not previously explored in medical context.

\section{Methodology}

\paragraph*{Representation Learning Methods}
We consider the following eight discriminative SSL methods: SimCLR \cite{chen2020simclr}, DINO \cite{caron2021dino}, BYOL \cite{grill2020bootstraplatentnewapproach}, ReSSL \cite{zheng2021resslrelationalselfsupervisedlearning}, MoCo v3 \cite{chen2021empiricalstudytrainingselfsupervised}, NNCLR \cite{dwibedi2021littlehelpfriendsnearestneighbor}, VICREG \cite{bardes2022vicregvarianceinvariancecovarianceregularizationselfsupervised}, and Barlow Twins \cite{zbontar2021barlowtwinsselfsupervisedlearning} which are explained briefly in Appendix~\ref{sec:appendix_method}. A comprehensive survey by \citet{huang2023self} highlights SimCLR, MoCo, and BYOL as the most frequently adopted SSL frameworks in medical imaging research. We include the other methods to make our study more comprehensive. 

\vspace{-12px}
\paragraph*{Tasks and Datasets}
\label{sec:methodology-tasks-and-datasets}

We used real-world medical datasets, including NCT-CRC-HE-100K, CRC-VAL-HE-7K \cite{kather2019predicting, kather2018}, HAM10000 \cite{tschandl2018ham10000}, OCT dataset \cite{kermany2018identifying}, pediatric chest X-Ray images \cite{kermany2018identifying}, DeepDRiD Retina dataset \cite{liu2022deepdrid}, breast ultrasound images \cite{al-dhabyani2020dataset}, microscopic peripheral blood cell images \cite{acevedo2020dataset}, BBBC051 \cite{ljosa2012annotated} and LiTS \cite{bilic2023lits}, all curated and standardized under the MedMNIST framework \cite{Yang_2023}. For the LiTS dataset, we followed the MedMNIST adaptation, which transforms the original segmentation task into an organ classification problem by extracting regions of interest.

To maintain our focus on multiclass medical classification and ordinal regression, we exclude ChestMNIST from our analysis, as it only offers multi-label disease classification. Our experiments span 11 MedMNIST datasets, additional information being provided in Appendix~\ref{sec:appendix_dset}.


\vspace{-12px}
\paragraph*{Architectures} For our study, we use ResNet-50 \cite{he2015resnet} with approximately 25 million parameters and ViT-Small \cite{dosovitskiy2021imageworth16x16words} with approximately 22 million parameters to ensure a fair comparison. We exclude other models like ViT-Tiny and ViT-Base due to their significantly different parameter counts of 5.7 million and 86 million, respectively.

\vspace{-12px}
\paragraph*{Pre-training}

We employed the \texttt{solo-learn}
library \cite{dacosta2022sololearnlibraryselfsupervisedmethods} with some modifications for pre-training, OOD detection and linear evaluation. Our implementation details are provided in Appendix \ref{sec:appendix_tr_pt}, and the code will be released after acceptance. 
In total, we focus on five different pre-training schemes: supervised training with (1) random initialization and (2) supervised \imagenet initialization; self-supervised pre-training with (3) random initialization, (4) supervised \imagenet initialization, and (5) self-supervised \imagenet initialization. 

\vspace{-12px}
\paragraph*{Linear Evaluation}
\label{par:lin-eval}
To evaluate the quality of self-supervised pre-trained encoders, we use linear probing, training linear classifiers on frozen features to assess downstream performance 
\cite{kolesnikov2019revisitingselfsupervisedvisualrepresentation, grill2020bootstraplatentnewapproach, he2020momentumcontrastunsupervisedvisual}. 
Additionally, to simulate realistic scenarios in the medical domain, we evaluate the low-shot performance by training a multi-class logistic regression on the frozen features with only 1\% and 10\% labeled data, following the evaluation protocol established by \citet{caron2021dino}. We report accuracy and Area Under the Curve (AUC) scores. Further details on the linear evaluation setup and evaluation metrics can be found in \ref{sec:appendix_tr_le} and \ref{sec:appendix_tr_eval}, respectively.

\vspace{-12px}
\paragraph*{Generalizability}
To assess generalizability, we conduct cross-dataset experiments using our dataset collection,
$\mathcal{D} = \{D_1, ..., D_{11}\}$. 
For each SSL method, we pre-train the model on a dataset $D_i$ and evaluate its transferability by training a linear classifier on frozen features on each of the remaining datasets, $\mathcal{D} \setminus {D_i}$. This process is repeated for each dataset $D_i$ in $\mathcal{D}$ to cover all cross-dataset pairs. 


\vspace{-12px}
\paragraph*{OOD Detection}
We evaluate various SSL methods for distinguishing between ID and OOD samples, focusing on backbone architectures, pre-training strategies, and multi-domain learning. We follow the approach defined in \cite{lee2018simpleunifiedframeworkdetecting}, where feature representations \( \mathbf{f}(x) \in \mathbb{R}^d \) are assumed to follow a class-conditional multivariate Gaussian, with parameters estimated from the ID training data \( \mathcal{P}_{\text{ID}} \) (i.e., samples used to train the encoder). For each class \( c \), the Mahalanobis distance is computed as
\[
D_\text{M}(x, c) = \sqrt{(\mathbf{f}(x) - \boldsymbol{\mu}_c)^\top \Sigma^{-1} (\mathbf{f}(x) - \boldsymbol{\mu}_c)}
\]
and the confidence score is given by
\[
S(x) = -\min_c D_\text{M}(x, c).
\]
Samples are considered OOD if they originate from a different dataset or subset (denoted as \( \mathcal{P}_{\text{OOD}} \)) than the one used for training. OOD detection performance is evaluated using threshold-independent metrics such as AUROC and AUPR. Further details are provided in Appendix.





\vspace{-12px}
\paragraph*{Multi-Domain Learning}
We examine the impact of multi-domain learning by comparing models trained on single-domain MedMNIST datasets with those trained on two dataset combinations. The first combination, named Organ\{A,C,S\} merges the Organ\{A,C,S\}MNIST datasets to represent a single-modality scenario. The second combination, referred to as Organ\{A,S\}PnePath, includes Organ\{A,S\}MNIST (CT), PathMNIST (Colon Pathology), and PneumoniaMNIST (Chest X-Ray), allowing us to assess the effect of combining different modalities.
We then train SSL methods from scratch
on these combined datasets to assess how dataset combination affects performance. 


\section{Experiments \& Results}
In this section, we analyze the in-domain (ID) performance, robustness to OOD samples, and generalizability of the SSL methods. For clarity in certain figures where direct comparisons between the SSL methods are not essential, we display results for a representative subset of five methods, with complete results available in the Appendix for reference.

\subsection{In-Domain Performance}
We assess each SSL method's in-distribution performance by pre-training ResNet-50 on the train set and evaluating with a linear classifier on the test set using frozen features. MoCo v3 exhibits strong performance, achieving the highest accuracy across 5 of the 11 datasets. BYOL and SimCLR also demonstrate competitive results, trailing closely behind MoCo v3. Notably, self-supervised learning outperforms supervised learning in 7 out of 11 datasets when both approaches start from random initialization. Numerical results can be found in Appendix~\ref{sec:appendix_ap_idp_lat}.

\begin{figure}[h]
\hspace*{-0.7cm} 
    \centering
    \includegraphics[width=0.8\columnwidth]{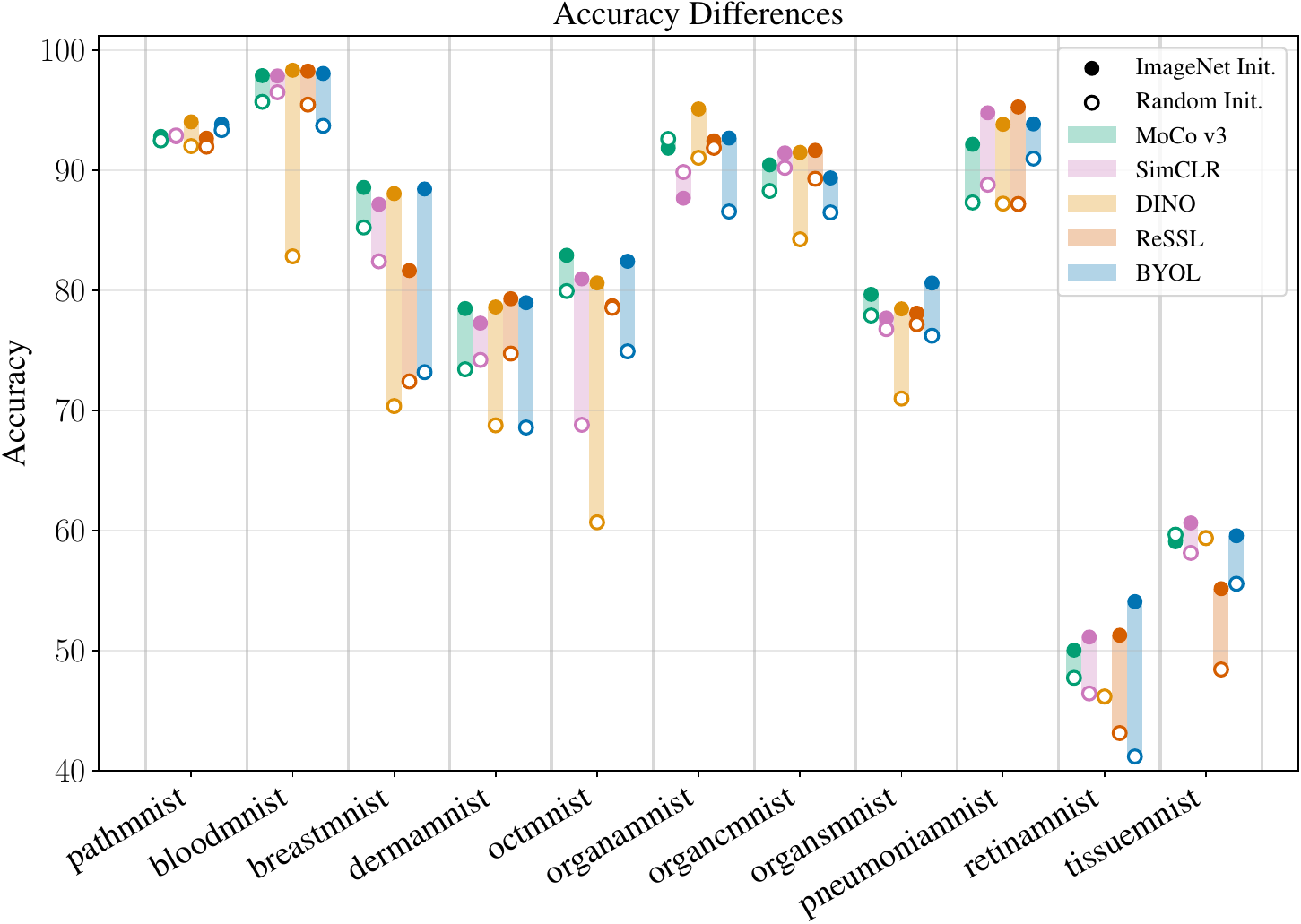}
    \caption{Accuracy differences between self-supervised pre-training with random (unfilled markers) vs. \imagenet (filled markers) initialization using ResNet-50 backbone.}
    \label{fig:id_pretrain.pdf}
\end{figure}

Next, we analyze the effect of supervised \imagenet initialization on self-supervised training performance for in-domain tasks. As shown in Figure \ref{fig:id_pretrain.pdf}, \imagenet initialization consistently improves performance on in-domain classification tasks, a trend also observed across other methods in Appendix~\ref{sec:appendix_ap_idp_lat}. Interestingly, DINO and BYOL show the most significant accuracy gains when transitioning from random to \imagenet initialization.

\begin{figure}[h]
\hspace*{-0.8cm} 
    \centering
    \includegraphics[width=0.8\columnwidth]{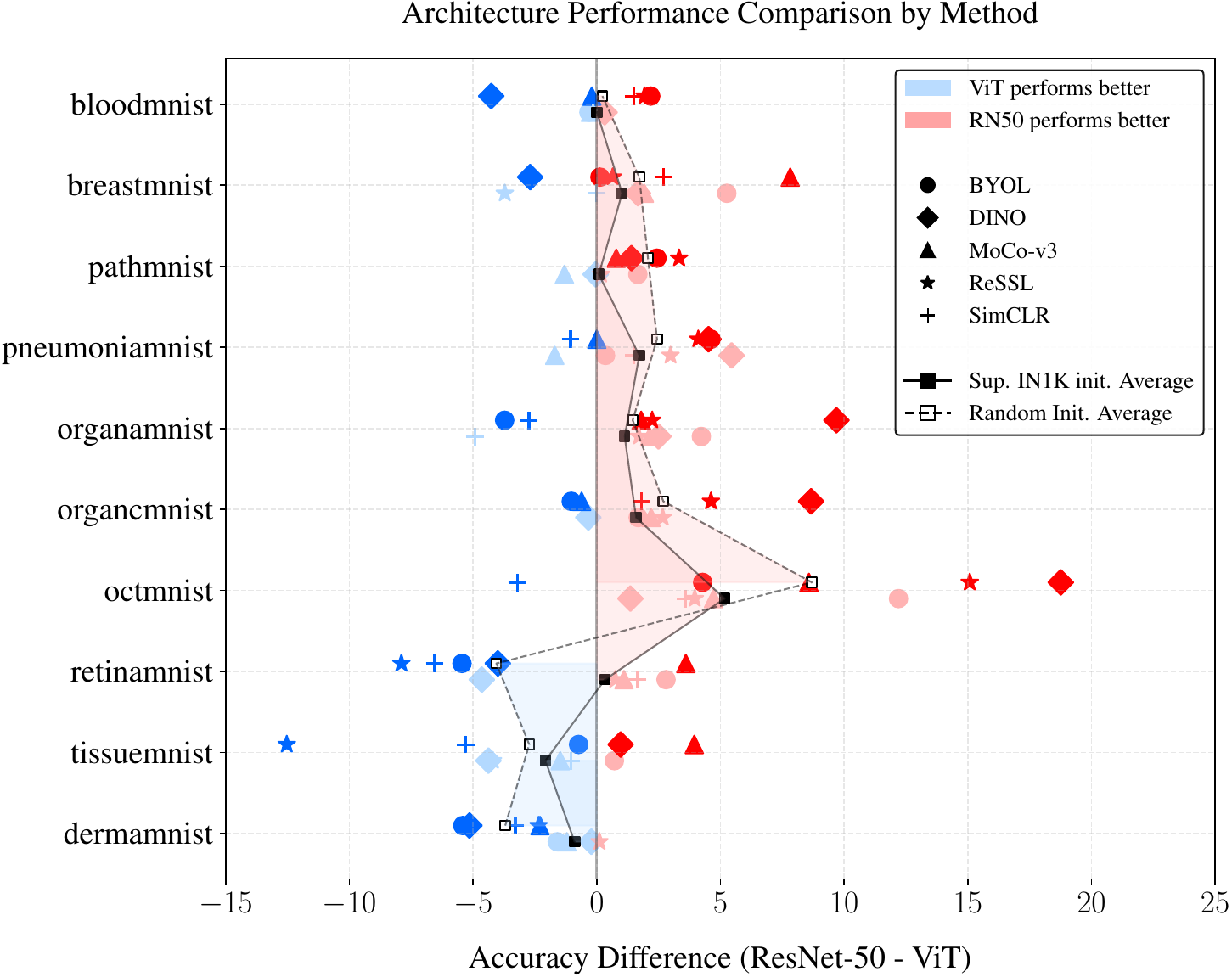}
    \caption{Performance differential between ResNet-50 and ViT across SSL methods, comparing initialization strategies. Dark colors stand for random initialization, whereas light colors stand for \imagenet initialization. For each dataset, geometric markers show individual method performance differentials. The connecting lines represent the mean performance differential across all methods for different initialization strategies.}
    \label{fig:id_arch.pdf}
\end{figure}

Figure \ref{fig:id_arch.pdf}
 further emphasizes the effect of \imagenet initialization by showing the accuracy differential between ResNet-50 and ViT ($\text{ACC}_\text{RN50} - \text{ACC}_\text{ViT}$) across various methods and datasets, presented for both random and \imagenet-initialized cases. In both scenarios, ResNet-50 generally outperforms ViT. However, the accuracy gap between these architectures generally narrows when transitioning from random initialization to \imagenet-supervised weights. This reduction can be attributed to two factors: firstly, as overall accuracy improves, incremental gains in performance diminish, naturally reducing the architecture gap. Secondly, as prior research suggests \cite{dosovitskiy2021imageworth16x16words, arnab2021vivitvideovisiontransformer, akbari2021vatttransformersmultimodalselfsupervised}, the data-intensive nature of transformer-based architectures can be harnessed effectively through large-scale pre-training, such as with \imagenet initialization. Complete numerical results for all methods are provided in Appendix~\ref{sec:appendix_ap_idp_lat}.

\label{sec:experiments-label-availability}
 Expanding on our in-domain performance evaluation, we also examine how the architectures perform under different levels of label availability. Table \ref{table:accuracy_drop_resnet50_vit_small} presents accuracy drops for various SSL methods when reducing label availability from 100\% to 1\% in random initialization setting. ResNet-50 not only outperforms other methods in the fully labeled (100\%) scenario, as shown in Figure \ref{fig:id_arch.pdf}, but also exhibits consistently lower average performance drops across nearly all methods. This indicates that ResNet-50 maintains more robust performance compared to ViT-Small even under severe label scarcity.
A similar pattern can also be observed across other methods and in \imagenet initialization setting in Appendix~\ref{sec:appendix_ap_idp_lat}, highlighting that transformers generally require more labeled data to reach optimal performance \cite{zhu2023understandingvittrainsbadly, xiao2022delvingmaskedautoencodersmultilabel}.

\begin{figure}[h]
    \centering
    \includegraphics[width=0.8\columnwidth]{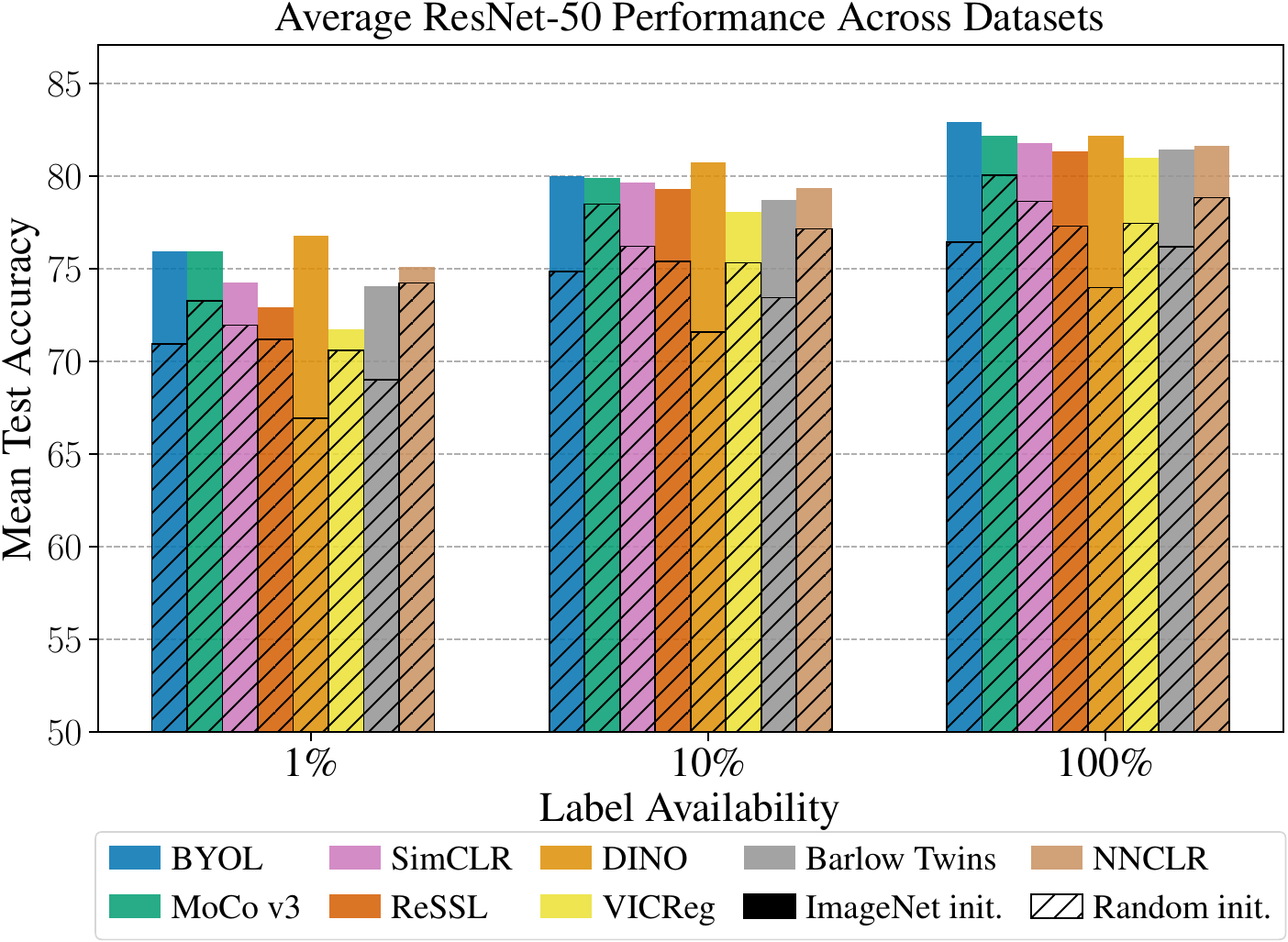}
    \caption{Mean test accuracy across various datasets with 1\%, 10\%, and 100\% label availability, comparing \imagenet against random initialization (hatched) using ResNet-50.}
    \label{fig:scarcity_test_acc_grouped_by_percentage (Resnet50)}
\end{figure}

Moreover, Figure \ref{fig:scarcity_test_acc_grouped_by_percentage (Resnet50)} presents a bar chart comparing the mean test accuracies of ResNet-50 across different label availability settings (1\%, 10\%, and 100\%), highlighting the effects of \imagenet versus random initialization. The results show that models initialized with \imagenet weights consistently outperform their randomly initialized counterparts, especially under label-scarce conditions. This finding underscores the value of pre-trained weights in boosting model performance when labeled data is limited. Notably, with \imagenet initialization, DINO excels in label-scarce scenarios, though it is slightly outperformed by BYOL at the 100\% label level. On the other hand, MoCo v3 shows promising results across most of the label percentages in the random initialization setting, slightly losing only to NNCLR at the 1\% label level. A similar trend captured in ViT models is shown in Appendix~\ref{sec:appendix_ap_idp_lat}.

\begin{table}[h]
  \centering
    \caption{
    Accuracy drops from 100\% to 1\% label availability across datasets for each method and architecture with random initialization. Larger drops are highlighted in \redunderline{red}, emphasizing higher sensitivity to limited labeled data.}
    \scalebox{0.55}{
    \begin{tabular}{lcc|cc|cc|cc|cc}
        \toprule
        \multirow{2}{*}{} & \multicolumn{2}{c}{SimCLR} & \multicolumn{2}{c}{DINO} & \multicolumn{2}{c}{BYOL} & \multicolumn{2}{c}{ReSSL} & \multicolumn{2}{c}{MoCo v3}  \\ 
        \cmidrule(lr){2-3} \cmidrule(lr){4-5} \cmidrule(lr){6-7} \cmidrule(lr){8-9} \cmidrule(lr){10-11}
         & RN50 & ViT & RN50 & ViT & RN50 & ViT & RN50 & ViT & RN50 & ViT \\
        \midrule
        Path & \redunderline{2.91} & 0.10 & 0.24 & \redunderline{1.22} & 1.10 & \redunderline{1.60} & \redunderline{2.38} & 0.22 & -0.12 & \redunderline{1.50} \\
        Derma & 4.54 & \redunderline{7.93} & 1.90 & \redunderline{6.56} & 1.68 & \redunderline{4.51} & 5.54 & \redunderline{8.41} & 3.00 & \redunderline{5.47} \\
         OCT & \redunderline{9.24} & 5.62 & 0.56 & \redunderline{6.08} & -2.24 & \redunderline{7.52} & -2.16 & \redunderline{10.76} & 4.20 & \redunderline{8.18} \\
      Pneumonia & -0.13 & \redunderline{9.84} &  -2.85 & \redunderline{0.22} & \redunderline{0.06} &  -2.21 & -3.07 & \redunderline{3.72} & \redunderline{0.45} & -0.67 \\
         Retina & 4.50 & \redunderline{9.65} & 3.30 & \redunderline{5.85} & -0.85 & \redunderline{1.90} & 0.40 & \redunderline{7.00} & \redunderline{4.70} & 0.60 \\
        Breast & \redunderline{7.82} & 3.72 & \redunderline{0.51} & 0.00 & \redunderline{0.13} & 0.00 & 0.77 & \redunderline{1.15} & \redunderline{6.54} & 0.90 \\
        Blood & 3.34 & \redunderline{13.11} & 12.34 & \redunderline{14.57} & 8.45 & \redunderline{17.43} & 9.92 & \redunderline{13.91} & 6.28 & \redunderline{12.04} \\

        Tissue & \redunderline{6.66} & 6.37 & \redunderline{7.45} & 5.04 & 7.10 & \redunderline{8.69} & 5.76 & \redunderline{6.58} & 7.18 & \redunderline{10.20} \\
        OrganA & \redunderline{9.40} & 9.10 & 8.72 & \redunderline{11.73} & \redunderline{9.42} & 6.99 & 7.36 & \redunderline{8.00} & \redunderline{6.17} & 4.63 \\
        OrganC & 11.45 & \redunderline{16.95} & \redunderline{23.87} & 16.93 & \redunderline{18.81} & 15.09 & \redunderline{20.32} & 16.23 & \redunderline{14.36} & 11.20 \\
        OrganS & 13.83 & \redunderline{15.71} & \redunderline{21.52} & 20.16 & \redunderline{16.77} & 15.38 & 20.13 & \redunderline{22.46} & \redunderline{21.66} & 18.33 \\
        \midrule
        \textbf{Average} & 6.69 & \redunderline{8.92} & 7.05 & \redunderline{8.03} & 5.49 & \redunderline{6.99} & 6.12 & \redunderline{8.95} & \redunderline{6.76} & 6.58 \\
        \midrule
        Counts & 5 & \redunderline{6} & 4 & \redunderline{7} & 5 & \redunderline{6} & 2 & \redunderline{10} & \redunderline{6} & 5 \\
        \bottomrule
    \end{tabular}
    }
    \label{table:accuracy_drop_resnet50_vit_small}
\end{table}

\label{sec:experiments-rn50-benchmark}

Next, we examine the impact of using self-supervised \imagenet initialization, as opposed to random or supervised \imagenet initialization, on in-domain performance using MoCo v3, SimCLR, BYOL, and DINO. 
A paired t-test comparing supervised and self-supervised \imagenet initialization across all datasets and methods, using normalized accuracies to account for varying dataset difficulty levels, revealed no statistically significant difference in performance (t(43) = 1.167, p = 0.2496, mean normalized difference = 0.26$\sigma$, 95\% CI: [-0.19$\sigma$, 0.70$\sigma$]). Individual accuracy scores can be found in Appendix~\ref{sec:appendix_ap_idp_lat}.

\begin{figure}[h]
    \centering
    \includegraphics[width=0.8\columnwidth]{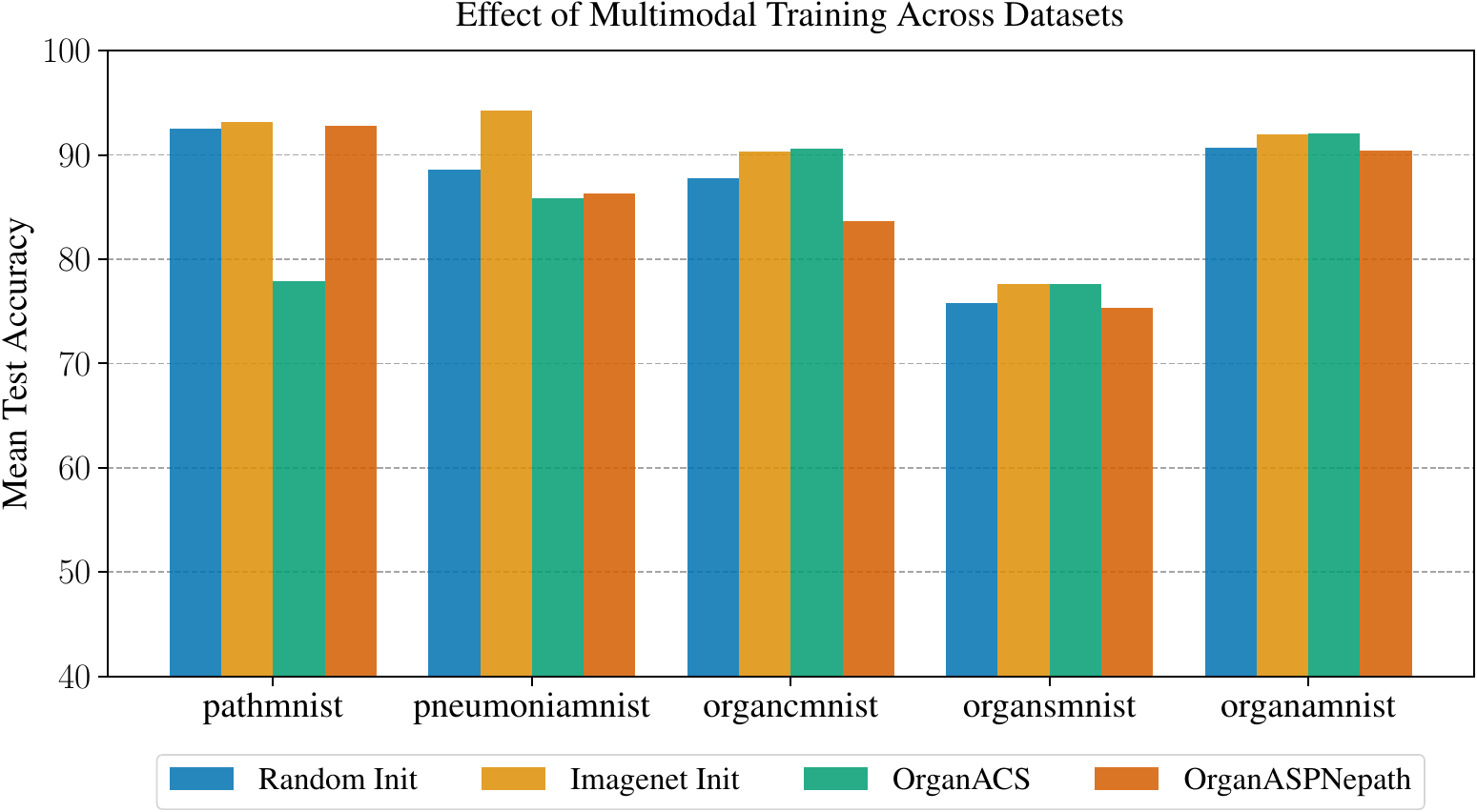}
    \captionsetup{justification=centering}
    \caption{
       Mean ID performance of SSL methods (ResNet-50), trained on single-domain datasets with different initialization strategies and multi-domain datasets with random initialization.
    }
    \label{fig:resnet50_pretraining_effect_average}
\end{figure}

Finally, we examine the impact of multi-domain training on in-domain performance. As shown in Figure \ref{fig:resnet50_pretraining_effect_average}, 
pre-training on a multi-domain dataset consisting of similar domains (OrganA, OrganC, and OrganS) with random initialization outperforms training on individual single-domain datasets with random initialization. Furthermore, it achieves performance comparable to single-domain training with \imagenet-initialization. In contrast, combining datasets from diverse domains (OrganA, OrganS, Pneumonia, and Path) does not significantly improve accuracy, and may even result in poorer performance.
These findings suggest that multi-domain pretraining is beneficial when the domains of combined datasets are similar, whereas combining diverse domains may not provide performance benefits.

    \subsection{Out-of-Distribution Detection}
We perform several experiments to assess OOD detection performance of each SSL method. In each experiment, one dataset serves as the ID dataset, while others are treated as OOD for evaluation; resulting in 110 OOD tests ($11 \times 10$) per method. The pre-trained encoder, trained on the ID dataset, is evaluated for its ability to detect OOD samples.


\begin{figure}[h]
    \centering
    \includegraphics[width=0.9\linewidth]{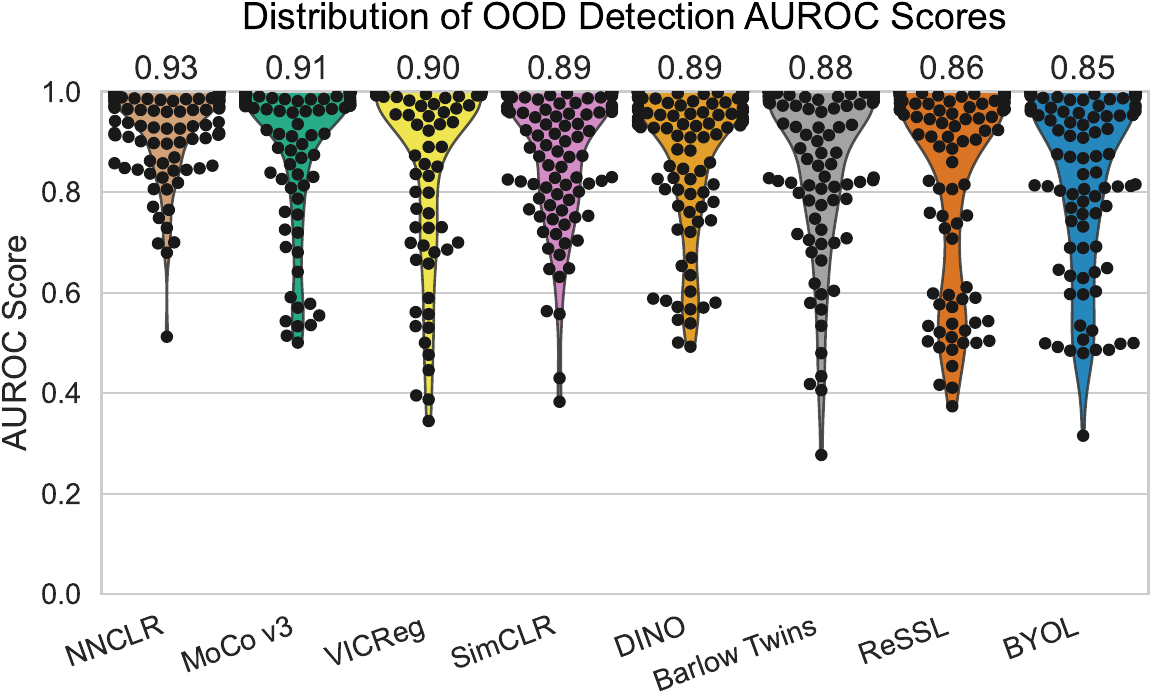}
    \caption{AUROC score distributions for models with randomly initialized weights and a ResNet-50 backbone in OOD detection. Models are ordered by mean AUROC scores (shown above), calculated by averaging over all $(\mathcal P_{\text{ID}}, \mathcal P_{\text{OOD}})$ combinations. Black dots indicate individual AUROC scores.}
    \label{fig:resnet50_method_auroc}
\end{figure}

Figure \ref{fig:resnet50_method_auroc}  illustrates the distribution of AUROC scores for OOD detection across various SSL methods using randomly initialized ResNet-50 backbones. Among these, NNCLR and MoCo v3 achieve the highest AUROC scores, suggesting that these methods are particularly effective in learning representations that differentiate ID from OOD samples. When evaluated with a ViT-Small backbone,  MoCo v3 demonstrated strong OOD detection performance, which aligns with expectations as it was originally developed for the ViT architecture.
Detailed analysis of effect of backbone choice on each method and in-distribution, OOD pairs $(\mathcal{P}_{\text{ID}}, \mathcal{P}_{\text{OOD}})$ can be found in Appendix \ref{sec:appendix_ap_ood_backbone}.

\begin{figure}[h]
    \centering
    \includegraphics[width=0.9\linewidth]{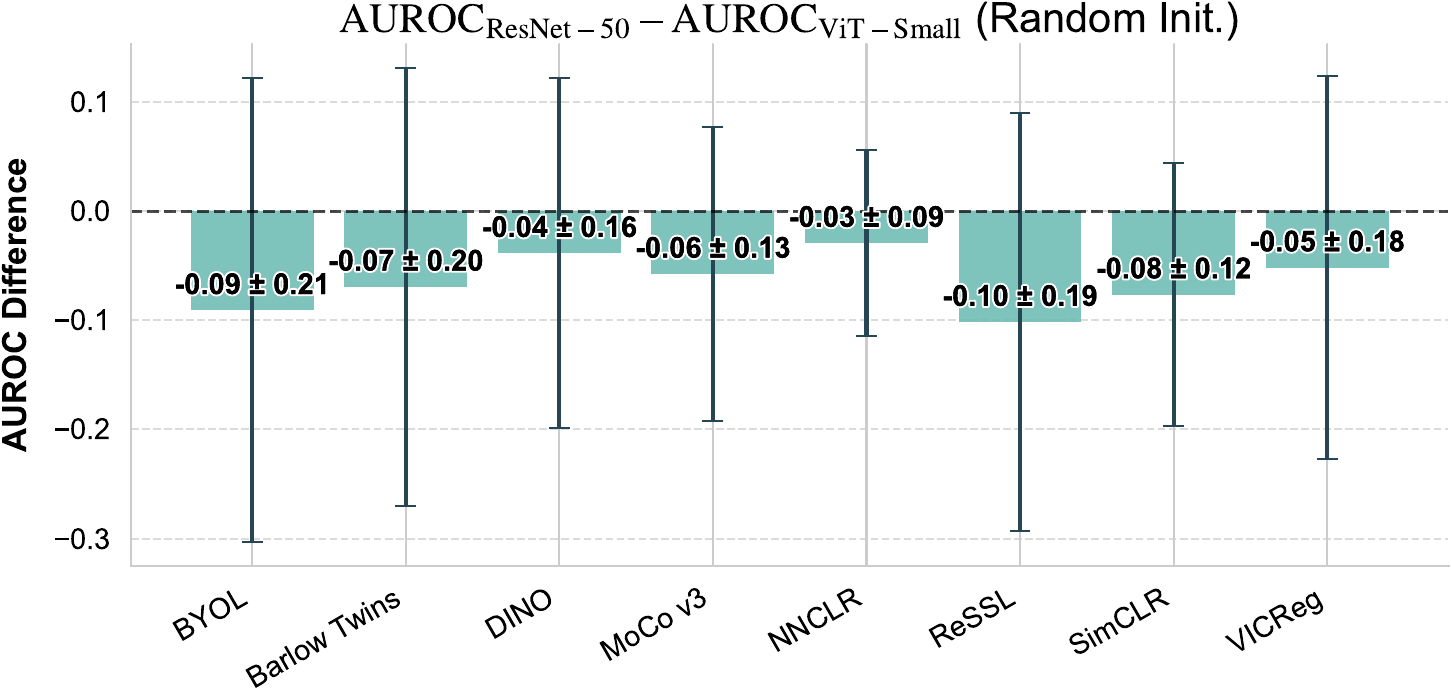}
    \caption{Mean AUROC score differences between ResNet-50 and ViT-Small models calculated over all $(\mathcal{P}_{\text{ID}},\mathcal{P}_{\text{OOD}})$ pairs. Negative values indicate a performance advantage for ViT-Small, while positive values favor ResNet-50. All differences are statistically significant ($p < 0.05$), demonstrating a clear overall advantage of ViT-Small models for OOD detection.}
    
    \label{fig:resnet50_vs_vit}
\end{figure}

In our experimental setup, the choice of backbone architecture played a crucial role in OOD detection performance. Specifically, models using ViT-Small consistently outperformed those using ResNet-50, as shown in Figure \ref{fig:resnet50_vs_vit}. This observation aligns with prior research, such as \cite{galil2023frameworkbenchmarkingclassoutofdistributiondetection,vit_under_shifts}, which demonstrated that Transformer-based architectures like ViT generally excel in OOD detection tasks. Furthermore, this trend held across all evaluated SSL methods, suggesting that the improved AUROC scores result from the ViT backbone itself rather than any specific SSL method. 

We observe that the impact of initialization on OOD detection varies depending on the choice of backbone, SSL method, and in-domain dataset distribution $\mathcal{P}_{\text{ID}}$. Generally, for smaller datasets such as BreastMNIST, RetinaMNIST, and DermaMNIST, models initialized with \imagenet weights demonstrate improved performance. For the ResNet-50 backbone, SSL methods such as SimCLR, ReSSL, MoCo v3, and VICReg enhanced OOD detection performance with random initialization. In the case of ViT-Small backbone models, although differences in OOD detection scores were less pronounced, there was still a noticeable tendency for models to favor random initialization, with VICReg showing the strongest preference.

Moreover, initializing the models with self-supervised trained \imagenet weights led to significant differences in AUROC scores compared to starting with supervised \imagenet weights for several datasets. In particular, significant negative differences (i.e., lower AUROC with self-supervised initialization) were observed for BreastMNIST ($\Delta \mathrm{AUC}=-0.12$, 95\% CI: $\pm0.18$), PathMNIST ($\Delta \mathrm{AUC}=-0.09$, 95\% CI: $\pm0.13$), and RetinaMNIST ($\Delta \mathrm{AUC} = -0.03$, 95\% CI: $\pm0.09$). Conversely, significant positive differences (i.e., higher AUROC with self-supervised initialization) were found for PneumoniaMNIST ($\Delta \mathrm{AUC} = 0.08$, 95\% CI: $\pm0.15$) and TissueMNIST ($\Delta \mathrm{AUC} = 0.08$, 95\% CI: $\pm0.19$). 
Detailed information about initialization preferences for each SSL method and backbone can be found in Appendix~\ref{sec:appendix_ap_ood_initialization}.

\begin{figure}[h]
    \centering
    \includegraphics[width=\linewidth]{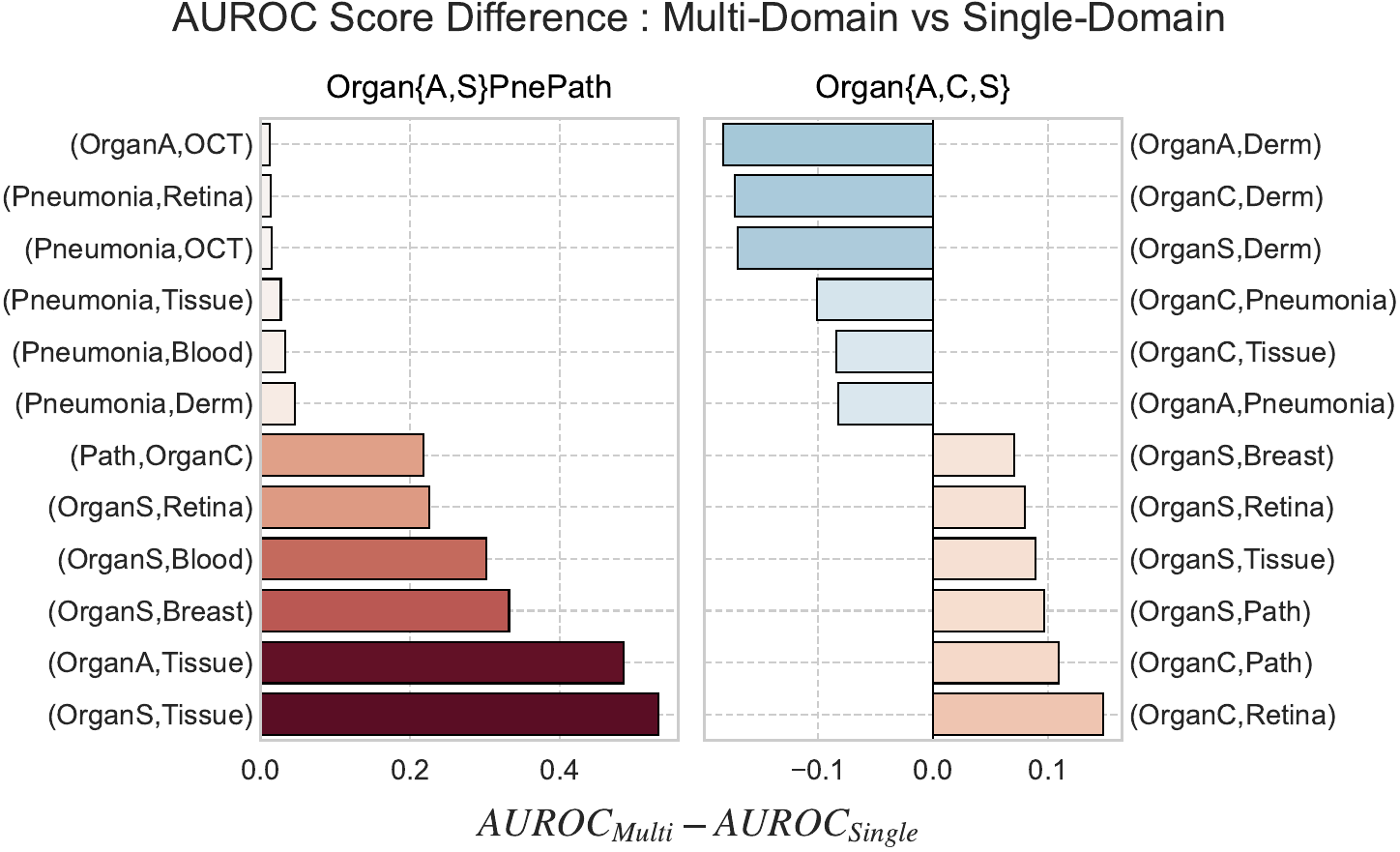}
    \caption{Mean average differences in AUROC scores between models trained on multi-domain and single-domain datasets. The y-axis lists $(\mathcal P_{\text{ID}}, \mathcal P_{\text{OOD}})$ pairs where. Positive values (red) indicates a favor for models trained on multi-domain dataset compared to its single-domain counterpart. For clarity, only the top/bottom 6 pairs with the highest/lowest AUROC score differences are shown. 
    }
    \label{fig:multimodality_comparison}
\end{figure}

Lastly, we compare OOD detection performance between models trained on multi-domain datasets (Organ\{A,C,S\} and Organ\{A,S\}PnePath) versus their constituent single-domain datasets. Figure~\ref{fig:multimodality_comparison} shows that for all SSL methods, Organ\{A,S\}PnePath consistently improves AUROC scores across most $(\mathcal P_{\text{ID}}, \mathcal P_{\text{OOD}})$ pairs, with the exception of a single pair. In contrast, Organ\{A,C,S\} generally decreases OOD detection performance across most $(\mathcal P_{\text{ID}}, \mathcal P_{\text{OOD}})$ pairs, particularly for models trained with MoCo v3, SimCLR, and ReSSL. These findings demonstrate that a multi-domain dataset with greater domain diversity (Organ\{A,S\}PnePath) provides a more substantial OOD detection performance boost compared to a more homogeneous multi-domain dataset (Organ\{A,C,S\}).

\subsection{Generalizability}

In this section, we evaluate transferability of representations learned from SSL methods by performing cross-dataset evaluations.
Specifically, we first pre-train the encoder on one of the datasets and then train a linear classifier on each of the remaining datasets individually, using a frozen encoder. 
We evaluate the generalizability of learned representations by calculating the drop in accuracy relative to the in-domain performance as, $\Delta\text{ACC} = \tfrac{\text{ACC}_\text{ID} - \text{ACC}_\text{transfer}}{\text{ACC}_\text{ID}} \times 100$, where $\text{ACC}_\text{ID}$ represents the accuracy when training and testing are performed on the same dataset (in-domain), and $\text{ACC}_\text{transfer}$ denotes the accuracy when the model is pre-trained on a different dataset. For each source dataset, we calculate the average accuracy drop ($\overline{\Delta\text{ACC}}$) across all target datasets, providing insights into how well a method trained on one dataset performs when transferred to others and which datasets yield the most transferable knowledge. 

Table~\ref{tab:transfer_acc_drop} shows that SimCLR and MoCo v3 excel in generalization, with both achieving the lowest accuracy drops in 4 datasets. Notably, models pre-trained on PathMNIST or Organ\{C,S\}MNIST show the lowest accuracy drops, suggesting these datasets offer the most transferable representations.

\begin{table}[h]
    \centering
    \caption{\textbf{Average Accuracy Drops in Transfer Performance:} The accuracy drop percentages, relative to the mean in-domain performance using ResNet-50 with random initialization are averaged over all possible target datasets. The lowest accuracy drops are highlighted in \colorbox{green!20}{green}, indicating better generalizability. Due to space constraints, we show only five of the eight evaluated methods, representing the best-performing candidates from the full set.}
    \label{tab:transfer_acc_drop}
\scalebox{0.65}{\begin{tabular}{l|c|c|c|c|c}
\toprule
{} & \multicolumn{5}{c}{Average Accuracy Drop  $\Delta_\text{ACC}$ (percentage)} \\ \cline{2-6}  \\[-0.9em]
\multirow{-2}{*}{\centering\makebox[1.5cm]{Source}} & SimCLR & MoCo v3 & Barlow Twins & NNCLR & ReSSL \\
\midrule
Path & 6.44\scriptsize{$\pm$5.78} & 6.38\scriptsize{$\pm$5.99} & \cellcolor{green!20}3.32\scriptsize{$\pm$5.82} & 6.80\scriptsize{$\pm$6.41} & 10.10\scriptsize{$\pm$7.54} \\
Derma & \cellcolor{green!20}11.16\scriptsize{$\pm$7.00} & 16.06\scriptsize{$\pm$8.55} & 13.32\scriptsize{$\pm$8.77} & 19.81\scriptsize{$\pm$10.60} & 14.08\scriptsize{$\pm$10.36} \\
OCT & 14.18\scriptsize{$\pm$5.06} & 8.62\scriptsize{$\pm$4.87} & \cellcolor{green!20}8.05\scriptsize{$\pm$4.52} & 11.40\scriptsize{$\pm$5.68} & 12.87\scriptsize{$\pm$6.67} \\
Pneumonia & 11.46\scriptsize{$\pm$7.84} & \cellcolor{green!20}9.77\scriptsize{$\pm$8.73} & 17.52\scriptsize{$\pm$9.23} & 10.40\scriptsize{$\pm$7.40} & 14.96\scriptsize{$\pm$8.29} \\
Retina & \cellcolor{green!20}13.73\scriptsize{$\pm$8.22} & 13.96\scriptsize{$\pm$8.25} & 31.55\scriptsize{$\pm$14.10} & 27.84\scriptsize{$\pm$13.86} & 16.14\scriptsize{$\pm$9.26} \\
Breast & \cellcolor{green!20}12.60\scriptsize{$\pm$7.84} & 23.16\scriptsize{$\pm$11.43} & 18.46\scriptsize{$\pm$10.14} & 39.06\scriptsize{$\pm$17.18} & 43.00\scriptsize{$\pm$18.46} \\
Blood & 11.55\scriptsize{$\pm$9.47} & \cellcolor{green!20}9.37\scriptsize{$\pm$8.53} & 33.78\scriptsize{$\pm$16.95} & 11.14\scriptsize{$\pm$7.59} & 12.04\scriptsize{$\pm$8.49} \\
Tissue & 9.09\scriptsize{$\pm$5.21} & \cellcolor{green!20}7.70\scriptsize{$\pm$7.43} & 8.86\scriptsize{$\pm$8.36} & 9.09\scriptsize{$\pm$7.66} & 12.57\scriptsize{$\pm$7.26} \\
OrganA & 10.14\scriptsize{$\pm$8.08} & 6.24\scriptsize{$\pm$7.50} & 7.17\scriptsize{$\pm$6.53} & \cellcolor{green!20}5.62\scriptsize{$\pm$8.09} & 7.97\scriptsize{$\pm$7.56} \\
OrganC & 6.12\scriptsize{$\pm$5.92} & \cellcolor{green!20}5.54\scriptsize{$\pm$7.73} & 7.15\scriptsize{$\pm$6.25} & 6.22\scriptsize{$\pm$6.22} & 6.81\scriptsize{$\pm$7.00} \\
OrganS & \cellcolor{green!20}5.68\scriptsize{$\pm$7.00} & 6.60\scriptsize{$\pm$6.37} & 7.01\scriptsize{$\pm$6.38} & 6.02\scriptsize{$\pm$6.12} & 8.73\scriptsize{$\pm$6.26} \\
\bottomrule
\end{tabular}
}
\end{table}

We further examine the impact of supervised \imagenet initialization on model generalizability.
To evaluate the overall impact of \imagenet initialization compared to random initialization, we conducted paired $t$-tests for each test dataset across all cross-dataset training combinations including the in-domain setting where test and train splits come from the same dataset. The results demonstrate statistically significant improvements $(p<0.05)$ across all datasets, with PneumoniaMNIST showing the most modest relative gain ($1.88\pm 3.28\%$) and OCTMNIST exhibiting the largest improvement ($13.22\pm 25.59\%$) when we switch from random weights to \imagenet initialization. 
Detailed quantitative results are provided in Appendix~\ref{sec:appendix_ap_g}. While previous studies have raised concerns about the efficacy of transfer learning from natural to medical images due to significant distributional differences~\cite{raghu2019transfer}, our findings show that continual pre-training~\cite{gururangan2020dontstoppretrainingadapt, taher2021systematicbenchmarkinganalysistransfer} improves both in-domain performance and generalization across diverse medical datasets.

\begin{figure}[h]
    \centering
    \hspace*{-0.2cm} 
    \includegraphics[width=0.8\columnwidth]{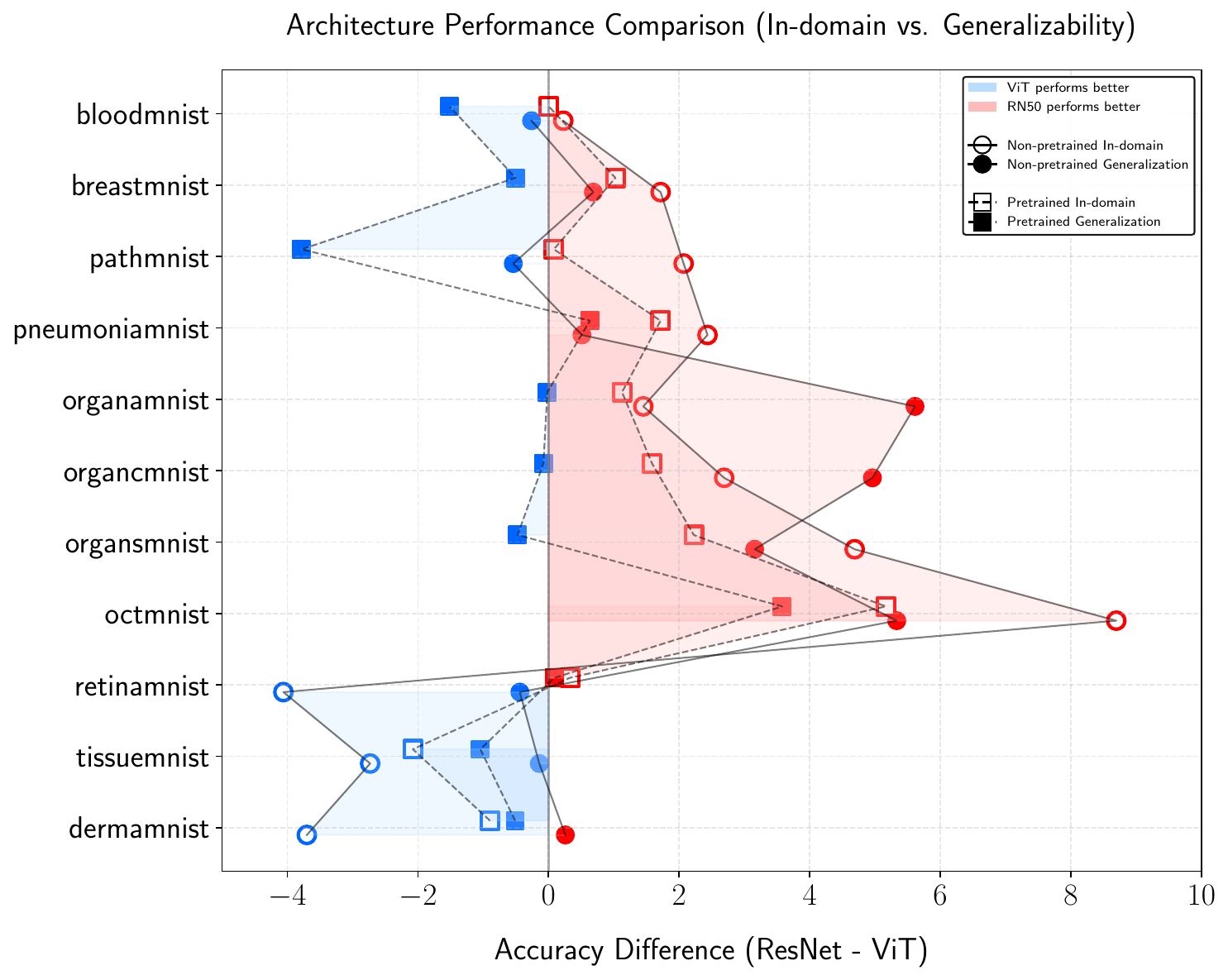}
    \caption{Performance differential between ResNet-50 and ViT (averaged over methods) for both in-domain and cross-dataset evaluations using two different initialization strategies. 
    } 
    \label{fig:generalizability_arch.pdf}
\end{figure}

ResNet-50 outperformed ViT-Small in most in-domain settings (8/11 datasets when trained from scratch and 9/11 when initialized with ImageNet weights). However, as Figure~\ref{fig:generalizability_arch.pdf} demonstrates, the performance gap between these architectures narrows considerably when assessing generalization rather than in-domain performance. Notably, in the \imagenet pre-trained generalization scenario, ViT-S actually surpasses ResNet-50 on 8/11 datasets, reversing the previous trend. This suggests that ViT-S benefits more from transfer learning than ResNet-50 when applied to new datasets - supporting a similar finding \cite{matsoukas2021timereplacecnnstransformers}.

\begin{figure}[h]
    \centering
    \hspace*{-0.4cm} 
    \includegraphics[width=1\columnwidth]{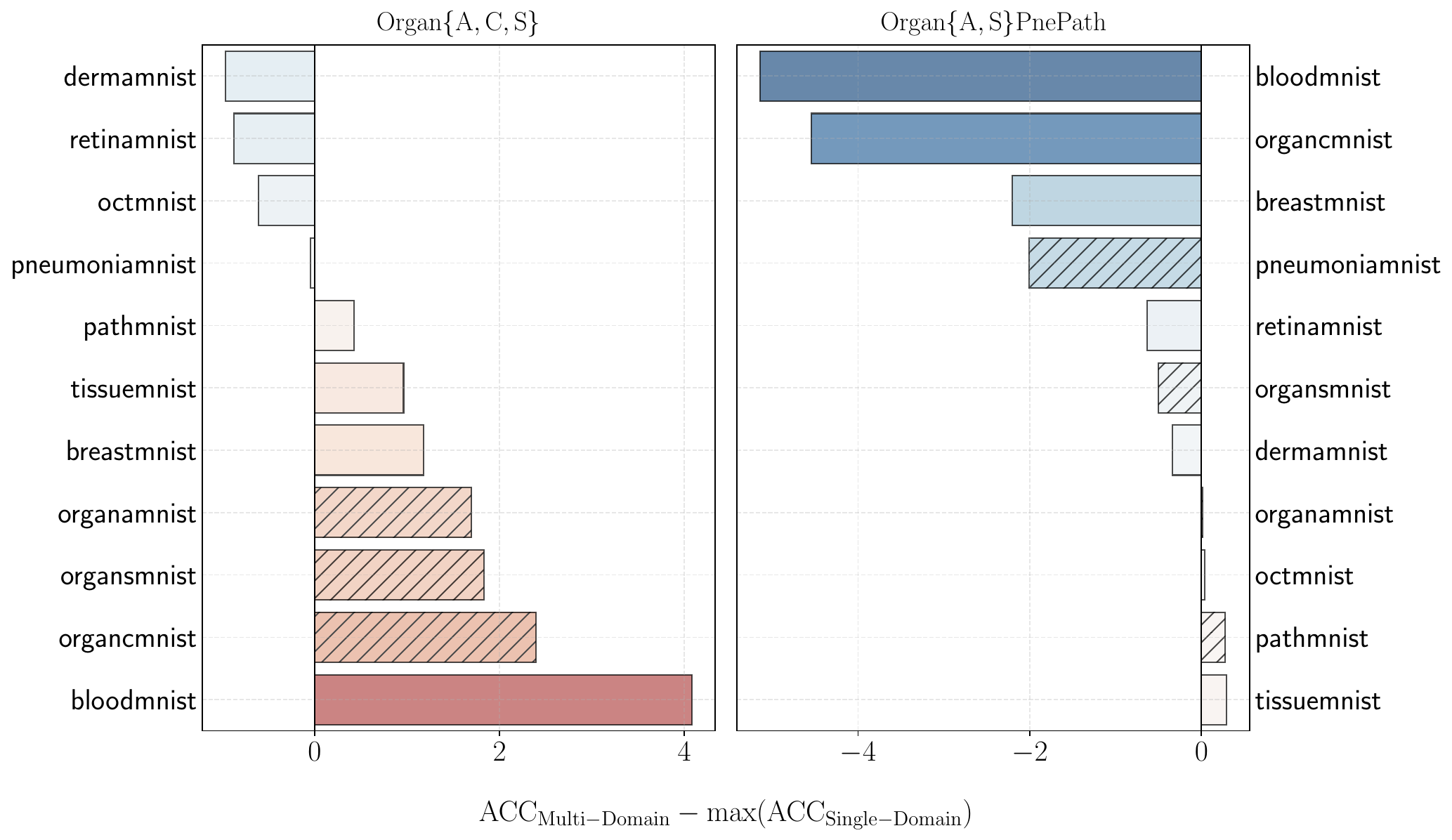}
    \caption{
   Comparison of performance differences between multi-domain and single-domain models. The bars show the average accuracy difference over five methods between the multi-domain model and the best-performing single-domain model accuracy for that dataset, with in-domain tasks highlighted using dashed lines.}
    \label{fig:organs_heatmap.pdf}
\end{figure}

Finally, we evaluate the generalizability of representations learned through multi-domain training. Figure~\ref{fig:organs_heatmap.pdf} compares the cross-dataset performance of models trained on multi-domain datasets with models trained on their individual constituents. Training with similar domain datasets (Organ\{A,C,S\}) enhances cross-dataset transfer performance, suggesting that multi-domain training within a single modality may act as a regularizer. However, this benefit does not hold for mixed-modality combinations (Organ\{A,S\}PnePath), showing that multi-domain training benefits are modality-dependent (see Appendix~\ref{sec:appendix_ap_g}).


\section{Discussion \& Conclusion}
\label{sec:conclusion}

We have conducted a comprehensive evaluation of popular SSL methods in medical imaging, addressing a crucial gap in understanding how these methods perform across diverse medical tasks. Based on our results and analysis, we offer the following recommendations for practitioners:

\vspace{-12px}
\paragraph*{Which self-supervised method to choose?} MoCo v3 demonstrates remarkable versatility in medical imaging tasks when trained from scratch, achieving superior performance in 5/11 datasets for in-domain tasks and maintaining minimal accuracy drops in 4/11 datasets for cross-domain generalization. It also shows competitive performance in OOD detection tasks.  However, the effectiveness of certain SSL methods is significantly influenced by initialization strategy. Notably, DINO and BYOL transform from being among the lowest performers to achieving competitive in-domain results when initialized with \imagenet weights, with DINO particularly excelling with ViT.

\vspace{-12px}
\paragraph*{Should we start self-supervised pre-training with supervised \imagenet weights?} Self-supervised pretraining on medical images, initialized with \imagenet weights, significantly outperforms random initialization for in-domain downstream tasks, mirroring the established benefits previously observed in supervised learning setting \cite{Tajbakhsh_2016}.  Despite concerns that supervised pre-training might yield less general representations due to label dependency \cite{liu2022selfsupervisedlearningrobustdataset, sariyildiz2023reasonsupervisionimprovedgeneralization}, \imagenet initialization enhances both in-domain and cross-dataset performance. However, this advantage does not consistently extend to OOD detection, as \imagenet initialization degrades performance for certain backbone and model combinations. Overall, supervised and self-supervised \imagenet initialization offer comparable performance for in-domain accuracy and OOD detection with no clear advantage for either.


\vspace{-12px}

\paragraph*{Which model architecture is better?} Choosing between these architectures involves key trade-offs: While ResNet-50 excels in in-domain classification (outperforming ViT-Small in 8/11 datasets when trained from scratch and 9/11 with ImageNet initialization), ViT-Small demonstrates superior performance in OOD detection. Interestingly, when trained with ImageNet initialization and evaluated on generalization tasks, ViT-Small reverses the in-domain trend, surpassing ResNet-50 in 8/11 datasets. However, ViT shows greater sensitivity to label scarcity, with performance dropping more steeply when training data is limited. Consistent with \citet{matsoukas2021timereplacecnnstransformers}, ImageNet initialization reduces the performance gap between architectures, though these fundamental trade-offs remain.

\vspace{-12px}
\paragraph*{Is multi-domain SSL effective?} 
Dataset composition plays a crucial role in model performance.
Models trained on heterogeneous datasets (Organ\{A,S\}PnePath) demonstrated superior OOD detection performance, achieving consistently higher AUROC scores compared to models trained on homogeneous (Organ\{A,C,S\}) or single-domain datasets. In contrast, Organ\{A,C,S\} often reduces OOD detection performance, underscoring the limitations of less diverse datasets. While Organ\{A,C,S\} yields better in-domain accuracy and more generalizable representations compared to individual Organ datasets, this pattern does not hold for the more diverse Organ\{A,S\}PnePath.

\vspace{-12px}
\paragraph*{Extended Evaluation of Key Findings:} We conducted a series of short experiments using bigger architectures, higher resolution images, and an additional dataset AIROGS \cite{devente2023airogsartificialintelligencerobust}.  

\textbf{Larger backbones:} Using RN-101 ($\sim$45M parameters) and ViT-B ($\sim$86M parameters), we conducted generalizability, OOD, and in-domain experiments on \underline{DermaMNIST}, plus in-domain tests on \underline{BloodMNIST} with MoCo v3 and DINO, yielding consistent results. In cross-dataset evaluations, \imagenet init. outperforms random init. ($\Delta{=}11.0\%$) and MoCo v3 surpasses DINO ($\Delta{=}10.5\%$) across all settings. For OOD detection, ViT shows superior performance ($\Delta\text{AUC}{=}0.026$, 95\% CI:$\pm 0.019$) (Fig. \ref{fig:resnet50_vs_vit}) and MoCo v3 outperforms DINO when using RN-50 with random init. ($\Delta\text{AUC}{=}0.037$, 95\% CI:$\pm 0.032$). For in-domain results (Fig. \ref{fig:id_pretrain.pdf}), \imagenet init. improves in-domain accuracy across methods and architectures ($\Delta{=}7.64\%$) considering both datasets. 

\textbf{Additional dataset:} We trained MoCo v3 and BYOL on \underline{AIROGS}, a large real-world dataset, containing \raisebox{0.5ex}{\texttildelow}113K color fundus images from \raisebox{0.5ex}{\texttildelow}60K patients. 
Aligned with our findings, ViT dominates 
RN-50 with random initialization ($\Delta\text{AUC}{=}0.05$, 95\% CI:$\pm 0.041$) in OOD detection and MoCo v3 outperforms BYOL in cross-dataset evaluations ($\Delta{=}2.72\%$) and across all initializations and architectures in OOD detection ($\Delta\text{AUC}{=}0.081$, 95\% CI:$\pm 0.041$).

 \textbf{Higher resolution:} We trained MoCo v3, BYOL, DINO, SimCLR using 224×224 \underline{OrganAMNIST}. Results were consistent with 64×64: For OOD detection, ViT-S outperforms RN50 ($\Delta$AUC{=}0.097, 95\% CI:$\pm 0.073$), and performance hierarchy (Fig. \ref{fig:resnet50_method_auroc}) among RN50 models with random init. is: MoCo v3 (0.93) $>$ DINO (0.91) $>$ BYOL (0.90) $>$ SimCLR (0.82). Also, \imagenet init. leads to higher cross-dataset performance across methods and architectures ($\Delta{=}0.84\%$ in average). RN50 achieves better in-domain performance than ViT-S (Fig. \ref{fig:id_arch.pdf}), where the performance gap shrinks when switching from random init. ($\Delta{=}2.44\%$) to \imagenet initialization ($\Delta{=}1.66\%$).
 \section*{Acknowledgements}

 The author acknowledges support from the Zuse School ELIZA, which provided essential funding that enabled their contributions to this work. 

{
    \small
    \bibliographystyle{ieeenat_fullname}
    \bibliography{main}
}
\onecolumn
\newpage

\renewcommand{\thesubsection}{\Alph{subsection}}
\startcontents[subsections]
\stopcontents[subsections]

\section*{Appendix}
\label{sec:appendix}

\resumecontents[subsections]
\printcontents[subsections]{l}{1}[4]{}  


\subsection{Dataset Details}
\label{sec:appendix_dset}
MedMNIST is a medical imaging dataset comprising 18 sub-datasets from various medical domains \cite{Yang_2023}. For our work, we focus on 12 of these sub-datasets, specifically those containing 2D images. MedMNIST offers a diverse representation of medical imaging modalities and supports multiple classification tasks, making it a comprehensive resource for benchmarking. Its extensive adoption in medical image recognition research \cite{ozkan2024multidomainimprovesoutofdistributiondatalimited, Narayanaswamy_2023_ICCV, doerrich2024rethinkingmodelprototypingmedmnist, huang2024systematiccomparisonsemisupervisedselfsupervised} further establishes its value as a standard benchmark. Accordingly, we utilize the 2D subset of MedMNIST, referred to as MedMNIST2D, which we simply denote as MedMNIST in the paper.

The MedMNIST dataset was initially introduced in a resolution of $28 \times 28$ pixels and has since been expanded by MedMNIST+ to additionally include $64 \times 64$, $128 \times 128$, and $224 \times 224$ resolutions. Due to limited GPU memory, we opted to use the $64 \times 64$ resolution version of the dataset throughout our experiments. As lower resolutions are also shown to yield reasonable accuracies in previous research \cite{huang2024systematiccomparisonsemisupervisedselfsupervised, Yang_2023}, we do not sacrifice much by not using a larger resolution. 
Detailed information regarding each dataset, such as data source, domain, classification task type (including the number of classes), and publicly available data splits that correspond to our benchmark are provided in \tablename~\ref{tab:medmnist2d}.

The datasets encompass diverse imaging modalities, including X-ray, CT, ultrasound, fundus camera, dermatoscope, and microscope images. Beyond spanning multiple medical domains, these modalities vary in technical characteristics such as color channels and level of detail. This diversity makes it particularly well-suited for self-supervised learning tasks, as it captures the unique challenges and nuances of different medical imaging fields. Such variability enables the development and rigorous evaluation of models that must generalize effectively across different image types—a critical requirement for robust medical image analysis. Leveraging MedMNIST not only facilitates the creation of domain-agnostic representations but also provides a platform for testing these representations across a wide range of medical imaging scenarios.

The classification tasks supported in MedMNIST are diverse, encompassing multi-class classification (MC), binary classification (BC), and ordinal regression (OR). Multi-class classification treats each class as distinct and independent, without accounting for relationships among them, while binary classification is a specific case of multi-class classification involving only two classes. Ordinal regression, on the other hand, is a regression task where the output represents a discrete value reflecting the ordered relationship between classes. In our experiments, we treat ordinal regression as multi-class classification by assigning each ordinal level as a separate class. While this method loses information on the ordinal relationships between classes, it enables consistent evaluation across all datasets in our benchmark. 
We excluded ChestMNIST, as explained in Section~\ref{sec:methodology-tasks-and-datasets}, since it involves multi-label classification, while our study focuses solely on single-label classification tasks.


\begin{table*}[h]
\scriptsize
\begin{adjustbox}{width=500pt,center}
\begin{tabular}{ccccc}
\hline
\textbf{Dataset}  & \textbf{Data Modality} & \textbf{Task (\# Classes / Labels)} & \textbf{\# Samples} & \textbf{\# Training / Validation / Test} \\ \hline 
PathMNIST \cite{kather2019predicting} & Colon Pathology & MC (9) & 107,180 & 89,996 / 10,004 / 7,180 \\ 
DermaMNIST \cite{tschandl2018ham10000, codella2019skin} & Dermatoscope & MC (7) & 10,015 & 7,007 / 1,003 / 2,005 \\ 
OCTMNIST \cite{kermany2018identifying}  & Retinal OCT & MC (4) & 109,309 & 97,477 / 10,832 / 1,000 \\
PneumoniaMNIST \cite{kermany2018identifying} & Chest X-Ray & BC (2) & 5,856 & 4,708 / 524 / 624 \\ 
RetinaMNIST \cite{liu2022deepdrid} & Fundus Camera & OR (5) & 1,600 & 1,080 / 120 / 400 \\ 
BreastMNIST \cite{al-dhabyani2020dataset} & Breast Ultrasound & BC (2) & 780 & 546 / 78 / 156 \\ 
BloodMNIST \cite{acevedo2020dataset} & Blood Cell Microscope & MC (8) & 17,092 & 11,959 / 1,712 / 3,421 \\ 
TissueMNIST \cite{ljosa2012annotated} & Kidney Cortex Microscope & MC (8) & 236,386 & 165,466 / 23,640 / 47,280 \\
OrganAMNIST \cite{bilic2023lits, xu2019efficient} & Abdominal CT & MC (11) & 58,850 & 34,581 / 6,491 / 17,778 \\
OrganCMNIST \cite{bilic2023lits, xu2019efficient} & Abdominal CT & MC (11) & 23,660 & 13,000 / 2,392 / 8,268 \\ 
OrganSMNIST \cite{bilic2023lits, xu2019efficient} & Abdominal CT & MC (11) & 25,221 & 13,940 / 2,452 / 8,829 \\ 
\bottomrule
\end{tabular}
\end{adjustbox}
\captionsetup{justification=centering}
\caption{Overview of the MedMNIST2D datasets employed in our benchmarking study, which include tasks involving multiclass classification, binary classification, and ordinal regression.}
\label{tab:medmnist2d}
\end{table*}

\subsection{Method Details}
\label{sec:appendix_method}

Self-supervised learning (SSL) techniques are a subset of unsupervised methods focused on extracting meaningful representations from unlabeled data. These techniques have gained significant attention due to their ability to leverage large datasets and improve model performance on various downstream tasks \cite{10559458}.

SSL methods can be broadly classified into generative and discriminative approaches. Generative methods aim to reconstruct or generate data samples to capture meaningful representations, whereas discriminative methods focus on distinguishing between data points to learn robust and invariant features.
In this study, we focus on discriminative self-supervised learning methods, which are specifically designed to maximize the similarity between augmented versions of the same image (``positive pairs'') while minimizing the similarity with other images (``negative pairs''). More recent approaches eliminate the need for negative pairs, instead focusing solely on maximizing the similarity between positive pairs. These methods enable the model to learn robust features, which can then be used in many downstream visual recognition tasks. We consider the following discriminative SSL methods:

\subparagraph*{SimCLR} \cite{chen2020simclr}  (Simple Framework for Contrastive Learning of Visual Representations) is a contrastive SSL method that learns representations by maximizing the agreement between differently augmented views of the same image. SimCLR utilizes a contrastive loss function, specifically the normalized temperature-scaled cross-entropy (NT-Xent) loss, to increase the similarity between augmented pairs while minimizing it with other samples.

\vspace{-0.4cm}
\subparagraph*{DINO} \cite{caron2021dino} (Distillation with No Labels) employs a teacher-student architecture to learn from self-distilled knowledge. \textit{Self-distillation} is a process where given two different views of a sample image, the student model directly predicts the output of the teacher model. The teacher model has the same architecture as the student, but its parameters are updated using a momentum encoder on the student's parameters. Thus, the student model ``distills'' knowledge from the teacher and extracts similar features for different views, without needing labels or negative samples.

\vspace{-0.4cm}
\subparagraph*{BYOL} \cite{grill2020byol} (Bootstrap Your Own Latent) leverages two neural networks, an \textit{online network} and a \textit{target network}, to iteratively improve each other's representations. BYOL does not rely on negative pairs; instead, it minimizes the difference between the two networks' representations of the same image. The target network is updated using a moving average of the online network parameters, which in turn is updated via back-propagation.

\vspace{-0.4cm}
\subparagraph*{ReSSL} \cite{zheng2021resslrelationalselfsupervisedlearning} (Relational Self-Supervised Learning) introduces a relation metric to better capture nuanced relationships between different samples. That is, instead of strictly enforcing positive and negative pairs as in contrastive SSL, ReSSL calculates the relationship distribution among the samples, and minimizes the KL divergence between those of the views of a sample. This approach enables ReSSL to learn more nuanced inter-sample relationships.

\vspace{-0.4cm}
\subparagraph*{MoCo v3} \cite{chen2021empiricalstudytrainingselfsupervised} (Momentum Contrast v3) builds upon the original Momentum Contrast framework, which frames contrastive learning as a dictionary look-up task. In this approach, encoders are trained to ensure that the representation of a \textit{query} (\ie a data sample) is similar to its corresponding \textit{key} (\ie a positive sample) while being dissimilar to other samples (``negative keys''). Earlier versions of MoCo utilized a dynamic and memory-efficient dictionary, maintaining key representations in a queue that was updated on-the-fly using a momentum-updated encoder. MoCo v3 simplifies this design by removing the memory queue entirely and instead leveraging a purely end-to-end transformer-based architecture.

\subparagraph*{VICReg} \cite{bardes2022vicregvarianceinvariancecovarianceregularizationselfsupervised} (Variance-Invariance-Covariance Regularization) avoids using negative samples or asymmetric networks. It prevents representation collapse by combining three key terms: a variance term to ensure diversity across feature dimensions, an invariance term to align features of augmented views of the same image, and a covariance term to reduce redundancy by decorrelating feature dimensions. Together, these components enable the learning of robust and informative representations.

\vspace{-0.4cm}
\subparagraph*{Barlow Twins} \cite{zbontar2021barlowtwinsselfsupervisedlearning} is another SSL method that aims to prevent representation collapse by employing an objective function inspired by neuroscientist H. Barlow's \textit{redundancy-reduction} principle. Specifically, the cross-correlation between the outputs of a Siamese network are made as close as possible to the identity matrix. This approach reduces redundancy across feature dimensions while encouraging different views of the same sample to have similar representations, thereby learning robust and diverse features.

\vspace{-0.4cm}
\subparagraph*{NNCLR} \cite{dwibedi2021littlehelpfriendsnearestneighbor} (Nearest-Neighbor Contrastive Learning of Representations) is a contrastive SSL method that incorporates nearest neighbors as additional positive pairs. In addition to using augmented views of a sample, NNCLR finds the closest semantic matches in a dynamically updated support set and uses them as extra positive pairs in contrastive loss. This reduces the reliance on heavy augmentations and enables the model to learn features that are more stable under larger semantic variations.

\subsection{Implementation Details}
\label{sec:appendix_tr}

\subsubsection{Pre-Training}
\label{sec:appendix_tr_pt}
For training, we employ mini-batch gradient descent using the AdamW optimizer \cite{loshchilov2019decoupledweightdecayregularization} and a cosine learning rate schedule \cite{cosine} beginning with a $10$-epoch linear warm-up from $3 \times 10^{-5}$. To ensure fairness, we conduct a grid search using learning rates of the form $3 \times 10^{x}$, where $x \in \{-1, -2, -3, -4\}$, and weight decays of the form $1 \times 10^{x}$, where $x \in \{-3, -4, -5\}$. We select the model with the best performance on the downstream validation set. The batch size is set to $256$, with coefficients for computing running averages of the gradient and its square as $\beta_1 = 0.9$ and $\beta_2 = 0.95$ respectively. We train the models for a total of 400 epochs. 

For our experimental evaluation with self-supervised \imagenet pre-training, we employ four established methods: MoCo v3, SimCLR, BYOL, and DINO. The MoCo v3 weights were obtained from the \texttt{mmselfsup} library \cite{mmselfsup2021}, specifically utilizing the model variant trained for $100$ epochs with a batch size of $4,096$ on the \imagenet dataset. For SimCLR, BYOL, and DINO, we leverage pre-trained weights from the \texttt{LightlySSL} framework \cite{Susmelj_Lightly}. These models were trained for 100 epochs, with varying batch sizes: 256 for SimCLR and BYOL, and 128 for DINO.

In the \imagenet initialization experiments, the backbone weights were initialized from the \imagenet checkpoint, while auxiliary components (such as the projection head, classifier, and other task-specific layers) were discarded. We conducted experiments with multi-domain datasets using 5 SSL methods, including SimCLR, DINO, BYOL, ReSSL, and MoCo v3. First, we merged the datasets and pre-trained the encoder using the selected SSL method. Then, hyperparameter selection was performed based on the average performance of linear classifiers, which were trained independently on each constituent dataset.

\subsubsection{Data Augmentations}
\citet{azizi2022re} work on different medical modalities, including dermatology photography, fundus imaging, digital pathology, chest radiography, and mammography. Similar to their default contrastive pretraining setting, we utilized random cropping, random color distortion (brightness, contrast, saturation, and hue changes), and random Gaussian blur, along with random horizontal flipping as the data augmentation strategy.

\subsubsection{Linear Evaluation} 
\label{sec:appendix_tr_le}
We utilize the train, validation, and test set splits as provided by MedMNIST. All images are converted to RGB format and normalized.
We use stochastic gradient descent with a step learning rate scheduler, decaying at epochs $[60, 80]$. For a fair comparison, we conduct a grid search over learning rates \{$0.1, 0.01, 0.001$\} and weight decay values \{$0, 0.1, 0.01$\}, selecting the best hyperparameters based on validation set performance. The test set is used only once at the end to evaluate the classifier with the optimal hyperparameters. We train the linear classifier for $100$ epochs.
 We report the mean and standard deviation of performance metrics on downstream tasks over five runs. Assuming a \textit{t}-distribution, we calculate the confidence intervals as $\bar{x} \pm t \times \text{SE}$, where $\bar{x}$ represents the mean metric and $\text{SE}$ is the standard error, calculated as $\text{SE} = \frac{\text{SD}}{\sqrt{n}}$ where $n$ is the number of trials and $\text{SD}$ is the 
 standard deviation. 

\subsubsection{Loss Criterion and Evaluation Metrics}
\label{sec:appendix_tr_eval}
Our downstream tasks include multiclass classification (MC), binary classification (BC), and ordinal regression (OR). To maintain consistency with the experiments conducted by \citet{Yang_2023}, we used the cross-entropy loss for all classification tasks, including ordinal regression. 
We adopted the same evaluation metrics as in the previous work, including accuracy (ACC) and the area under the receiver operating characteristic curve (AUC or AUROC), to assess the models' ability to differentiate between classes. For multiclass classification, we employ macro averaging, calculating the metric independently for each class and then averaging the results. We used different numbers of epochs for pre-training, to accommodate the differences in the training set sizes. 

\subsubsection{Out-of-distribution Detection}
\label{sec:appendix_tr_ood}

A sample $x$ is considered \textit{in-distribution} if drawn from the training distribution $\mathcal{P}_\text{ID}$, 
and \textit{out-of-distribution} if drawn from a different distribution $\mathcal{P}_\text{OOD} \neq \mathcal{P}_\text{ID}$, representing different domains or modalities. To determine whether a sample originates from the in-distribution (ID) $\mathcal{P}_{\text{ID}}$ or out-of-distribution (OOD) $\mathcal{P}_{\text{OOD}}$, we primarily adopt the \textit{Mahalanobis Distance}~\cite{lee2018simpleunifiedframeworkdetecting} as the core metric. While energy-based~\cite{energy_ood} and softmax-based~\cite{softmax_ood} methods are also considered, our experiments consistently demonstrate that the Mahalanobis distance-based approach yields superior OOD detection performance. This approach computes the distance between the extracted feature vector of an input sample and the nearest class-conditional Gaussian distribution, facilitating the assignment of pseudo-labels to features. Formally, let $\mathbf{f}(x) \in \mathbb{R}^d$ denote the feature representation of input $x$ extracted by a trained encoder. Assuming that the feature representations of each class follow a multivariate Gaussian distribution, we define the class-conditional distributions as:
\[
p(\mathbf{f}(x) | y = c) = \mathcal{N}(\mathbf{f}(x) | \boldsymbol{\mu}_c, \Sigma),
\]
where $\boldsymbol{\mu}_c \in \mathbb{R}^d$ and $\Sigma \in \mathbb{R}^{d \times d}$ represent the class mean and a shared covariance matrix, respectively. These parameters are estimated empirically from the training data  $\mathcal{P}_{\text{ID}}^{\text{train}}$:
\[
\boldsymbol{\mu}_c = \frac{1}{N_c} \sum_{i: y_i = c} \mathbf{f}(x_i), \quad \Sigma = \frac{1}{N} \sum_{c=1}^m \sum_{i: y_i = c} (\mathbf{f}(x_i) - \boldsymbol{\mu}_c)(\mathbf{f}(x_i) - \boldsymbol{\mu}_c)^\top,
\]
where $N_c$ is the number of training samples in class $c$, and $N$ is the total number of training samples.

The Mahalanobis distance between a sample's feature vector $\mathbf{f}(x)$ and a class-conditional distribution is given by:
\[
D_\text{M}(x, c) = \sqrt{(\mathbf{f}(x) - \boldsymbol{\mu}_c)^\top \Sigma^{-1} (\mathbf{f}(x) - \boldsymbol{\mu}_c)}
\]
The confidence score $S(x)$ for an input sample is defined as the negative of the minimum Mahalanobis distance:
\[
S(x) = -\min_c D_\text{M}(x, c)
\]

To assess the effectiveness of different SSL methods for OOD detection, we evaluate the widely used AUROC and AUPR metrics, both of which are threshold-independent~\cite{li2023rethinkingoutofdistributionooddetection, hendrycks2019usingselfsupervisedlearningimprove, galil2023frameworkbenchmarkingclassoutofdistributiondetection}. The performance is assessed by comparing Mahalanobis-based scores derived from $\mathcal{P}_{\text{ID}}^{\text{test}}$ and $\mathcal{P}_{\text{OOD}}$. Following the framework of~\cite{lee2018simpleunifiedframeworkdetecting}, we validate the method across various datasets and configurations. Specifically, models trained on $\mathcal{P}_{\text{ID}}$ are evaluated against other 10 MedMNIST datasets, yielding $11 \times 10 \times 2 \ (\text{initializations}) \times 2 \ (\text{backbone types}) = 440$ OOD detection scores per each method, backbone, and initialization combination.

This comprehensive evaluation framework allows us to rigorously compare methods across diverse settings, highlighting the robustness and consistency of the Mahalanobis distance-based approach in handling OOD detection. If further clarity is needed on the methodology, readers are encouraged to refer to the foundational work in~\cite{lee2018simpleunifiedframeworkdetecting}.

\subsection{Additional Analysis}
\label{sec:appendix_ap}

\subsubsection{In-Domain Performance}
\label{sec:appendix_ap_idp_lat}

\paragraph{Linear Evaluation with All Labels}

\indent The in-domain performance of the self-supervised learning methods using ResNet-50 with random initialization is presented in Section~\ref{sec:experiments-rn50-benchmark} in Table~\ref{table:benchmark_resnet50_random_1}. Additional results for ResNet-50 with \imagenet initialization, ViT-Small with random initialization, and ViT-Small with \imagenet initialization are provided in Tables~\ref{table:benchmark_resnet50_imagenet_1},~\ref{table:benchmark_vit-small_random_1},~and~\ref{table:benchmark_vit-small_imagenet_1} respectively. Similar to Table~\ref{table:benchmark_resnet50_random_1}, Area Under the Curve (AUC) and accuracy (ACC) metrics are reported, and supervised learning results are provided for reference. The highest accuracy scores among the SSL methods are highlighted in \colorbox{green!30}{green}, and conversely, the lowest ones
are highlighted in \colorbox{red!30}{red}.

\begin{figure*}[h]
\begin{flushleft}
\setlength{\parindent}{10pt}
\textbf{Effect of Initialization with respect to Method and Dataset:} Figure~\ref{fig:init_comp_id_heatmap} illustrates the impact of random versus \imagenet initialization on the performance of various SSL methods across diverse datasets, evaluated on both ResNet-50 and ViT-Small backbones. \imagenet initialization generally enhances model accuracy, although the extent of improvement differs across methods and datasets. Some datasets receive greater gains in general such as BreastMNIST and OctMNIST. However, it can also be seen that TissueMNIST and OrganAMNIST prefer random initialization on ViT backbone most of the time as an exception. These insights suggest that selecting an appropriate pre-training method and initialization strategy is crucial for optimizing performance on particular tasks, especially in medical imaging domains where dataset characteristics vary widely. Nonetheless, working with \imagenet initialization is most of the time better.
\end{flushleft}
\vspace{2mm}
    \centering
    \thinspace
    \includegraphics[width=0.9\textwidth]{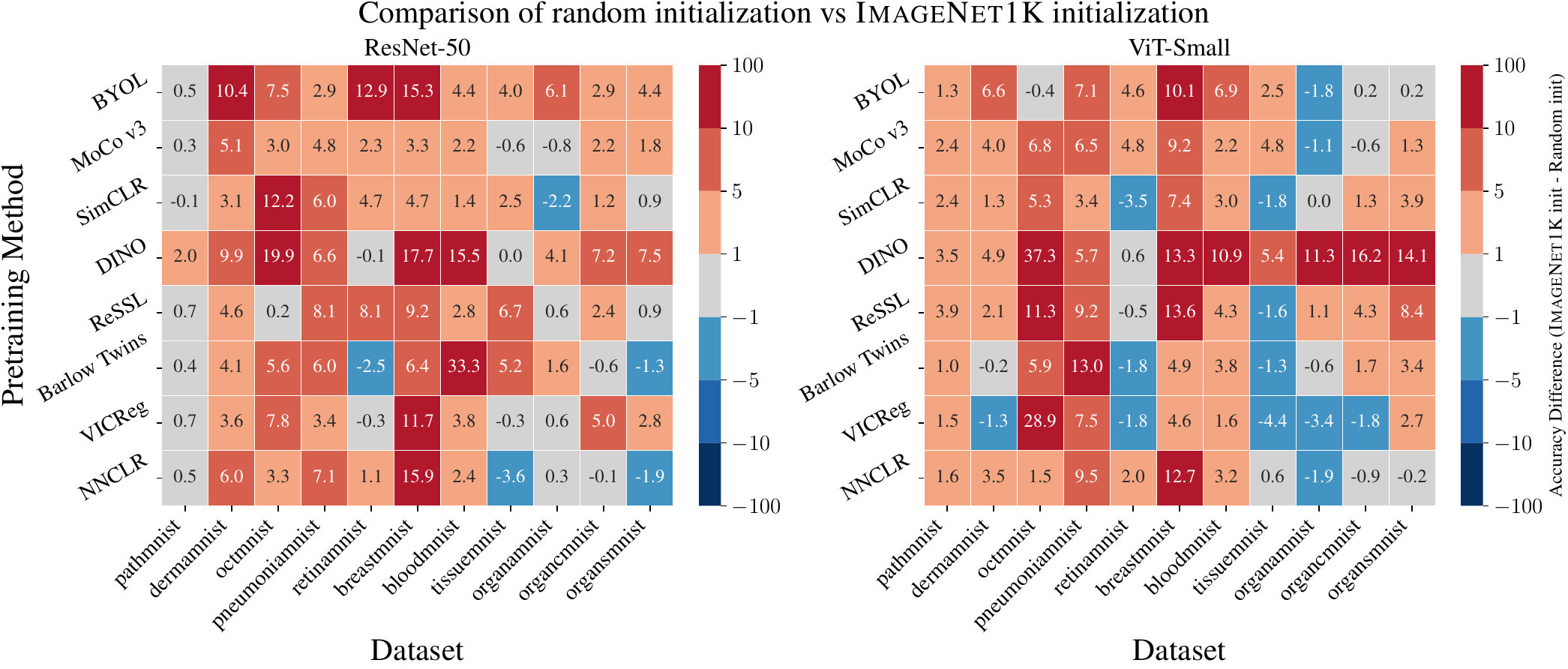}
    \captionsetup{justification=centering}
    \caption{
        Heatmap showing the accuracy differences for various methods on different datasets using ResNet-50 and ViT-Small backbones. Positive values (shades of red) indicate higher accuracy with \imagenet initialization compared to random, while negative values (shades of blue) show the opposite trend.
    }
    \label{fig:init_comp_id_heatmap}
\end{figure*}

These tables illustrate how initialization and architecture choices impact the performance of different SSL methods. Notably, DINO performs significantly better with \imagenet initialization on both ResNet-50 and ViT-Small, whereas it ranks among the worst with random initialization. Moreover, compared to the random initialization results for ResNet-50 in Table~\ref{table:benchmark_resnet50_random_2}, SimCLR achieves notably higher accuracy with ViT-Small. Further analyses on the impact of initialization and backbone architecture are provided in Tables ~\ref{tab:init_comparison_id}~and~\ref{tab:arch_comparison_id}, respectively. Additionally, Figure~\ref{fig:init_comp_id_heatmap} shows the performance differences between random and \imagenet initialization for each dataset and SSL method.

    \begin{table*}[t]
  \centering
    \caption{In-domain performance of SSL methods on a randomly initialized ResNet-50 backbone, reported as Area Under the Curve (AUC) and Accuracy (ACC). Supervised learning results with random and \imagenet initialization are included for comparison..}
    
    \renewcommand\thetable{3.1}
    \begin{adjustbox}{width=450pt,center}
\begin{tabular}{ccc|cc|cc|cc|ccc}
      \midrule
           & \multicolumn{2}{c}{PathMNIST} & \multicolumn{2}{c}{DermaMNIST} & \multicolumn{2}{c}{OCTMNIST} & \multicolumn{2}{c}{PneumoniaMNIST} & \multicolumn{2}{c}{RetinaMNIST} \\
      \midrule
          Method
          & AUC $\uparrow$ & ACC $\uparrow$
          & AUC $\uparrow$ & ACC $\uparrow$
          & AUC $\uparrow$ & ACC $\uparrow$
          & AUC $\uparrow$ & ACC $\uparrow$
          & AUC $\uparrow$ & ACC $\uparrow$ \\
      \midrule
          Supervised (random init.) & 98.38\scriptsize{$\pm$0.53} & 91.52\scriptsize{$\pm$1.03} & 90.79\scriptsize{$\pm$0.45} & 74.68\scriptsize{$\pm$0.89} & 95.78\scriptsize{$\pm$0.84} & 83.14\scriptsize{$\pm$1.34} & 93.48\scriptsize{$\pm$3.32} & 84.26\scriptsize{$\pm$2.54} & 71.78\scriptsize{$\pm$2.64} & 52.10\scriptsize{$\pm$3.03}  \\
          Supervised (\imagenet init.) & 99.46\scriptsize{$\pm$0.25} & 94.13\scriptsize{$\pm$0.43} & 94.41\scriptsize{$\pm$1.43} & 82.79\scriptsize{$\pm$0.81} & 96.93\scriptsize{$\pm$0.57} & 83.24\scriptsize{$\pm$1.59} & 94.84\scriptsize{$\pm$2.73} & 89.39\scriptsize{$\pm$0.88} & 73.86\scriptsize{$\pm$1.34} & 51.55\scriptsize{$\pm$4.68}  \\
      \midrule 
          SimCLR & 99.45\scriptsize{$\pm$0.01} & 92.91\scriptsize{$\pm$0.08} & 89.20\scriptsize{$\pm$0.32} & 74.22\scriptsize{$\pm$0.18} & 92.17\scriptsize{$\pm$0.27} & 68.82\scriptsize{$\pm$0.24} & 94.36\scriptsize{$\pm$2.54} & 88.81\scriptsize{$\pm$4.79} & 63.38\scriptsize{$\pm$6.30} & 46.45\scriptsize{$\pm$2.20}  \\
          DINO & 99.28\scriptsize{$\pm$0.01} & 92.03\scriptsize{$\pm$0.13} & 80.67\scriptsize{$\pm$0.49} & 68.77\scriptsize{$\pm$0.22} & 90.01\scriptsize{$\pm$0.05} & \cellcolor{red!30}{60.70}\scriptsize{$\pm$0.29} & 90.29\scriptsize{$\pm$3.01} & 87.24\scriptsize{$\pm$3.76} & 58.72\scriptsize{$\pm$1.87} & 46.20\scriptsize{$\pm$3.55}  \\
\hline
      \end{tabular}
    \end{adjustbox}
    \label{table:benchmark_resnet50_random_1}

    \begin{adjustbox}{width=450pt,center}
      \begin{tabular}{ccc|cc|cc|cc|cc|cc}
           & \multicolumn{2}{c}{BreastMNIST} & \multicolumn{2}{c}{BloodMNIST} & \multicolumn{2}{c}{TissueMNIST} & \multicolumn{2}{c}{OrganAMNIST} & \multicolumn{2}{c}{OrganCMNIST} & \multicolumn{2}{c}{OrganSMNIST} \\
      \midrule
          Method
          & AUC $\uparrow$ & ACC $\uparrow$
          & AUC $\uparrow$ & ACC $\uparrow$
          & AUC $\uparrow$ & ACC $\uparrow$
          & AUC $\uparrow$ & ACC $\uparrow$
          & AUC $\uparrow$ & ACC $\uparrow$
          & AUC $\uparrow$ & ACC $\uparrow$ \\
      \midrule
          Supervised (random init.) & 88.99\scriptsize{$\pm$2.46} & 83.71\scriptsize{$\pm$2.73} & 99.88\scriptsize{$\pm$0.02} & 97.83\scriptsize{$\pm$0.21} & 92.37\scriptsize{$\pm$0.05} & 69.95\scriptsize{$\pm$0.19} & 99.49\scriptsize{$\pm$0.11} & 92.32\scriptsize{$\pm$0.38} & 99.14\scriptsize{$\pm$0.09} & 90.11\scriptsize{$\pm$0.66} & 96.48\scriptsize{$\pm$0.30} & 78.14\scriptsize{$\pm$0.51}  \\
          Supervised (\imagenet init.) & 88.20\scriptsize{$\pm$4.92} & 85.64\scriptsize{$\pm$2.96} & 99.88\scriptsize{$\pm$0.04} & 98.34\scriptsize{$\pm$0.17} & 93.35\scriptsize{$\pm$0.12} & 71.72\scriptsize{$\pm$0.16} & 99.70\scriptsize{$\pm$0.09} & 95.19\scriptsize{$\pm$0.74} & 99.61\scriptsize{$\pm$0.07} & 94.22\scriptsize{$\pm$0.23} & 97.12\scriptsize{$\pm$0.32} & 82.52\scriptsize{$\pm$0.85}  \\
      \midrule
                    SimCLR & 85.56\scriptsize{$\pm$1.75} & 82.43\scriptsize{$\pm$2.36} & 99.80\scriptsize{$\pm$0.00} & \cellcolor{green!30}{96.51}\scriptsize{$\pm$0.12} & 87.38\scriptsize{$\pm$0.00} & 58.15\scriptsize{$\pm$0.07} & 97.92\scriptsize{$\pm$0.82} & 89.87\scriptsize{$\pm$2.27} & 99.29\scriptsize{$\pm$0.03} & \cellcolor{green!30}{90.22}\scriptsize{$\pm$0.24} & 97.35\scriptsize{$\pm$0.37} & 76.78\scriptsize{$\pm$0.24}  \\
\hline
      \end{tabular}
      \end{adjustbox}
    \label{table:benchmark_resnet50_random_2}
    
  \end{table*}
    \begin{table*}[h!]
  \centering
    \addtocounter{table}{-1}
    \captionlistentry{}  
   
    \begin{adjustbox}{width=500pt,center}
      \begin{tabular}{cccccccccccc}
       \toprule 
           & \multicolumn{2}{c}{PathMNIST} & \multicolumn{2}{c}{DermaMNIST} & \multicolumn{2}{c}{OCTMNIST} & \multicolumn{2}{c}{PneumoniaMNIST} & \multicolumn{2}{c}{RetinaMNIST} \\
      \midrule
          Method
          & AUC $\uparrow$ & ACC $\uparrow$
          & AUC $\uparrow$ & ACC $\uparrow$
          & AUC $\uparrow$ & ACC $\uparrow$
          & AUC $\uparrow$ & ACC $\uparrow$
          & AUC $\uparrow$ & ACC $\uparrow$ \\
      \midrule
          Supervised (random init.) & 98.38\scriptsize{$\pm$0.53} & 91.52\scriptsize{$\pm$1.03} & 90.79\scriptsize{$\pm$0.45} & 74.68\scriptsize{$\pm$0.89} & 95.78\scriptsize{$\pm$0.84} & 83.14\scriptsize{$\pm$1.34} & 93.48\scriptsize{$\pm$3.32} & 84.26\scriptsize{$\pm$2.54} & 71.78\scriptsize{$\pm$2.64} & 52.10\scriptsize{$\pm$3.03}  \\
          Supervised (\imagenet init.) & 99.46\scriptsize{$\pm$0.25} & 94.13\scriptsize{$\pm$0.43} & 94.41\scriptsize{$\pm$1.43} & 82.79\scriptsize{$\pm$0.81} & 96.93\scriptsize{$\pm$0.57} & 83.24\scriptsize{$\pm$1.59} & 94.84\scriptsize{$\pm$2.73} & 89.39\scriptsize{$\pm$0.88} & 73.86\scriptsize{$\pm$1.34} & 51.55\scriptsize{$\pm$4.68}  \\
      \midrule 
          SimCLR & 99.43\scriptsize{$\pm$0.05} & 92.81\scriptsize{$\pm$0.39} & 92.86\scriptsize{$\pm$0.13} & 77.28\scriptsize{$\pm$0.45} & 98.05\scriptsize{$\pm$0.32} & 80.98\scriptsize{$\pm$2.19} & 97.45\scriptsize{$\pm$0.38} & 94.80\scriptsize{$\pm$1.02} & 73.28\scriptsize{$\pm$2.08} & 51.15\scriptsize{$\pm$2.07}  \\
          DINO & 99.59\scriptsize{$\pm$0.02} & \cellcolor{green!30}{94.05}\scriptsize{$\pm$0.17} & 93.87\scriptsize{$\pm$0.60} & 78.63\scriptsize{$\pm$0.98} & 98.15\scriptsize{$\pm$0.15} & 80.64\scriptsize{$\pm$0.92} & 99.08\scriptsize{$\pm$0.17} & 93.84\scriptsize{$\pm$1.10} & 66.53\scriptsize{$\pm$1.58} & \cellcolor{red!30}{46.15}\scriptsize{$\pm$1.21}  \\
\hline
      \end{tabular}
    \end{adjustbox}
    \label{table:benchmark_resnet50_imagenet_1}
    
    \begin{adjustbox}{width=500pt,center}
      \begin{tabular}{ccccccccccccc}
           & \multicolumn{2}{c}{BreastMNIST} & \multicolumn{2}{c}{BloodMNIST} & \multicolumn{2}{c}{TissueMNIST} & \multicolumn{2}{c}{OrganAMNIST} & \multicolumn{2}{c}{OrganCMNIST} & \multicolumn{2}{c}{OrganSMNIST} \\
      \midrule
          Method
          & AUC $\uparrow$ & ACC $\uparrow$
          & AUC $\uparrow$ & ACC $\uparrow$
          & AUC $\uparrow$ & ACC $\uparrow$
          & AUC $\uparrow$ & ACC $\uparrow$
          & AUC $\uparrow$ & ACC $\uparrow$
          & AUC $\uparrow$ & ACC $\uparrow$ \\
      \midrule
          Supervised (random init.) & 88.99\scriptsize{$\pm$2.46} & 83.71\scriptsize{$\pm$2.73} & 99.88\scriptsize{$\pm$0.02} & 97.83\scriptsize{$\pm$0.21} & 92.37\scriptsize{$\pm$0.05} & 69.95\scriptsize{$\pm$0.19} & 99.49\scriptsize{$\pm$0.11} & 92.32\scriptsize{$\pm$0.38} & 99.14\scriptsize{$\pm$0.09} & 90.11\scriptsize{$\pm$0.66} & 96.48\scriptsize{$\pm$0.30} & 78.14\scriptsize{$\pm$0.51}  \\
          Supervised (\imagenet init.) & 88.20\scriptsize{$\pm$4.92} & 85.64\scriptsize{$\pm$2.96} & 99.88\scriptsize{$\pm$0.04} & 98.34\scriptsize{$\pm$0.17} & 93.35\scriptsize{$\pm$0.12} & 71.72\scriptsize{$\pm$0.16} & 99.70\scriptsize{$\pm$0.09} & 95.19\scriptsize{$\pm$0.74} & 99.61\scriptsize{$\pm$0.07} & 94.22\scriptsize{$\pm$0.23} & 97.12\scriptsize{$\pm$0.32} & 82.52\scriptsize{$\pm$0.85}  \\
      \midrule
          SimCLR & 91.90\scriptsize{$\pm$1.42} & 87.17\scriptsize{$\pm$0.56} & 99.86\scriptsize{$\pm$0.01} & 97.87\scriptsize{$\pm$0.18} & 89.19\scriptsize{$\pm$0.00} & \cellcolor{green!30}{60.64}\scriptsize{$\pm$0.05} & 98.64\scriptsize{$\pm$0.11} & \cellcolor{red!30}{87.69}\scriptsize{$\pm$1.22} & 99.46\scriptsize{$\pm$0.01} & 91.46\scriptsize{$\pm$0.09} & 97.18\scriptsize{$\pm$0.49} & 77.71\scriptsize{$\pm$0.42}  \\
\hline
      \end{tabular}
      \end{adjustbox}
          \caption{In-domain performance of the SSL methods using ResNet-50 with \imagenet initialization.}
    \label{table:benchmark_resnet50_imagenet_2}
  \end{table*}
    \begin{table*}[h]
  \centering
    \addtocounter{table}{-1}
    \captionlistentry{}
    \begin{adjustbox}{width=500pt,center}
      \begin{tabular}{cccccccccccc}
      \toprule
           & \multicolumn{2}{c}{PathMNIST} & \multicolumn{2}{c}{DermaMNIST} & \multicolumn{2}{c}{OCTMNIST} & \multicolumn{2}{c}{PneumoniaMNIST} & \multicolumn{2}{c}{RetinaMNIST} \\
      \midrule
          Method
          & AUC $\uparrow$ & ACC $\uparrow$
          & AUC $\uparrow$ & ACC $\uparrow$
          & AUC $\uparrow$ & ACC $\uparrow$
          & AUC $\uparrow$ & ACC $\uparrow$
          & AUC $\uparrow$ & ACC $\uparrow$ \\
      \midrule
        Supervised (random init.)  & 96.22\scriptsize{$\pm$0.04} & 79.57\scriptsize{$\pm$0.15} & 86.52\scriptsize{$\pm$0.58} & 73.99\scriptsize{$\pm$0.35} & 91.90\scriptsize{$\pm$0.29} & 70.04\scriptsize{$\pm$0.14} & 84.81\scriptsize{$\pm$3.03} & 83.71\scriptsize{$\pm$0.52} & 72.60\scriptsize{$\pm$0.74} & 51.95\scriptsize{$\pm$0.60}  \\
        Supervised (\imagenet init.) & 98.72\scriptsize{$\pm$0.15} & 92.53\scriptsize{$\pm$0.09} & 92.76\scriptsize{$\pm$0.20} & 80.49\scriptsize{$\pm$0.20} & 92.95\scriptsize{$\pm$1.35} & 80.16\scriptsize{$\pm$0.11} & 94.69\scriptsize{$\pm$1.69} & 86.69\scriptsize{$\pm$0.24} & 72.02\scriptsize{$\pm$0.09} & 51.05\scriptsize{$\pm$0.94}  \\
      \midrule 
          SimCLR & 99.06\scriptsize{$\pm$0.00} & 90.51\scriptsize{$\pm$0.04} & 93.09\scriptsize{$\pm$0.23} & \cellcolor{green!30}{77.51}\scriptsize{$\pm$0.44} & 95.21\scriptsize{$\pm$0.03} & 72.04\scriptsize{$\pm$0.42} & 97.07\scriptsize{$\pm$0.12} & \cellcolor{green!30}{89.87}\scriptsize{$\pm$1.20} & 72.46\scriptsize{$\pm$0.50} & \cellcolor{green!30}{53.00}\scriptsize{$\pm$0.79}  \\
          DINO & 99.16\scriptsize{$\pm$0.02} & 90.63\scriptsize{$\pm$0.13} & 90.44\scriptsize{$\pm$0.09} & \cellcolor{red!30}{73.92}\scriptsize{$\pm$0.36} & 75.14\scriptsize{$\pm$0.21} & \cellcolor{red!30}{41.94}\scriptsize{$\pm$0.66} & 93.38\scriptsize{$\pm$0.32} & 82.72\scriptsize{$\pm$1.32} & 68.42\scriptsize{$\pm$2.46} & 50.20\scriptsize{$\pm$1.79}  \\
\hline
      \end{tabular}
    \end{adjustbox}
    \label{table:benchmark_vit-small_random_1}

    \begin{adjustbox}{width=500pt,center}
      \begin{tabular}{ccccccccccccc}
           & \multicolumn{2}{c}{BreastMNIST} & \multicolumn{2}{c}{BloodMNIST} & \multicolumn{2}{c}{TissueMNIST} & \multicolumn{2}{c}{OrganAMNIST} & \multicolumn{2}{c}{OrganCMNIST} & \multicolumn{2}{c}{OrganSMNIST} \\
      \midrule
          Method
          & AUC $\uparrow$ & ACC $\uparrow$
          & AUC $\uparrow$ & ACC $\uparrow$
          & AUC $\uparrow$ & ACC $\uparrow$
          & AUC $\uparrow$ & ACC $\uparrow$
          & AUC $\uparrow$ & ACC $\uparrow$
          & AUC $\uparrow$ & ACC $\uparrow$ \\
      \midrule
          Supervised (random init.) & 75.68\scriptsize{$\pm$0.31} & 77.82\scriptsize{$\pm$0.71} & 98.90\scriptsize{$\pm$0.44} & 93.76\scriptsize{$\pm$0.15} & 87.98\scriptsize{$\pm$7.65} & 66.53\scriptsize{$\pm$16.46} & 98.87\scriptsize{$\pm$0.26} & 89.90\scriptsize{$\pm$0.85} & 97.89\scriptsize{$\pm$0.21} & 88.26\scriptsize{$\pm$0.09} & 93.29\scriptsize{$\pm$0.33} & 73.65\scriptsize{$\pm$0.08}  \\
          Supervised (\imagenet init.) & 85.69\scriptsize{$\pm$3.42} & 87.82\scriptsize{$\pm$1.26} & 99.72\scriptsize{$\pm$0.16} & 97.23\scriptsize{$\pm$0.46} & 87.01\scriptsize{$\pm$0.63} & 67.72\scriptsize{$\pm$0.04} & 99.52\scriptsize{$\pm$0.05} & 95.21\scriptsize{$\pm$0.05} & 99.62\scriptsize{$\pm$0.08} & 94.76\scriptsize{$\pm$0.05} & 95.09\scriptsize{$\pm$1.27} & 82.28\scriptsize{$\pm$0.24}  \\
      \midrule
          SimCLR & 80.17\scriptsize{$\pm$2.99} & 79.74\scriptsize{$\pm$1.83} & 99.66\scriptsize{$\pm$0.01} & 95.02\scriptsize{$\pm$0.21} & 90.66\scriptsize{$\pm$0.00} & \cellcolor{green!30}{63.44}\scriptsize{$\pm$0.06} & 99.58\scriptsize{$\pm$0.01} & \cellcolor{green!30}{92.61}\scriptsize{$\pm$0.16} & 99.07\scriptsize{$\pm$0.01} & 88.40\scriptsize{$\pm$0.18} & 96.83\scriptsize{$\pm$0.02} & 73.49\scriptsize{$\pm$0.08}  \\
\hline
      \end{tabular}
      \end{adjustbox}
          \caption{In-domain performance of the SSL methods using ViT-Small with random initialization.}
    \label{table:benchmark_vit-small_random_2}
  \end{table*}
    \begin{table*}[h]
  \centering
    \addtocounter{table}{-1}
    \captionlistentry{}
    \begin{adjustbox}{width=500pt,center}
      \begin{tabular}{cccccccccccc}
      \toprule
           & \multicolumn{2}{c}{PathMNIST} & \multicolumn{2}{c}{DermaMNIST} & \multicolumn{2}{c}{OCTMNIST} & \multicolumn{2}{c}{PneumoniaMNIST} & \multicolumn{2}{c}{RetinaMNIST} \\
      \midrule
          Method
          & AUC $\uparrow$ & ACC $\uparrow$
          & AUC $\uparrow$ & ACC $\uparrow$
          & AUC $\uparrow$ & ACC $\uparrow$
          & AUC $\uparrow$ & ACC $\uparrow$
          & AUC $\uparrow$ & ACC $\uparrow$ \\
      \midrule
        Supervised (random init.)  & 96.22\scriptsize{$\pm$0.04} & 79.57\scriptsize{$\pm$0.15} & 86.52\scriptsize{$\pm$0.58} & 73.99\scriptsize{$\pm$0.35} & 91.90\scriptsize{$\pm$0.29} & 70.04\scriptsize{$\pm$0.14} & 84.81\scriptsize{$\pm$3.03} & 83.71\scriptsize{$\pm$0.52} & 72.60\scriptsize{$\pm$0.74} & 51.95\scriptsize{$\pm$0.60}  \\
        Supervised (\imagenet init.) & 98.72\scriptsize{$\pm$0.15} & 92.53\scriptsize{$\pm$0.09} & 92.76\scriptsize{$\pm$0.20} & 80.49\scriptsize{$\pm$0.20} & 92.95\scriptsize{$\pm$1.35} & 80.16\scriptsize{$\pm$0.11} & 94.69\scriptsize{$\pm$1.69} & 86.69\scriptsize{$\pm$0.24} & 72.02\scriptsize{$\pm$0.09} & 51.05\scriptsize{$\pm$0.94}  \\
      \midrule 
          SimCLR & 99.33\scriptsize{$\pm$0.01} & 92.88\scriptsize{$\pm$0.11} & 94.36\scriptsize{$\pm$0.06} & 78.85\scriptsize{$\pm$0.08} & 97.41\scriptsize{$\pm$0.07} & 77.38\scriptsize{$\pm$0.57} & 96.54\scriptsize{$\pm$0.48} & 93.30\scriptsize{$\pm$1.71} & 72.71\scriptsize{$\pm$3.22} & 49.50\scriptsize{$\pm$4.36}  \\
          DINO & 99.40\scriptsize{$\pm$0.01} & 94.09\scriptsize{$\pm$0.06} & 94.39\scriptsize{$\pm$0.09} & 78.85\scriptsize{$\pm$0.72} & 97.24\scriptsize{$\pm$0.16} & \cellcolor{green!30}{79.28}\scriptsize{$\pm$1.03} & 97.71\scriptsize{$\pm$1.12} & \cellcolor{red!30}{88.39}\scriptsize{$\pm$2.39} & 74.65\scriptsize{$\pm$0.73} & 50.80\scriptsize{$\pm$1.75}  \\
\hline
      \end{tabular}
    \end{adjustbox}
    \label{table:benchmark_vit-small_imagenet_1}
  
    \begin{adjustbox}{width=500pt,center}
      \begin{tabular}{ccccccccccccc}
           & \multicolumn{2}{c}{BreastMNIST} & \multicolumn{2}{c}{BloodMNIST} & \multicolumn{2}{c}{TissueMNIST} & \multicolumn{2}{c}{OrganAMNIST} & \multicolumn{2}{c}{OrganCMNIST} & \multicolumn{2}{c}{OrganSMNIST} \\
      \midrule
          Method
          & AUC $\uparrow$ & ACC $\uparrow$
          & AUC $\uparrow$ & ACC $\uparrow$
          & AUC $\uparrow$ & ACC $\uparrow$
          & AUC $\uparrow$ & ACC $\uparrow$
          & AUC $\uparrow$ & ACC $\uparrow$
          & AUC $\uparrow$ & ACC $\uparrow$ \\
      \midrule
          Supervised (random init.) & 75.68\scriptsize{$\pm$0.31} & 77.82\scriptsize{$\pm$0.71} & 98.90\scriptsize{$\pm$0.44} & 93.76\scriptsize{$\pm$0.15} & 87.98\scriptsize{$\pm$7.65} & 66.53\scriptsize{$\pm$16.46} & 98.87\scriptsize{$\pm$0.26} & 89.90\scriptsize{$\pm$0.85} & 97.89\scriptsize{$\pm$0.21} & 88.26\scriptsize{$\pm$0.09} & 93.29\scriptsize{$\pm$0.33} & 73.65\scriptsize{$\pm$0.08}  \\
          Supervised (\imagenet init.) & 85.69\scriptsize{$\pm$3.42} & 87.82\scriptsize{$\pm$1.26} & 99.72\scriptsize{$\pm$0.16} & 97.23\scriptsize{$\pm$0.46} & 87.01\scriptsize{$\pm$0.63} & 67.72\scriptsize{$\pm$0.04} & 99.52\scriptsize{$\pm$0.05} & 95.21\scriptsize{$\pm$0.05} & 99.62\scriptsize{$\pm$0.08} & 94.76\scriptsize{$\pm$0.05} & 95.09\scriptsize{$\pm$1.27} & 82.28\scriptsize{$\pm$0.24}  \\
      \midrule
          SimCLR & 89.99\scriptsize{$\pm$1.51} & 87.17\scriptsize{$\pm$3.23} & 99.91\scriptsize{$\pm$0.00} & 98.01\scriptsize{$\pm$0.06} & 89.95\scriptsize{$\pm$0.01} & 61.69\scriptsize{$\pm$0.03} & 99.61\scriptsize{$\pm$0.01} & 92.62\scriptsize{$\pm$0.22} & 99.16\scriptsize{$\pm$0.13} & 89.73\scriptsize{$\pm$0.60} & 97.23\scriptsize{$\pm$0.11} & 77.34\scriptsize{$\pm$0.11}  \\
\hline
      \end{tabular}
      \end{adjustbox}
          \caption{In-domain performance of the SSL methods using ViT-Small with \imagenet initialization.}
    \label{table:benchmark_vit-small_imagenet_1}
  \end{table*}

    \begin{table}[h]
\begin{flushleft}
    \setlength{\parindent}{10pt} 
    \textbf{Random vs.\ \imagenet Initialization Performance Comparison:} We analyze the impact of initialization strategies on self-supervised learning methods across various medical datasets in Table ~\ref{tab:init_comparison_id}. The best scores in random initialization are \underline{underlined} and the best scores in \imagenet initialization are \textbf{bolded}.
MoCo v3 emerges as the best method under random initialization, achieving superior performance in 5 out of 11 datasets. 
The transition to \imagenet initialization shifts the performance landscape; DINO and BYOL show the most substantial improvements. DINO becomes the new leader, achieving the best accuracy in 3 out of 11 datasets.
The results show that while \imagenet initialization generally enhances performance, the choice of self-supervised method should be tailored to specific medical imaging tasks, considering both the initialization strategy and the target domain's characteristics.
\end{flushleft}
\vspace{0.2cm}
\centering
\definecolor{lightblue}{RGB}{230,240,255}
\setul{}{1.25pt}
\resizebox{\textwidth}{!}{
\begin{tabular}{l||c|>{}c||c|>{}c||c|>{}c||c|>{}c||c|>{}c||c|>{}c||c|>{}c||c|>{}c}
\toprule
Dataset & \multicolumn{2}{c||}{MoCo v3} & \multicolumn{2}{c||}{SimCLR} & \multicolumn{2}{c||}{DINO} & \multicolumn{2}{c||}{ReSSL} & \multicolumn{2}{c||}{BYOL} & \multicolumn{2}{c||}{VICReg} & \multicolumn{2}{c||}{NNCLR} & \multicolumn{2}{c}{Barlow Twins} \\
& Rand. & \imnet & Rand. & \imnet & Rand. & \imnet & Rand. & \imnet & Rand. & \imnet & Rand. & \imnet & Rand. &\imnet & Rand. & \imnet \\
\midrule
PathMNIST & 92.50 & \cellcolor{lightblue}92.84 & 92.92 & \cellcolor{lightblue}92.81 & 92.03 & \textbf{\cellcolor{lightblue}94.05} & 91.99 & \cellcolor{lightblue}92.69 & \ul{93.36} & \cellcolor{lightblue}93.85 & 92.32 & \cellcolor{lightblue}93.02 & 92.72 & \cellcolor{lightblue}93.18 & 92.43 & \cellcolor{lightblue}92.82 \\
BloodMNIST & 95.72 & \cellcolor{lightblue}97.90 &  \ul{96.52} & \cellcolor{lightblue}97.88 & 82.85 & \textbf{\cellcolor{lightblue}98.35} & 95.49 & \cellcolor{lightblue}98.28 & 93.72 & \cellcolor{lightblue}98.08 & 93.25 & \cellcolor{lightblue}97.06 & 95.95 & \cellcolor{lightblue}98.30 & 64.44 & \cellcolor{lightblue}97.72 \\
BreastMNIST &  \ul{85.26} & \cellcolor{lightblue}88.59 & 82.44 & \cellcolor{lightblue}87.18 & 70.38 & \cellcolor{lightblue}88.08 & 72.44 & \cellcolor{lightblue}81.67 & 73.21 & \cellcolor{lightblue}88.46 & 73.46 & \cellcolor{lightblue}85.13 & 73.08 & \textbf{\cellcolor{lightblue}88.97} & 79.62 & \cellcolor{lightblue}86.03 \\
DermaMNIST & 73.45 & \cellcolor{lightblue}78.50 & 74.22 & \cellcolor{lightblue}77.29 & 68.78 & \cellcolor{lightblue}78.63 &  \ul{74.75} & \textbf{\cellcolor{lightblue}79.32} & 68.60 & \cellcolor{lightblue}78.99 & 73.06 & \cellcolor{lightblue}76.67 & 72.68 & \cellcolor{lightblue}78.70 & 72.83 & \cellcolor{lightblue}76.90 \\
OCTMNIST & \ul{79.96} & \cellcolor{lightblue}82.94 & 68.82 & \cellcolor{lightblue}80.98 & 60.70 & \cellcolor{lightblue}80.64 & 78.56 & \cellcolor{lightblue}78.74 & 74.94 & \cellcolor{lightblue}82.44 & 75.20 & \textbf{\cellcolor{lightblue}83.00} & 79.16 & \cellcolor{lightblue}82.48 & 77.28 & \cellcolor{lightblue}82.86 \\
OrganAMNIST &  \ul{92.63} & \cellcolor{lightblue}91.85 & 89.87 & \cellcolor{lightblue}87.69 & 91.06 & \textbf{\cellcolor{lightblue}95.13} & 91.89 & \cellcolor{lightblue}92.47 & 86.58 & \cellcolor{lightblue}92.70 & 89.67 & \cellcolor{lightblue}90.25 & 92.56 & \cellcolor{lightblue}92.83 & 91.21 & \cellcolor{lightblue}92.80 \\
OrganCMNIST & 88.30 & \cellcolor{lightblue}90.47 & \ul{90.22} & \cellcolor{lightblue}91.47 & 84.27 & \cellcolor{lightblue}91.51 & 89.31 & \textbf{\cellcolor{lightblue}91.68} & 86.51 & \cellcolor{lightblue}89.38 & 84.88 & \cellcolor{lightblue}89.90 & 89.45 & \cellcolor{lightblue}89.38 & 89.29 & \cellcolor{lightblue}88.70 \\
OrganSMNIST & \ul{77.92} & \cellcolor{lightblue}79.69 & 76.78 & \cellcolor{lightblue}77.72 & 71.01 & \cellcolor{lightblue}78.48 & 77.20 & \cellcolor{lightblue}78.12 & 76.24 & \textbf{\cellcolor{lightblue}80.63} & 73.86 & \cellcolor{lightblue}76.69 & 76.65 & \cellcolor{lightblue}74.73 & 76.54 & \cellcolor{lightblue}75.20 \\
PneumoniaMNIST & 87.34 & \cellcolor{lightblue}92.18 & 88.81 & \cellcolor{lightblue}94.81 & 87.24 & \cellcolor{lightblue}93.85 & 87.21 & \cellcolor{lightblue}95.29 & 90.99 & \cellcolor{lightblue}93.88 & \ul{91.03} & \cellcolor{lightblue}94.42 & 88.24 & \textbf{\cellcolor{lightblue}95.35} & 88.08 & \cellcolor{lightblue}94.07 \\
RetinaMNIST & 47.75 & \cellcolor{lightblue}50.05 & 46.45 & \cellcolor{lightblue}51.15 & 46.20 & \cellcolor{lightblue}46.15 & 43.15 & \cellcolor{lightblue}51.30 & 41.20 & \textbf{\cellcolor{lightblue}54.10} & 49.15 & \cellcolor{lightblue}48.85 & 47.40 & \cellcolor{lightblue}48.45 & \ul{52.50} & \cellcolor{lightblue}49.95 \\
TissueMNIST & \ul{59.68} & \cellcolor{lightblue}59.09 & 58.15 & \textbf{\cellcolor{lightblue}60.65} & 59.39 & \cellcolor{lightblue}59.43 & 48.45 & \cellcolor{lightblue}55.18 & 55.59 & \cellcolor{lightblue}59.58 & 56.20 & \cellcolor{lightblue}55.92 & 59.39 & \cellcolor{lightblue}55.79 & 53.89 & \cellcolor{lightblue}59.09 \\
\bottomrule
\end{tabular}}
\captionsetup{justification=centering}
\caption{Comparison of mean accuracy scores between random initialization (Rand.) and \imagenet initialization (\imnet) for self-supervised training using ResNet-50 across different methods and datasets. }
\label{tab:init_comparison_id}
\end{table}
    \begin{table}[h]
\begin{flushleft}
    \setlength{\parindent}{10pt} 
    \textbf{ResNet-50 vs. ViT-S Performance Comparison:} To systematically assess the impact of architectural differences, we compare ResNet-50 and ViT-Small under both random and \imagenet initialization strategies. Table~\ref{tab:arch_comparison_id} presents the performance differential, computed as the accuracy of ResNet-50 minus that of ViT-Small. For visual clarity, we employ a color-coding scheme where \colorbox{red!10}{red} indicates superior ResNet-50 performance and \colorbox{blue!10}{blue} denotes better ViT-Small performance. Cells are left uncolored when the absolute performance difference is negligible ($< 0.01$ percentage points). \imagenet initialization generally reduces the performance gap between architectures, suggesting that pre-training helps mitigate architectural biases. Some datasets (e.g., OCT, OrganA) exhibit consistent architectural preferences regardless of initialization, while others (e.g., Blood, Breast) show initialization-dependent trends, where the preferred architecture shifts based on the initialization strategy. 
    
\end{flushleft}
\vspace{0.2cm}
\centering
\resizebox{\textwidth}{!}{
\begin{tabular}{l||c|c||c|c||c|c||c|c||c|c||c|c||c|c||c|c}
\toprule
Dataset & \multicolumn{2}{c||}{MoCo v3} & \multicolumn{2}{c||}{SimCLR} & \multicolumn{2}{c||}{DINO} & \multicolumn{2}{c||}{ReSSL} & \multicolumn{2}{c||}{BYOL} & \multicolumn{2}{c||}{VICReg} & \multicolumn{2}{c||}{NNCLR} & \multicolumn{2}{c}{Barlow Twins} \\
& Rand. & \imnet & Rand. & \imnet  & Rand. & \imnet  & Rand. & \imnet  & Rand. & \imnet  & Rand. & \imnet  & Rand. & \imnet  & Rand. & \imnet  \\
\midrule
PathMNIST & \cellcolor{red!10}0.79 & \cellcolor{blue!10}-1.31 & \cellcolor{red!10}2.40 & \cellcolor{blue!10}-0.07 & \cellcolor{red!10}1.40 & \cellcolor{blue!10}-0.04 & \cellcolor{red!10}3.33 & \cellcolor{red!10}0.17 & \cellcolor{red!10}2.43 & \cellcolor{red!10}1.66 & \cellcolor{red!10}2.17 & \cellcolor{red!10}1.41 & \cellcolor{red!10}1.56 & \cellcolor{red!10}0.44 & \cellcolor{red!10}1.81 & \cellcolor{red!10}1.19 \\
BloodMNIST & \cellcolor{blue!10}-0.20 & \cellcolor{blue!10}-0.26 & \cellcolor{red!10}1.49 & \cellcolor{blue!10}-0.14 & \cellcolor{blue!10}-4.27 & \cellcolor{red!10}0.32 & \cellcolor{red!10}1.93 & \cellcolor{red!10}0.43 & \cellcolor{red!10}2.18 & \cellcolor{blue!10}-0.32 & \cellcolor{blue!10}-1.56 & \cellcolor{red!10}0.65 & \cellcolor{red!10}1.02 & \cellcolor{red!10}0.19 & \cellcolor{blue!10}-29.24 & \cellcolor{red!10}0.20 \\
BreastMNIST & \cellcolor{red!10}7.82 & \cellcolor{red!10}1.92 & \cellcolor{red!10}2.69 & 0.00 & \cellcolor{blue!10}-2.69 & \cellcolor{red!10}1.67 & \cellcolor{red!10}0.64 & \cellcolor{blue!10}-3.72 & \cellcolor{red!10}0.13 & \cellcolor{red!10}5.26 & \cellcolor{blue!10}-9.10 & \cellcolor{blue!10}-2.05 & \cellcolor{blue!10}-2.44 & \cellcolor{red!10}0.77 & \cellcolor{red!10}2.95 & \cellcolor{red!10}4.49 \\
DermaMNIST & \cellcolor{blue!10}-2.29 & \cellcolor{blue!10}-1.21 & \cellcolor{blue!10}-3.29 & \cellcolor{blue!10}-1.57 & \cellcolor{blue!10}-5.15 & \cellcolor{blue!10}-0.22 & \cellcolor{blue!10}-2.35 & \cellcolor{red!10}0.11 & \cellcolor{blue!10}-5.42 & \cellcolor{blue!10}-1.60 & \cellcolor{blue!10}-3.38 & \cellcolor{red!10}1.54 & \cellcolor{blue!10}-2.33 & \cellcolor{red!10}0.22 & \cellcolor{blue!10}-4.45 & \cellcolor{blue!10}-0.16 \\
OCTMNIST & \cellcolor{red!10}8.58 & \cellcolor{red!10}4.72 & \cellcolor{blue!10}-3.22 & \cellcolor{red!10}3.60 & \cellcolor{red!10}18.76 & \cellcolor{red!10}1.36 & \cellcolor{red!10}15.08 & \cellcolor{red!10}3.96 & \cellcolor{red!10}4.28 & \cellcolor{red!10}12.20 & \cellcolor{red!10}27.96 & \cellcolor{red!10}6.82 & \cellcolor{red!10}4.98 & \cellcolor{red!10}6.76 & \cellcolor{red!10}7.06 & \cellcolor{red!10}6.74 \\
OrganAMNIST & \cellcolor{red!10}1.80 & \cellcolor{red!10}2.14 & \cellcolor{blue!10}-2.74 & \cellcolor{blue!10}-4.93 & \cellcolor{red!10}9.69 & \cellcolor{red!10}2.50 & \cellcolor{red!10}2.24 & \cellcolor{red!10}1.70 & \cellcolor{blue!10}-3.72 & \cellcolor{red!10}4.23 & \cellcolor{blue!10}-1.09 & \cellcolor{red!10}2.85 & \cellcolor{red!10}1.05 & \cellcolor{red!10}3.26 & \cellcolor{red!10}1.96 & \cellcolor{red!10}4.11 \\
OrganCMNIST & \cellcolor{blue!10}-0.61 & \cellcolor{red!10}2.20 & \cellcolor{red!10}1.82 & \cellcolor{red!10}1.74 & \cellcolor{red!10}8.67 & \cellcolor{blue!10}-0.34 & \cellcolor{red!10}4.62 & \cellcolor{red!10}2.67 & \cellcolor{blue!10}-1.03 & \cellcolor{red!10}1.67 & \cellcolor{blue!10}-1.49 & \cellcolor{red!10}5.36 & \cellcolor{red!10}1.44 & \cellcolor{red!10}2.22 & \cellcolor{red!10}2.75 & \cellcolor{red!10}0.50 \\
OrganSMNIST & \cellcolor{red!10}2.97 & \cellcolor{red!10}3.42 & \cellcolor{red!10}3.29 & \cellcolor{red!10}0.37 & \cellcolor{red!10}7.02 & \cellcolor{red!10}0.40 & \cellcolor{red!10}8.18 & \cellcolor{red!10}0.73 & \cellcolor{red!10}1.98 & \cellcolor{red!10}6.21 & \cellcolor{red!10}2.88 & \cellcolor{red!10}3.00 & \cellcolor{red!10}1.29 & \cellcolor{blue!10}-0.40 & \cellcolor{red!10}4.10 & \cellcolor{blue!10}-0.66 \\
PneumoniaMNIST & 0.00 & \cellcolor{blue!10}-1.70 & \cellcolor{blue!10}-1.06 & \cellcolor{red!10}1.51 & \cellcolor{red!10}4.52 & \cellcolor{red!10}5.45 & \cellcolor{red!10}4.10 & \cellcolor{red!10}2.98 & \cellcolor{red!10}4.62 & \cellcolor{red!10}0.35 & \cellcolor{red!10}4.36 & \cellcolor{red!10}0.26 & \cellcolor{red!10}3.56 & \cellcolor{red!10}1.19 & \cellcolor{red!10}6.89 & \cellcolor{blue!10}-0.13 \\
RetinaMNIST & \cellcolor{red!10}3.60 & \cellcolor{red!10}1.10 & \cellcolor{blue!10}-6.55 & \cellcolor{red!10}1.65 & \cellcolor{blue!10}-4.00 & \cellcolor{blue!10}-4.65 & \cellcolor{blue!10}-7.90 & \cellcolor{red!10}0.75 & \cellcolor{blue!10}-5.45 & \cellcolor{red!10}2.80 & \cellcolor{red!10}0.65 & \cellcolor{red!10}2.15 & \cellcolor{blue!10}-0.65 & \cellcolor{red!10}36.95 & \cellcolor{red!10}2.45 & \cellcolor{red!10}1.65 \\
TissueMNIST & \cellcolor{red!10}3.94 & \cellcolor{blue!10}-1.48 & \cellcolor{blue!10}-5.29 & \cellcolor{blue!10}-1.04 & \cellcolor{red!10}0.97 & \cellcolor{blue!10}-4.38 & \cellcolor{blue!10}-12.53 & \cellcolor{blue!10}-4.20 & \cellcolor{blue!10}-0.74 & \cellcolor{red!10}0.71 & \cellcolor{blue!10}-5.46 & \cellcolor{blue!10}-1.32 & \cellcolor{blue!10}-0.77 & \cellcolor{blue!10}-4.95 & \cellcolor{blue!10}-7.59 & \cellcolor{blue!10}-1.12 \\
\bottomrule
\end{tabular}}
\captionsetup{justification=centering}
\caption{Performance difference between ResNet-50 and ViT across different initialization strategies and methods. Red indicates better ResNet-50 performance, and blue indicates better ViT performance.}
\label{tab:arch_comparison_id}
\end{table}

\begin{table}
\begin{flushleft}
\setlength{\parindent}{10pt} 
\textbf{Self-Supervised \imagenet Weights vs. Supervised \imagenet Weights:} Supervised \imagenet initialization outperformed self-supervised in 24/44 (54.5\%) dataset-method combinations, while self-supervised performed better in 20/44 (45.5\%). The difference between methods ranged from -9.25\% (favoring self-supervised) to +12.35\% (favoring supervised), with a median absolute difference of 1.52\%. A paired t-test across all dataset-method combinations confirmed no statistically significant difference between supervised and self-supervised \imagenet initialization for in-domain performance (t(43) = 1.040, p = 0.3043, 95\% CI: [-0.52\%, 1.64\%]), though both initialization techniques substantially outperformed random initialization.
\end{flushleft}

\caption{Comparison between supervised and self-supervised \imagenet initialization across different methods and datasets. Best accuracies for each method are \underline{underlined}.}
\centering
\definecolor{lightblue}{RGB}{230,240,255}
\setul{}{1.25pt}
\resizebox{0.45\textwidth}{!}{
\begin{tabular}{l||c|c||c|c||c|c||c|c}
\toprule
\multirow{2}{*}{Dataset} & \multicolumn{2}{c||}{MoCo v3} & \multicolumn{2}{c||}{SimCLR} & \multicolumn{2}{c||}{DINO} & \multicolumn{2}{c}{BYOL} \\ \cline{2-9} 
\\[-0.99em]
& Sup. & SSL & Sup. & SSL & Sup. & SSL & Sup. & SSL \\
\midrule
Path & 92.84 & \cellcolor{lightblue}93.52 & 92.81 & \cellcolor{lightblue}92.52 & \ul{94.05} & \cellcolor{lightblue}93.85 & 93.85 & \cellcolor{lightblue}93.43 \\
Derma & 78.50 & \cellcolor{lightblue}77.29 & 77.29 & \cellcolor{lightblue}76.66 & 78.63 & \cellcolor{lightblue}\ul{80.21} & 78.99 & \cellcolor{lightblue}77.12 \\
OCT & \ul{82.94} & \cellcolor{lightblue}81.48 & 80.98 & \cellcolor{lightblue}75.46 & 80.64 & \cellcolor{lightblue}80.10 & 82.44 & \cellcolor{lightblue}80.56 \\
Pneumonia & 92.18 & \cellcolor{lightblue}94.36 & \ul{94.81} & \cellcolor{lightblue}93.78 & 93.85 & \cellcolor{lightblue}92.12 & 93.88 & \cellcolor{lightblue}91.47 \\
Retina & 50.05 & \cellcolor{lightblue}51.40 & 51.15 & \cellcolor{lightblue}52.35 & 46.15 & \cellcolor{lightblue}\ul{55.40} & 54.10 & \cellcolor{lightblue}41.75 \\
Breast & 88.59 & \cellcolor{lightblue}84.62 & 87.18 & \cellcolor{lightblue}\ul{90.00} & 88.08 & \cellcolor{lightblue}89.49 & 88.46 & \cellcolor{lightblue}85.77 \\
Blood & 97.90 & \cellcolor{lightblue}98.12 & 97.88 & \cellcolor{lightblue}97.99 & \ul{98.35} & \cellcolor{lightblue}98.22 & 98.08 & \cellcolor{lightblue}98.29 \\
Tissue & 59.09 & \cellcolor{lightblue}\ul{63.27} & 60.65 & \cellcolor{lightblue}62.90 & 59.43 & \cellcolor{lightblue}47.21 & 59.58 & \cellcolor{lightblue}61.23 \\
OrganA & 91.85 & \cellcolor{lightblue}91.70 & 87.69 & \cellcolor{lightblue}91.02 & \ul{95.13} & \cellcolor{lightblue}93.43 & 92.70 & \cellcolor{lightblue}91.58 \\
OrganC & 90.47 & \cellcolor{lightblue}91.26 & 91.47 & \cellcolor{lightblue}89.58 & 91.51 & \cellcolor{lightblue}\ul{91.81} & 89.38 & \cellcolor{lightblue}91.01 \\
OrganS & 79.69 & \cellcolor{lightblue}77.39 & 77.72 & \cellcolor{lightblue}78.91 & 78.48 & \cellcolor{lightblue}79.75 & \ul{80.63} & \cellcolor{lightblue}76.18 \\
\bottomrule
\end{tabular}}
\label{tab:selfsup_vs_sup_in1k}
\end{table}

    \newpage
    \phantom{.}
    \newpage
    \phantom{.}
    \newpage
    
    \paragraph{Linear Evaluation with Limited Labels} The in-domain performance results for various label availability scenarios are presented in Tables~\ref{table:scarcity_resnet50_random_2},~\ref{table:scarcity_resnet50_imagenet_2},~\ref{table:scarcity_vit_random_2},~and~\ref{table:scarcity_vit_imagenet_2}; corresponding to ResNet-50 and ViT-Small architectures with random or \imagenet initialization. As discussed in Section~\ref{sec:experiments-label-availability}, only 1\% or 10\% of the labels from each dataset were used during downstream training in these scenarios. Notably, all backbones were pre-trained with 100\% of their respective datasets. 
    The highest and lowest accuracy values are again highlighted in \colorbox{green!30}{green} and \colorbox{red!30}{red}, respectively.

\begin{figure}[h]
\raggedright
\setlength{\parindent}{10pt} 
\subparagraph*{Comparison of Fine-Tuning Strategies with Limited Labeled Data:} Figure~\ref{fig:scarcity_dino_finetune_comparison} presents the experiments using DINO to assess whether fine-tuning the encoder alongside the linear classifier in a low-shot setting improves performance compared to training only the linear classifier. Specifically, we perform two experiments:

\begin{itemize}
\item \textbf{Frozen Backbone:} We first train a ResNet-50 backbone, initialized with \imagenet weights, on 100\% of the training data in a self-supervised manner using DINO. Next, we train a linear classifier on top of the frozen backbone using only 1\% of the labeled data.
\item \textbf{Fine-Tuned Backbone:} Using the same pre-trained ResNet-50 backbone, we fine-tune both the backbone and the linear classifier with 1\% of the labeled training data.

\end{itemize}

Our results align with the findings of \citet{caron2021dino}, demonstrating that the representations learned by DINO are not only robust but often surpass fine-tuned models in terms of accuracy. Interestingly, we observe that fine-tuned models exhibited higher variance across downstream classification tasks. This increased variability is likely attributed to the continuous parameter updates during gradient-based fine-tuning, which can make the model more sensitive to the limited labeled data available in the low-shot setting. 


\vspace{0.3cm}
\centering
\includegraphics[width=0.9\textwidth]{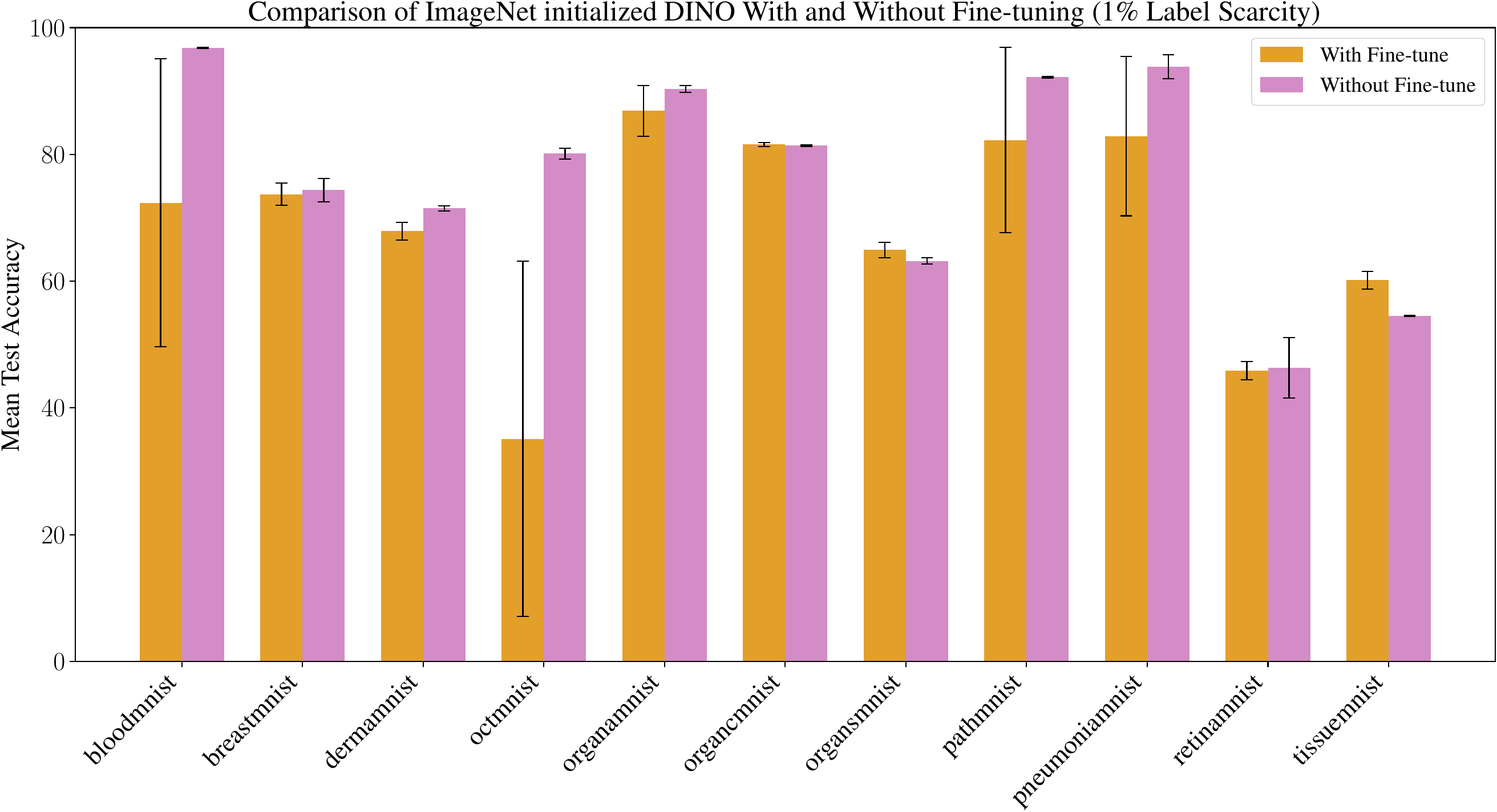}
\captionsetup{justification=centering}
        \caption{
In-domain performance of DINO with ResNet-50 backbone and \imagenet initialization, with and without fine-tuning, across datasets.
}

\label{fig:scarcity_dino_finetune_comparison}
\end{figure}

\begin{figure*}[h]
    \centering
    \begin{minipage}{1\textwidth}
    \setlength{\parindent}{10pt} 
    \subparagraph*{Analysis of Performance Drop due to Label Scarcity Across Methods and Initializations:} Figure \ref{fig:scarcity_test_acc_drop_combined_scatter} depicts the percentage performance drops of various self-supervised learning methods across different datasets, comparing ResNet-50 and ViT-Small backbones as well as \imagenet and random initialization. Each data point represents the accuracy drop for a specific dataset and method, with larger values indicating greater performance drops. Dashed lines mark the average performance drop for each backbone or initialization setting, summarizing overall trends.

    \indent In the top two plots, comparing ResNet-50 with ViT-Small across both \imagenet and random initializations, ResNet-50 consistently exhibits lower average performance drops than ViT-Small across methods and datasets. This suggests that ResNet-50 offers more stable accuracy across a range of medical imaging tasks under these conditions.

    \indent Conversely, the bottom two plots comparing \imagenet with random initialization for different backbones show less consistency, with average performance drop lines frequently crossing. This indicates no definitive advantage of one initialization strategy over the other across all datasets and methods, suggesting that the impact of initialization is more context-dependent, varying significantly with the specific dataset or backbone used. \\
    
    \end{minipage}\hfill
    \begin{minipage}{1\textwidth}
        \centering
        \includegraphics[width=\textwidth]{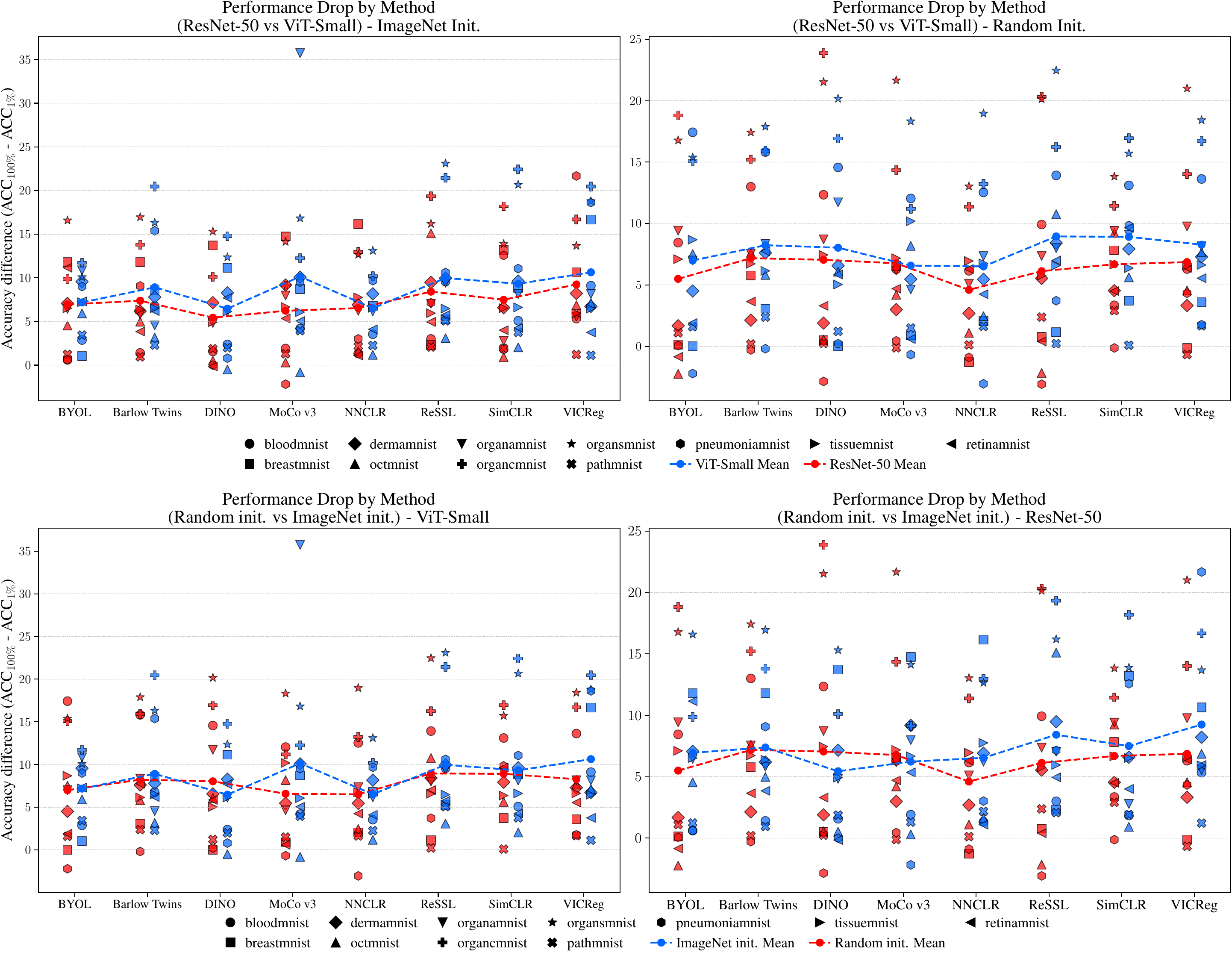}
        \captionsetup{justification=centering}
        \caption{
        Scatter plots showing the performance drop (\%) for different SSL methods across various datasets. The top row compares ResNet-50 and ViT-Small backbones, while the bottom row contrasts \imagenet and random initialization strategies. Each data point represents the performance drop of a specific method on a particular dataset, with dashed lines indicating the average performance drop for each comparison.
        }
        \label{fig:scarcity_test_acc_drop_combined_scatter}
    \end{minipage}
\end{figure*}

\begin{figure}[h]
    \centering
    \begin{minipage}{1\textwidth}
       
        \subparagraph*{Performance Comparison of SSL Methods Under Label Scarcity:}
        \setlength{\parindent}{10pt}
         
        Figure \ref{fig:scarcity_test_acc_grouped_by_method} shows the mean test accuracy across all datasets using various self-supervised learning methods at different levels of label availability (1\%, 10\%, and 100\%). This visualization highlights how accuracy scales with the availability of labeled data for each method, encompassing all combinations of backbones (ResNet-50 and ViT) and initialization types (Random Initialization and \imagenet Initialization).

        \indent The results show that at 1\% label availability, there is significantly higher variance in test accuracy compared to 10\% and 100\%, indicating greater instability with limited labels.
        As label availability increases, the mean accuracy naturally improves, demonstrating enhanced performance with more labeled data. Furthermore, the increase in mean accuracy from 10\% to 100\% label availability seems to be smaller than the increase from 1\% to 10\%, suggesting that additional label access yields diminishing returns in accuracy gains. 
        \vspace{0.5em}
    \end{minipage}\hfill
    \begin{minipage}{1\textwidth}
        \centering
        \includegraphics[width=0.5\textwidth]{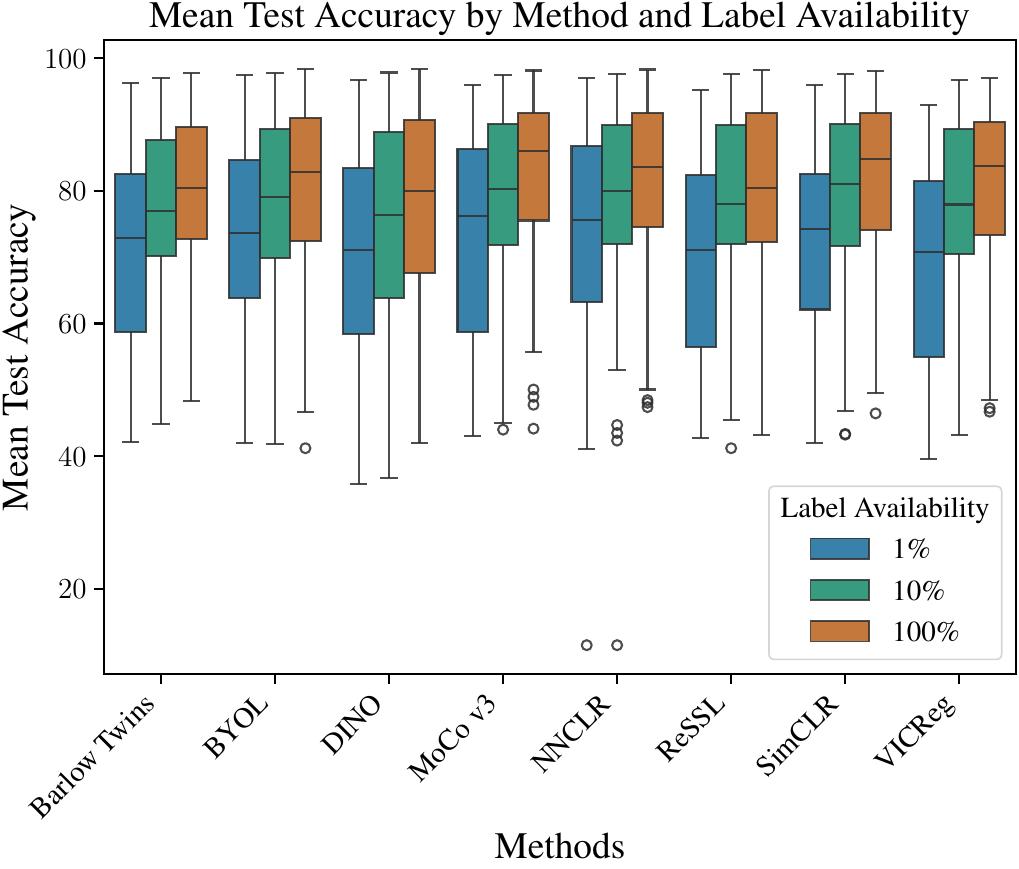}
        \captionsetup{justification=centering}
        \caption{
            Mean test accuracy across different self-supervised learning methods at varying levels of label availability (1\%, 10\%, and 100\%). The plot illustrates how accuracy scales with label availability for each method. All combinations of backbones (ResNet50, ViT) and initialization types (Random Initialization, \imagenet Initialization) were used to calculate the means.
        }
        \label{fig:scarcity_test_acc_grouped_by_method}
    \end{minipage}
\end{figure}

\begin{figure}[h]
    \centering
    \begin{minipage}{\textwidth}
       
        \subparagraph*{Comparison of \imagenet vs Random Initialization using ViT-Small in Low-Label Setting:}
         \setlength{\parindent}{10pt} Figure ~\ref{fig:scarcity_vitsmall_initcomparison} illustrates the mean test accuracy of ViT-Small across datasets under 1\%, 10\%, and 100\% label availability, comparing models initialized with \imagenet weights to those with random initialization (hatched bars).
       
        
        \indent As is evident from the plot, models initialized with \imagenet weights consistently outperform those with random initialization, particularly in label-scarce scenarios. Notably, DINO exhibits significant performance gains at 1\% label availability, transitioning from the worst-performing method under random initialization to the best-performing one when initialized with \imagenet weights. Furthermore, the increase across all methods in mean test accuracy from 10\% to 100\% label availability seems smaller than the increase from 1\% to 10\%, suggesting that additional label access yields diminishing returns in accuracy gains.
        
    \end{minipage}\hfill
    \vspace{2mm}
    \begin{minipage}{1\textwidth}
        \centering
        \includegraphics[width=1\textwidth]{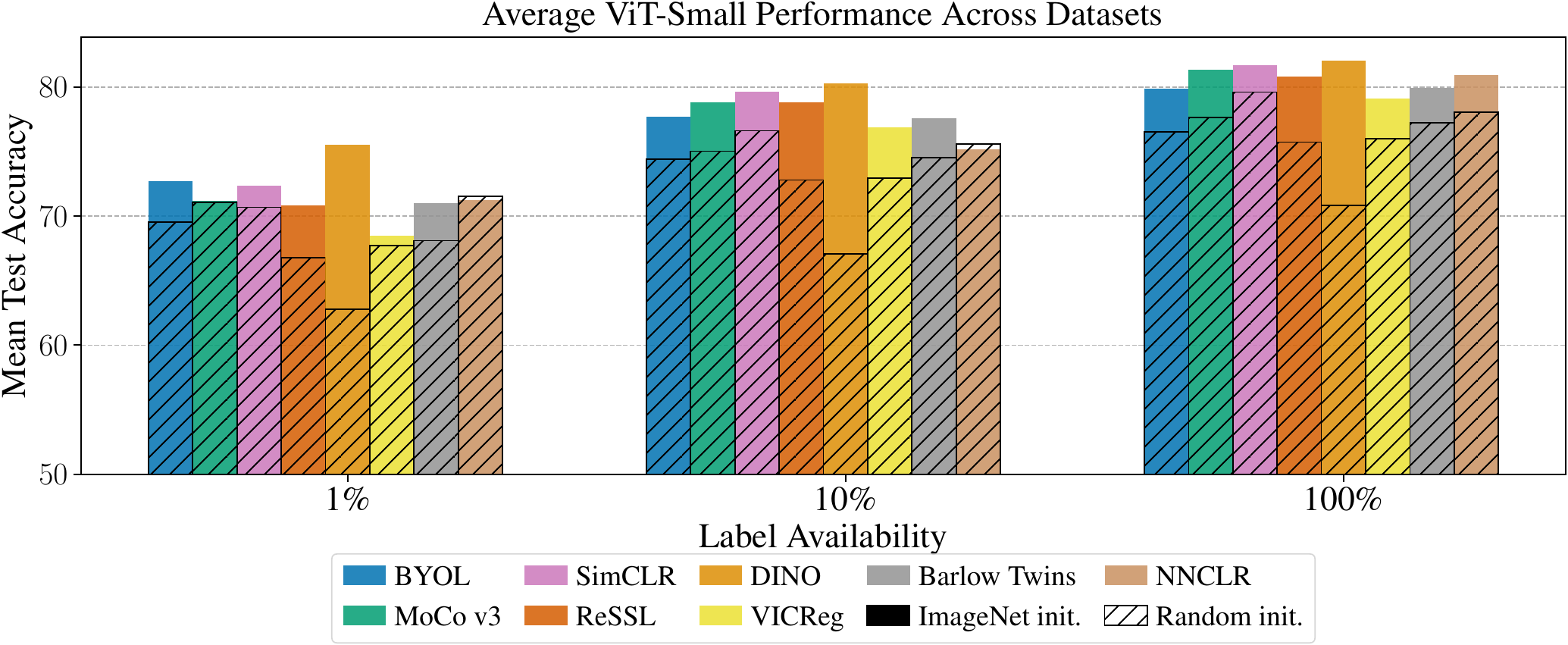}
        \captionsetup{justification=centering}
        \caption{
            Mean test accuracy across all datasets with 1\%, 10\%, and 100\% label availability, comparing \imagenet against random initialization (hatched) using ViT-Small for different methods.
        }
        \label{fig:scarcity_vitsmall_initcomparison}
    \end{minipage}
\end{figure}

    The aforementioned tables illustrate how label scarcity impacts the performance of different methods. DINO notably shows significant robustness to label scarcity compared to other methods when tested on \imagenet-initialized backbones, prevailing as the best performing method on several datasets in both label scarcity scenarios and with both architectures. For further inspection, DINO's performance is further evaluated with a fine-tuned backbone as presented in Figure~\ref{fig:scarcity_dino_finetune_comparison}.
    On the other hand, detailed analysis of the effects of label scarcity across different SSL methods, architectures, and initialization strategies is provided in Figure~\ref{fig:scarcity_test_acc_drop_combined_scatter}. Moreover, Figure~\ref{fig:scarcity_test_acc_grouped_by_method} presents details on individual methods' robustness to label scarcity and Figure~\ref{fig:scarcity_vitsmall_initcomparison} includes additional analysis of label scarcity on ViT-Small.

    \begin{table*}[h!]
  \centering
    \addtocounter{table}{-1}
    \captionlistentry{}  
    \begin{adjustbox}{width=500pt,center}
\begin{tabular}{ccc|cc|cc|cc|cc}
          \toprule
           & \multicolumn{2}{c}{PathMNIST} & \multicolumn{2}{c}{DermaMNIST} & \multicolumn{2}{c}{OCTMNIST} & \multicolumn{2}{c}{PneumoniaMNIST} & \multicolumn{2}{c}{RetinaMNIST} \\
      \midrule
          Method & 10\% & 1\% & 10\% & 1\% & 10\% & 1\% & 10\% & 1\% & 10\% & 1\% \\
      \midrule
          SimCLR & 92.13\scriptsize{ $\pm$ 0.51} & 90.00\scriptsize{ $\pm$ 0.50} & 71.71\scriptsize{ $\pm$ 0.48} & 69.68\scriptsize{ $\pm$ 0.82} & 67.64\scriptsize{ $\pm$ 0.78} & \cellcolor{red!30}{59.58}\scriptsize{ $\pm$ 1.84} & 91.41\scriptsize{ $\pm$ 0.57} & 88.94\scriptsize{ $\pm$ 3.84} & 43.35\scriptsize{ $\pm$ 2.18} & 41.95\scriptsize{ $\pm$ 3.36}  \\
          DINO & 91.86\scriptsize{ $\pm$ 0.05} & 91.79\scriptsize{ $\pm$ 0.29} & 68.05\scriptsize{ $\pm$ 0.80} & \cellcolor{red!30}{66.88}\scriptsize{ $\pm$ 0.65} & \cellcolor{red!30}{58.46}\scriptsize{ $\pm$ 0.38} & 60.14\scriptsize{ $\pm$ 1.61} & 91.63\scriptsize{ $\pm$ 1.04} & 90.09\scriptsize{ $\pm$ 0.79} & 45.70\scriptsize{ $\pm$ 2.12} & 42.90\scriptsize{ $\pm$ 1.67}  \\
          BYOL & 92.28\scriptsize{ $\pm$ 0.12} & 92.26\scriptsize{ $\pm$ 0.08} & \cellcolor{red!30}{67.45}\scriptsize{ $\pm$ 0.60} & 66.92\scriptsize{ $\pm$ 0.57} & 77.60\scriptsize{ $\pm$ 0.72} & 77.18\scriptsize{ $\pm$ 1.14} & 91.79\scriptsize{ $\pm$ 0.70} & \cellcolor{green!30}{90.92}\scriptsize{ $\pm$ 0.62} & 41.90\scriptsize{ $\pm$ 2.68} & 42.05\scriptsize{ $\pm$ 2.30}  \\
          ReSSL & \cellcolor{red!30}{91.40}\scriptsize{ $\pm$ 0.22} & \cellcolor{red!30}{89.60}\scriptsize{ $\pm$ 0.36} & 71.19\scriptsize{ $\pm$ 0.46} & 69.21\scriptsize{ $\pm$ 0.59} & 78.26\scriptsize{ $\pm$ 1.98} & \cellcolor{green!30}{80.72}\scriptsize{ $\pm$ 1.37} & 90.44\scriptsize{ $\pm$ 1.26} & 90.28\scriptsize{ $\pm$ 1.12} & \cellcolor{red!30}{41.20}\scriptsize{ $\pm$ 1.15} & 42.75\scriptsize{ $\pm$ 2.08}  \\
\hline
  \end{tabular}
    \end{adjustbox}
  \label{table:scarcity_resnet50_random_1}
  
    \begin{adjustbox}{width=500pt,center}
      \begin{tabular}{ccc|cc|cc|cc|cc|cc}
           & \multicolumn{2}{c}{BreastMNIST} & \multicolumn{2}{c}{BloodMNIST} & \multicolumn{2}{c}{TissueMNIST} & \multicolumn{2}{c}{OrganAMNIST} & \multicolumn{2}{c}{OrganCMNIST} & \multicolumn{2}{c}{OrganSMNIST} \\
      \midrule
          Method & 10\% & 1\% & 10\% & 1\% & 10\% & 1\% & 10\% & 1\% & 10\% & 1\% & 10\% & 1\% \\
      \midrule
          SimCLR & 77.30\scriptsize{ $\pm$ 1.65} & 74.61\scriptsize{ $\pm$ 1.45} & \cellcolor{green!30}{95.35}\scriptsize{ $\pm$ 0.23} & \cellcolor{green!30}{93.17}\scriptsize{ $\pm$ 0.54} & 55.94\scriptsize{ $\pm$ 0.05} & 51.49\scriptsize{ $\pm$ 0.13} & 85.92\scriptsize{ $\pm$ 1.58} & 80.47\scriptsize{ $\pm$ 0.84} & 85.13\scriptsize{ $\pm$ 0.53} & \cellcolor{green!30}{78.77}\scriptsize{ $\pm$ 0.82} & 72.45\scriptsize{ $\pm$ 0.47} & 62.95\scriptsize{ $\pm$ 0.74}  \\
          DINO & \cellcolor{red!30}{71.02}\scriptsize{ $\pm$ 2.06} & \cellcolor{red!30}{69.87}\scriptsize{ $\pm$ 2.52} & 77.43\scriptsize{ $\pm$ 0.23} & 70.51\scriptsize{ $\pm$ 0.68} & 56.60\scriptsize{ $\pm$ 0.09} & 51.94\scriptsize{ $\pm$ 0.21} & 87.08\scriptsize{ $\pm$ 1.39} & 82.34\scriptsize{ $\pm$ 0.63} & \cellcolor{red!30}{75.49}\scriptsize{ $\pm$ 0.38} & \cellcolor{red!30}{60.39}\scriptsize{ $\pm$ 1.88} & \cellcolor{red!30}{64.26}\scriptsize{ $\pm$ 0.54} & \cellcolor{red!30}{49.48}\scriptsize{ $\pm$ 1.60}  \\
          BYOL & 73.33\scriptsize{ $\pm$ 0.71} & 73.07\scriptsize{ $\pm$ 0.00} & 90.79\scriptsize{ $\pm$ 0.09} & 85.27\scriptsize{ $\pm$ 0.21} & 53.26\scriptsize{ $\pm$ 0.13} & 48.48\scriptsize{ $\pm$ 0.55} & \cellcolor{red!30}{83.82}\scriptsize{ $\pm$ 0.80} & \cellcolor{red!30}{77.16}\scriptsize{ $\pm$ 0.45} & 79.73\scriptsize{ $\pm$ 0.48} & 67.70\scriptsize{ $\pm$ 0.90} & 71.51\scriptsize{ $\pm$ 0.45} & 59.46\scriptsize{ $\pm$ 0.46}  \\
\hline
  \end{tabular}
    \end{adjustbox}
    \caption{In-domain performance of the self-supervised learning methods using ResNet-50 with random initialization with 1\% and 10\% of the labels.}
  \label{table:scarcity_resnet50_random_2}
\end{table*}

    \begin{table*}[h!]
  \centering
    \addtocounter{table}{-1}
    \captionlistentry{}  
    \begin{adjustbox}{width=500pt,center}
      \begin{tabular}{ccc|cc|cc|cc|cc}
      \toprule
           & \multicolumn{2}{c}{PathMNIST} & \multicolumn{2}{c}{DermaMNIST} & \multicolumn{2}{c}{OCTMNIST} & \multicolumn{2}{c}{PneumoniaMNIST} & \multicolumn{2}{c}{RetinaMNIST} \\
      \midrule
          Method & 10\% & 1\% & 10\% & 1\% & 10\% & 1\% & 10\% & 1\% & 10\% & 1\% \\
      \midrule
          SimCLR & 92.20\scriptsize{ $\pm$ 0.17} & 90.88\scriptsize{ $\pm$ 0.07} & 73.78\scriptsize{ $\pm$ 0.98} & 70.70\scriptsize{ $\pm$ 0.18} & 81.66\scriptsize{ $\pm$ 1.10} & 80.06\scriptsize{ $\pm$ 0.14} & 93.97\scriptsize{ $\pm$ 1.18} & 82.24\scriptsize{ $\pm$ 1.50} & 47.80\scriptsize{ $\pm$ 1.02} & 47.15\scriptsize{ $\pm$ 2.15}  \\
          DINO & \cellcolor{green!30}{93.64}\scriptsize{ $\pm$ 0.06} & 92.17\scriptsize{ $\pm$ 0.10} & \cellcolor{green!30}{75.20}\scriptsize{ $\pm$ 0.36} & 71.48\scriptsize{ $\pm$ 0.39} & 83.10\scriptsize{ $\pm$ 1.25} & 80.12\scriptsize{ $\pm$ 0.85} & 95.48\scriptsize{ $\pm$ 0.17} & 93.84\scriptsize{ $\pm$ 1.90} & 47.70\scriptsize{ $\pm$ 2.29} & 46.30\scriptsize{ $\pm$ 4.78}  \\
          BYOL & 93.48\scriptsize{ $\pm$ 0.23} & \cellcolor{green!30}{92.61}\scriptsize{ $\pm$ 0.16} & 74.04\scriptsize{ $\pm$ 0.46} & \cellcolor{green!30}{71.93}\scriptsize{ $\pm$ 0.51} & 84.26\scriptsize{ $\pm$ 0.30} & 77.90\scriptsize{ $\pm$ 0.65} & 95.22\scriptsize{ $\pm$ 0.55} & 93.26\scriptsize{ $\pm$ 1.93} & 45.70\scriptsize{ $\pm$ 1.56} & 42.95\scriptsize{ $\pm$ 2.12}  \\
          ReSSL & 91.95\scriptsize{ $\pm$ 0.07} & \cellcolor{red!30}{90.61}\scriptsize{ $\pm$ 0.20} & 74.60\scriptsize{ $\pm$ 0.50} & 69.84\scriptsize{ $\pm$ 0.64} & \cellcolor{red!30}{76.40}\scriptsize{ $\pm$ 0.29} & \cellcolor{red!30}{63.64}\scriptsize{ $\pm$ 0.14} & 95.80\scriptsize{ $\pm$ 0.09} & 88.14\scriptsize{ $\pm$ 0.24} & \cellcolor{green!30}{48.20}\scriptsize{ $\pm$ 2.54} & 46.35\scriptsize{ $\pm$ 2.16}  \\
\hline
  \end{tabular}
    \end{adjustbox}
  \label{table:scarcity_resnet50_imagenet_1}

    \begin{adjustbox}{width=500pt,center}
      \begin{tabular}{ccc|cc|cc|cc|cc|cc}
           & \multicolumn{2}{c}{BreastMNIST} & \multicolumn{2}{c}{BloodMNIST} & \multicolumn{2}{c}{TissueMNIST} & \multicolumn{2}{c}{OrganAMNIST} & \multicolumn{2}{c}{OrganCMNIST} & \multicolumn{2}{c}{OrganSMNIST} \\
      \midrule
          Method & 10\% & 1\% & 10\% & 1\% & 10\% & 1\% & 10\% & 1\% & 10\% & 1\% & 10\% & 1\% \\
      \midrule
          SimCLR & 80.51\scriptsize{ $\pm$ 0.91} & 73.97\scriptsize{ $\pm$ 0.44} & 97.58\scriptsize{ $\pm$ 0.15} & 96.03\scriptsize{ $\pm$ 0.20} & \cellcolor{green!30}{57.98}\scriptsize{ $\pm$ 0.07} & 54.12\scriptsize{ $\pm$ 0.15} & 89.78\scriptsize{ $\pm$ 0.27} & 84.90\scriptsize{ $\pm$ 0.72} & 87.56\scriptsize{ $\pm$ 0.63} & 73.27\scriptsize{ $\pm$ 2.31} & 73.58\scriptsize{ $\pm$ 1.29} & 63.84\scriptsize{ $\pm$ 0.29}  \\
          DINO & 80.25\scriptsize{ $\pm$ 2.48} & 74.35\scriptsize{ $\pm$ 1.87} & 97.11\scriptsize{ $\pm$ 0.29} & 96.77\scriptsize{ $\pm$ 0.08} & 57.91\scriptsize{ $\pm$ 0.15} & \cellcolor{green!30}{54.50}\scriptsize{ $\pm$ 0.07} & \cellcolor{green!30}{93.12}\scriptsize{ $\pm$ 0.20} & \cellcolor{green!30}{90.30}\scriptsize{ $\pm$ 0.54} & \cellcolor{green!30}{89.40}\scriptsize{ $\pm$ 0.42} & \cellcolor{green!30}{81.40}\scriptsize{ $\pm$ 0.11} & 75.55\scriptsize{ $\pm$ 0.22} & 63.17\scriptsize{ $\pm$ 0.49}  \\
          BYOL & 79.23\scriptsize{ $\pm$ 1.07} & 76.66\scriptsize{ $\pm$ 1.83} & \cellcolor{green!30}{97.82}\scriptsize{ $\pm$ 0.13} & \cellcolor{green!30}{97.47}\scriptsize{ $\pm$ 0.38} & 56.83\scriptsize{ $\pm$ 0.12} & 53.06\scriptsize{ $\pm$ 0.13} & 90.76\scriptsize{ $\pm$ 0.38} & 86.25\scriptsize{ $\pm$ 0.45} & 86.25\scriptsize{ $\pm$ 0.34} & 79.51\scriptsize{ $\pm$ 0.44} & \cellcolor{green!30}{76.66}\scriptsize{ $\pm$ 0.50} & 64.05\scriptsize{ $\pm$ 0.12}  \\
\hline
  \end{tabular}
    \end{adjustbox}
    \caption{In-domain performance of the self-supervised learning methods using ResNet-50 with \imagenet initialization with 1\% and 10\% of the labels.}
  \label{table:scarcity_resnet50_imagenet_2}
\end{table*}

    \begin{table*}[h!]
  \centering
    \addtocounter{table}{-1}
    \captionlistentry{}  
    \begin{adjustbox}{width=500pt,center}
      \begin{tabular}{ccc|cc|cc|cc|cc}
      \toprule
           & \multicolumn{2}{c}{PathMNIST} & \multicolumn{2}{c}{DermaMNIST} & \multicolumn{2}{c}{OCTMNIST} & \multicolumn{2}{c}{PneumoniaMNIST} & \multicolumn{2}{c}{RetinaMNIST} \\
      \midrule
          Method & 10\% & 1\% & 10\% & 1\% & 10\% & 1\% & 10\% & 1\% & 10\% & 1\% \\
      \midrule
          SimCLR & 90.19\scriptsize{ $\pm$ 0.29} & \cellcolor{green!30}{90.42}\scriptsize{ $\pm$ 0.14} & \cellcolor{green!30}{74.12}\scriptsize{ $\pm$ 0.80} & 69.58\scriptsize{ $\pm$ 0.60} & 71.58\scriptsize{ $\pm$ 0.84} & 66.42\scriptsize{ $\pm$ 1.03} & \cellcolor{green!30}{89.48}\scriptsize{ $\pm$ 0.73} & 80.03\scriptsize{ $\pm$ 1.63} & \cellcolor{red!30}{43.25}\scriptsize{ $\pm$ 2.19} & 43.35\scriptsize{ $\pm$ 2.16}  \\
          DINO & 88.79\scriptsize{ $\pm$ 0.62} & 89.40\scriptsize{ $\pm$ 0.59} & \cellcolor{red!30}{70.37}\scriptsize{ $\pm$ 1.45} & \cellcolor{red!30}{67.36}\scriptsize{ $\pm$ 0.78} & \cellcolor{red!30}{36.64}\scriptsize{ $\pm$ 0.19} & \cellcolor{red!30}{35.86}\scriptsize{ $\pm$ 0.24} & 82.53\scriptsize{ $\pm$ 1.93} & 82.50\scriptsize{ $\pm$ 1.21} & \cellcolor{green!30}{47.10}\scriptsize{ $\pm$ 4.46} & 44.35\scriptsize{ $\pm$ 2.46}  \\
          BYOL & 89.53\scriptsize{ $\pm$ 0.16} & 89.33\scriptsize{ $\pm$ 0.26} & 71.45\scriptsize{ $\pm$ 0.84} & 69.50\scriptsize{ $\pm$ 0.26} & 69.56\scriptsize{ $\pm$ 0.84} & 63.14\scriptsize{ $\pm$ 1.87} & 89.23\scriptsize{ $\pm$ 1.17} & \cellcolor{green!30}{88.59}\scriptsize{ $\pm$ 0.53} & 43.40\scriptsize{ $\pm$ 3.06} & \cellcolor{green!30}{44.75}\scriptsize{ $\pm$ 1.32}  \\
          ReSSL & \cellcolor{red!30}{86.95}\scriptsize{ $\pm$ 1.00} & 88.43\scriptsize{ $\pm$ 0.13} & 73.46\scriptsize{ $\pm$ 0.27} & 68.68\scriptsize{ $\pm$ 0.50} & 60.88\scriptsize{ $\pm$ 0.37} & 52.72\scriptsize{ $\pm$ 3.07} & 82.53\scriptsize{ $\pm$ 0.37} & \cellcolor{red!30}{79.39}\scriptsize{ $\pm$ 1.06} & 45.80\scriptsize{ $\pm$ 4.08} & 44.05\scriptsize{ $\pm$ 2.25}  \\
\hline
  \end{tabular}
    \end{adjustbox}
  \label{table:scarcity_vit_random_1}

    \begin{adjustbox}{width=500pt,center}
      \begin{tabular}{ccc|cc|cc|cc|cc|cc}
           & \multicolumn{2}{c}{BreastMNIST} & \multicolumn{2}{c}{BloodMNIST} & \multicolumn{2}{c}{TissueMNIST} & \multicolumn{2}{c}{OrganAMNIST} & \multicolumn{2}{c}{OrganCMNIST} & \multicolumn{2}{c}{OrganSMNIST} \\
      \midrule
          Method & 10\% & 1\% & 10\% & 1\% & 10\% & 1\% & 10\% & 1\% & 10\% & 1\% & 10\% & 1\% \\
      \midrule
          SimCLR & 76.15\scriptsize{ $\pm$ 0.67} & 76.02\scriptsize{ $\pm$ 2.55} & 91.62\scriptsize{ $\pm$ 0.36} & 81.91\scriptsize{ $\pm$ 0.48} & \cellcolor{green!30}{61.47}\scriptsize{ $\pm$ 0.03} & \cellcolor{green!30}{57.07}\scriptsize{ $\pm$ 0.11} & 90.11\scriptsize{ $\pm$ 0.19} & 83.51\scriptsize{ $\pm$ 0.27} & 84.60\scriptsize{ $\pm$ 0.28} & 71.45\scriptsize{ $\pm$ 0.41} & 70.12\scriptsize{ $\pm$ 0.36} & 57.78\scriptsize{ $\pm$ 0.26}  \\
          DINO & 73.07\scriptsize{ $\pm$ 0.00} & 73.07\scriptsize{ $\pm$ 0.00} & \cellcolor{red!30}{81.84}\scriptsize{ $\pm$ 0.43} & \cellcolor{red!30}72.55\scriptsize{ $\pm$ 1.25} & 56.17\scriptsize{ $\pm$ 0.13} & 53.38\scriptsize{ $\pm$ 0.56} & \cellcolor{red!30}{77.23}\scriptsize{ $\pm$ 0.16} & \cellcolor{red!30}{69.64}\scriptsize{ $\pm$ 0.95} & \cellcolor{red!30}{67.02}\scriptsize{ $\pm$ 0.50} & \cellcolor{red!30}{58.67}\scriptsize{ $\pm$ 0.43} & \cellcolor{red!30}{56.93}\scriptsize{ $\pm$ 0.30} & \cellcolor{red!30}{43.82}\scriptsize{ $\pm$ 0.54}  \\
          BYOL & 73.07\scriptsize{ $\pm$ 0.00} & 73.07\scriptsize{ $\pm$ 0.00} & 86.71\scriptsize{ $\pm$ 0.29} & 74.11\scriptsize{ $\pm$ 0.91} & 53.71\scriptsize{ $\pm$ 0.30} & 47.63\scriptsize{ $\pm$ 0.18} & 89.11\scriptsize{ $\pm$ 0.28} & 83.31\scriptsize{ $\pm$ 0.53} & 82.92\scriptsize{ $\pm$ 1.05} & 72.45\scriptsize{ $\pm$ 2.19} & 69.94\scriptsize{ $\pm$ 0.18} & \cellcolor{green!30}{58.87}\scriptsize{ $\pm$ 0.61}  \\
\hline
  \end{tabular}
    \end{adjustbox}
    \caption{In-domain performance of the self-supervised learning methods using ViT-Small with random initialization with 1\% and 10\% of the labels.}
  \label{table:scarcity_vit_random_2}
\end{table*}

    \begin{table*}[h!]
  \centering
    \addtocounter{table}{-1}
    \captionlistentry{}  
    \begin{adjustbox}{width=500pt,center}
      \begin{tabular}{ccc|cc|cc|cc|cc}
      \toprule
           & \multicolumn{2}{c}{PathMNIST} & \multicolumn{2}{c}{DermaMNIST} & \multicolumn{2}{c}{OCTMNIST} & \multicolumn{2}{c}{PneumoniaMNIST} & \multicolumn{2}{c}{RetinaMNIST} \\
      \midrule
          Method & 10\% & 1\% & 10\% & 1\% & 10\% & 1\% & 10\% & 1\% & 10\% & 1\% \\
      \midrule
          SimCLR & 92.19\scriptsize{ $\pm$ 0.25} & 89.12\scriptsize{ $\pm$ 0.14} & 74.10\scriptsize{ $\pm$ 0.45} & 69.22\scriptsize{ $\pm$ 0.85} & 76.90\scriptsize{ $\pm$ 0.65} & 75.34\scriptsize{ $\pm$ 0.64} & 94.87\scriptsize{ $\pm$ 0.34} & 82.24\scriptsize{ $\pm$ 1.65} & 46.80\scriptsize{ $\pm$ 1.21} & 45.25\scriptsize{ $\pm$ 0.49}  \\
          DINO & \cellcolor{green!30}{92.72}\scriptsize{ $\pm$ 0.19} & \cellcolor{green!30}{92.10}\scriptsize{ $\pm$ 0.11} & 75.07\scriptsize{ $\pm$ 0.44} & 70.57\scriptsize{ $\pm$ 0.45} & \cellcolor{green!30}{81.62}\scriptsize{ $\pm$ 0.67} & \cellcolor{green!30}{79.78}\scriptsize{ $\pm$ 0.56} & 93.84\scriptsize{ $\pm$ 2.13} & \cellcolor{green!30}{87.59}\scriptsize{ $\pm$ 2.82} & 46.85\scriptsize{ $\pm$ 0.92} & 43.10\scriptsize{ $\pm$ 3.56}  \\
          BYOL & \cellcolor{red!30}{89.99}\scriptsize{ $\pm$ 0.15} & 88.74\scriptsize{ $\pm$ 0.21} & \cellcolor{green!30}{75.86}\scriptsize{ $\pm$ 0.34} & \cellcolor{green!30}{71.02}\scriptsize{ $\pm$ 0.56} & \cellcolor{red!30}{69.34}\scriptsize{ $\pm$ 1.04} & \cellcolor{red!30}{64.32}\scriptsize{ $\pm$ 1.57} & \cellcolor{green!30}{94.90}\scriptsize{ $\pm$ 1.05} & 84.51\scriptsize{ $\pm$ 1.32} & \cellcolor{green!30}48.00\scriptsize{ $\pm$ 1.10} & 44.00\scriptsize{ $\pm$ 0.66}  \\
          ReSSL & 92.25\scriptsize{ $\pm$ 0.03} & \cellcolor{red!30}{87.45}\scriptsize{ $\pm$ 0.18} & 74.98\scriptsize{ $\pm$ 0.22} & 69.24\scriptsize{ $\pm$ 0.58} & 77.62\scriptsize{ $\pm$ 0.32} & 71.70\scriptsize{ $\pm$ 3.89} & 94.71\scriptsize{ $\pm$ 1.93} & 81.69\scriptsize{ $\pm$ 2.62} & 46.55\scriptsize{ $\pm$ 0.46} & 44.80\scriptsize{ $\pm$ 1.83}  \\
\hline
  \end{tabular}
    \end{adjustbox}
  \label{table:scarcity_vit_imagenet_1}

    \begin{adjustbox}{width=500pt,center}
      \begin{tabular}{ccc|cc|cc|cc|cc|cc}
           & \multicolumn{2}{c}{BreastMNIST} & \multicolumn{2}{c}{BloodMNIST} & \multicolumn{2}{c}{TissueMNIST} & \multicolumn{2}{c}{OrganAMNIST} & \multicolumn{2}{c}{OrganCMNIST} & \multicolumn{2}{c}{OrganSMNIST} \\
      \midrule
          Method & 10\% & 1\% & 10\% & 1\% & 10\% & 1\% & 10\% & 1\% & 10\% & 1\% & 10\% & 1\% \\
      \midrule
          SimCLR & \cellcolor{green!30}{82.56}\scriptsize{ $\pm$ 2.60} & 78.33\scriptsize{ $\pm$ 0.87} & 97.13\scriptsize{ $\pm$ 0.18} & 92.93\scriptsize{ $\pm$ 0.25} & 60.38\scriptsize{ $\pm$ 0.06} & 55.07\scriptsize{ $\pm$ 0.07} & 90.05\scriptsize{ $\pm$ 0.31} & 84.48\scriptsize{ $\pm$ 0.23} & 86.71\scriptsize{ $\pm$ 0.46} & 67.31\scriptsize{ $\pm$ 1.68} & 74.27\scriptsize{ $\pm$ 0.92} & 56.68\scriptsize{ $\pm$ 0.61}  \\
          DINO & \cellcolor{red!30}{77.30}\scriptsize{ $\pm$ 3.50} & 75.25\scriptsize{ $\pm$ 0.71} & \cellcolor{green!30}{97.84}\scriptsize{ $\pm$ 0.12} & \cellcolor{green!30}{95.66}\scriptsize{ $\pm$ 0.35} & \cellcolor{green!30}{62.33}\scriptsize{ $\pm$ 0.03} & \cellcolor{green!30}{57.67}\scriptsize{ $\pm$ 0.06} & \cellcolor{green!30}{92.38}\scriptsize{ $\pm$ 0.36} & \cellcolor{green!30}{86.50}\scriptsize{ $\pm$ 0.23} & \cellcolor{green!30}{88.37}\scriptsize{ $\pm$ 0.79} & \cellcolor{green!30}{77.08}\scriptsize{ $\pm$ 2.49} & \cellcolor{green!30}{74.56}\scriptsize{ $\pm$ 0.61} & \cellcolor{green!30}{65.72}\scriptsize{ $\pm$ 0.61}  \\
          BYOL & 78.84\scriptsize{ $\pm$ 4.74} & \cellcolor{green!30}{82.17}\scriptsize{ $\pm$ 0.67} & 97.48\scriptsize{ $\pm$ 0.12} & 95.52\scriptsize{ $\pm$ 0.05} & 56.84\scriptsize{ $\pm$ 0.15} & 51.66\scriptsize{ $\pm$ 0.05} & 86.47\scriptsize{ $\pm$ 1.41} & 77.61\scriptsize{ $\pm$ 3.21} & 85.02\scriptsize{ $\pm$ 0.40} & 75.99\scriptsize{ $\pm$ 2.10} & 72.24\scriptsize{ $\pm$ 0.41} & 64.33\scriptsize{ $\pm$ 0.45}  \\
\hline
  \end{tabular}
    \end{adjustbox}
    \caption{In-domain performance of the self-supervised learning methods using ViT-Small with \imagenet initialization with 1\% and 10\% of the labels.}
  \label{table:scarcity_vit_imagenet_2}
\end{table*}


\newpage
\phantom{.}
\newpage
\phantom{.}
\newpage
\phantom{.}
\newpage
\phantom{.}  
\newpage

\subsubsection{Out-of-distribution Detection}
\label{sec:appendix_ap_ood}

\paragraph{Effect of SSL Method}
\label{sec:appendix_ap_ood_density}

Tables \ref{table:ood_resnet}, \ref{table:ood_resnet_in1k}, \ref{table:ood_vit} and \ref{table:ood_vit_in1k} show performance metrics (AUROC and AUPR) for OOD detection across 8 self-supervised learning models on diverse medical imaging datasets. Each cell shows the average scores across all $(\mathcal{P}_{\text{ID}},\mathcal{P}_{\text{OOD}})$ pairs when the dataset mentioned in that row is considered $\mathcal{P}_{\text{ID}}$ and used for pre-training. Each model's performance is evaluated in terms of AUROC (higher values indicate better separability) and AUPR (higher values indicate better precision-recall performance) with the best results highlighted in \colorbox{green!30}{green} and the worst in \colorbox{red!30}{red} for AUROC. Similarly, for AUPR, the best results are highlighted in \colorbox{green!15}{light green} and the worst in \colorbox{red!15}{light red}.

For ResNet-50 architectures, we observe from Tables \ref{table:ood_resnet} and \ref{table:ood_resnet_in1k} that NNCLR consistently outperforms other models, achieving the highest average AUROC and AUPR across datasets. MoCo v3 demonstrates stable performance under both random and \imagenet initialization strategies, highlighting its robustness. In contrast, BYOL exhibits excellent performance under \imagenet initialization but performs poorly when initialized with random weights, indicating sensitivity to initialization strategies. SimCLR, on the other hand, achieves the lowest OOD detection scores when models are initialized with \imagenet weights, reflecting its relative inefficiency in such scenarios.

For ViT-Small architectures, Tables \ref{table:ood_vit} and \ref{table:ood_vit_in1k} reveal that MoCo v3 delivers the best performance across both random and \imagenet initialization strategies, establishing itself as the most robust approach for OOD detection in this setting. DINO achieves strong results under \imagenet initialization but experiences a significant decline in performance when initialized with random weights, indicating a reliance on pretrained feature representations. In contrast, VICReg shows the largest drop in average AUROC and AUPR when switching from random initialization to \imagenet initialization, suggesting a potential mismatch between its pretrained features and OOD detection requirements.

\begin{table}[h!]
\centering
\label{tab:performance_metrics}
\resizebox{1\textwidth}{!}{%
\begin{tabular}{l||cc|cc|cc|cc|cc|cc|cc|cc}
\toprule
Dataset & \multicolumn{2}{c|}{MoCo v3} & \multicolumn{2}{c|}{SimCLR} & \multicolumn{2}{c|}{DINO} & \multicolumn{2}{c|}{ReSSL} & \multicolumn{2}{c|}{BYOL} & \multicolumn{2}{c|}{VICReg} & \multicolumn{2}{c|}{NNCLR} & \multicolumn{2}{c}{Barlow Twins} \\
 & AUROC & AUPR & AUROC & AUPR & AUROC & AUPR & AUROC & AUPR & AUROC & AUPR & AUROC & AUPR & AUROC & AUPR & AUROC & AUPR \\
\midrule
PathMNIST & 0.982 & 0.952 & \cellcolor{red!30}0.872 & \cellcolor{red!15}0.774 & 0.974 & 0.956 & 0.936 & 0.904 & 0.982 & 0.964 & 0.986 & 0.968 & 0.973 & 0.945 & \cellcolor{green!30}0.990 & \cellcolor{green!15}0.973 \\
DermaMNIST & 0.988 & 0.949 & \cellcolor{green!30}0.994 & \cellcolor{green!15}0.978 & 0.925 & 0.556 & 0.978 & 0.909 & \cellcolor{red!30}0.516 & \cellcolor{red!15}0.173 & 0.984 & 0.927 & 0.965 & 0.900 & 0.987 & 0.965 \\
OCTMNIST & 0.904 & 0.674 & \cellcolor{red!30}0.888 & \cellcolor{red!15}0.644 & \cellcolor{green!30}0.998 & \cellcolor{green!15}0.994 & 0.952 & 0.819 & 0.983 & 0.962 & 0.989 & 0.934 & 0.975 & 0.907 & 0.955 & 0.874 \\
PneumoniaMNIST & 0.998 & 0.994 & 0.998 & 0.995 & 0.924 & 0.311 & 0.990 & 0.964 & \cellcolor{red!30}0.808 & \cellcolor{red!15}0.189 & 0.997 & 0.980 & \cellcolor{green!30}0.998 & \cellcolor{green!15}0.996 & 0.988 & 0.925 \\
BreastMNIST & 0.568 & \cellcolor{red!15}0.019 & 0.751 & 0.028 & \cellcolor{green!30}0.943 & \cellcolor{green!15}0.714 & 0.525 & 0.020 & 0.911 & 0.596 & \cellcolor{red!30}0.498 & 0.033 & 0.840 & 0.122 & 0.843 & 0.068 \\
BloodMNIST & \cellcolor{green!30}0.997 & 0.986 & 0.984 & 0.930 & 0.997 & \cellcolor{green!15}0.991 & 0.991 & 0.977 & 0.985 & 0.965 & 0.994 & 0.977 & 0.997 & 0.989 & \cellcolor{red!30}0.954 & \cellcolor{red!15}0.908 \\
TissueMNIST & 0.995 & 0.996 & 0.995 & 0.997 & 0.998 & 0.999 & 0.998 & 0.999 & 0.999 & 0.999 & 0.998 & 0.999 & \cellcolor{red!30}0.990 & \cellcolor{red!15}0.994 & \cellcolor{green!30}0.999 & \cellcolor{green!15}0.999 \\
OrganAMNIST & 0.864 & 0.823 & 0.740 & \cellcolor{red!15}0.649 & 0.823 & 0.804 & 0.725 & 0.711 & \cellcolor{red!30}0.672 & 0.702 & 0.772 & 0.722 & \cellcolor{green!30}0.894 & \cellcolor{green!15}0.874 & 0.851 & 0.816 \\
OrganCMNIST & 0.878 & 0.791 & 0.871 & 0.782 & \cellcolor{red!30}0.720 & \cellcolor{red!15}0.627 & \cellcolor{green!30}0.906 & \cellcolor{green!15}0.821 & 0.780 & 0.701 & 0.831 & 0.724 & 0.866 & 0.776 & 0.823 & 0.717 \\
OrganSMNIST & \cellcolor{green!30}0.894 & 0.789 & 0.852 & 0.760 & \cellcolor{red!30}0.703 & \cellcolor{red!15}0.605 & 0.894 & \cellcolor{green!15}0.800 & 0.786 & 0.709 & 0.813 & 0.733 & 0.881 & 0.795 & 0.741 & 0.649 \\
\midrule
Average & 0.907 & 0.797 & 0.894 & 0.754 & 0.901 & 0.756 & 0.889 & 0.792 & \cellcolor{red!30}0.842 & \cellcolor{red!15}0.696 & 0.886 & 0.800 & \cellcolor{green!30}0.938 & \cellcolor{green!15}0.830 & 0.913 & 0.789 \\
\bottomrule
\end{tabular}%
}
\caption{Average OOD detection performance of ResNet-50 models initialized with random weights for different datasets}
\label{table:ood_resnet}
\end{table}

\begin{table}[h!]
\centering
\label{tab:performance_metrics}
\resizebox{1\textwidth}{!}{%
\begin{tabular}{l||cc|cc|cc|cc|cc|cc|cc|cc}
\toprule
Dataset & \multicolumn{2}{c|}{MoCo v3} & \multicolumn{2}{c|}{SimCLR} & \multicolumn{2}{c|}{DINO} & \multicolumn{2}{c|}{ReSSL} & \multicolumn{2}{c|}{BYOL} & \multicolumn{2}{c|}{VICReg} & \multicolumn{2}{c|}{NNCLR} & \multicolumn{2}{c}{Barlow Twins} \\
 & AUROC & AUPR & AUROC & AUPR & AUROC & AUPR & AUROC & AUPR & AUROC & AUPR & AUROC & AUPR & AUROC & AUPR & AUROC & AUPR \\
\midrule
PathMNIST & 0.974 & 0.955 & \cellcolor{red!30}0.765 & \cellcolor{red!15}0.627 & 0.979 & 0.963 & 0.944 & 0.906 & \cellcolor{green!30}0.982 & \cellcolor{green!15}0.966 & 0.878 & 0.805 & 0.979 & 0.955 & 0.914 & 0.858 \\
DermaMNIST & 0.959 & 0.850 & \cellcolor{red!30}0.795 & 0.522 & 0.981 & 0.944 & 0.976 & 0.935 & \cellcolor{green!30}0.994 & \cellcolor{green!15}0.972 & 0.835 & \cellcolor{red!15}0.509 & 0.993 & 0.959 & 0.959 & 0.833 \\
OCTMNIST & 0.920 & 0.667 & 0.819 & 0.503 & \cellcolor{green!30}0.998 & \cellcolor{green!15}0.991 & \cellcolor{red!30}0.751 & \cellcolor{red!15}0.257 & 0.994 & 0.967 & 0.915 & 0.648 & 0.969 & 0.878 & 0.954 & 0.805 \\
PneumoniaMNIST & 0.971 & 0.916 & \cellcolor{red!30}0.580 & \cellcolor{red!15}0.189 & 0.986 & 0.958 & 0.987 & 0.958 & 0.970 & 0.941 & 0.886 & 0.418 & \cellcolor{green!30}0.992 & \cellcolor{green!15}0.964 & 0.971 & 0.931 \\
BreastMNIST & 0.996 & 0.928 & 0.998 & 0.941 & 0.964 & 0.451 & \cellcolor{red!30}0.715 & \cellcolor{red!15}0.043 & 0.997 & 0.912 & 0.983 & 0.697 & \cellcolor{green!30}0.998 & \cellcolor{green!15}0.943 & 0.993 & 0.876 \\
BloodMNIST & 0.992 & 0.968 & \cellcolor{red!30}0.908 & \cellcolor{red!15}0.727 & 0.986 & 0.956 & \cellcolor{green!30}0.999 & \cellcolor{green!15}0.997 & 0.991 & 0.952 & 0.941 & 0.829 & 0.996 & 0.986 & 0.979 & 0.919 \\
TissueMNIST & 0.995 & 0.997 & 0.994 & 0.996 & \cellcolor{red!30}0.650 & \cellcolor{red!15}0.695 & 0.994 & 0.995 & \cellcolor{green!30}0.999 & \cellcolor{green!15}0.999 & 0.983 & 0.994 & 0.999 & 0.999 & 0.993 & 0.997 \\
OrganAMNIST & 0.830 & 0.754 & \cellcolor{red!30}0.524 & \cellcolor{red!15}0.526 & 0.812 & 0.734 & 0.613 & 0.614 & 0.822 & 0.760 & 0.867 & 0.827 & \cellcolor{green!30}0.899 & \cellcolor{green!15}0.829 & 0.875 & 0.826 \\
OrganCMNIST & 0.770 & 0.663 & \cellcolor{red!30}0.604 & \cellcolor{red!15}0.480 & 0.725 & 0.575 & 0.623 & 0.508 & 0.859 & 0.753 & 0.809 & 0.694 & 0.848 & 0.742 & \cellcolor{green!30}0.879 & \cellcolor{green!15}0.772 \\
OrganSMNIST & 0.755 & 0.619 & 0.632 & \cellcolor{red!15}0.497 & \cellcolor{red!30}0.620 & 0.520 & 0.627 & 0.514 & 0.757 & 0.637 & 0.744 & 0.577 & 0.821 & 0.705 & \cellcolor{green!30}0.857 & \cellcolor{green!15}0.748 \\
\midrule
Average & 0.916 & 0.832 & \cellcolor{red!30}0.762 & \cellcolor{red!15}0.601 & 0.870 & 0.779 & 0.823 & 0.673 & 0.937 & 0.886 & 0.884 & 0.700 & \cellcolor{green!30}0.949 & \cellcolor{green!15}0.896 & 0.937 & 0.856 \\
\bottomrule
\end{tabular}%
}
\caption{Average OOD detection performance of ResNet-50 models initialized with \imagenet weights evaluated across different datasets.}
\label{table:ood_resnet_in1k}
\end{table}

\begin{table}[h!]
\centering
\label{tab:performance_metrics}
\resizebox{1\textwidth}{!}{%
\begin{tabular}{l||cc|cc|cc|cc|cc|cc|cc|cc}
\toprule
Dataset & \multicolumn{2}{c|}{MoCo v3} & \multicolumn{2}{c|}{SimCLR} & \multicolumn{2}{c|}{DINO} & \multicolumn{2}{c|}{ReSSL} & \multicolumn{2}{c|}{BYOL} & \multicolumn{2}{c|}{VICReg} & \multicolumn{2}{c|}{NNCLR} & \multicolumn{2}{c}{Barlow Twins} \\
 & AUROC & AUPR & AUROC & AUPR & AUROC & AUPR & AUROC & AUPR & AUROC & AUPR & AUROC & AUPR & AUROC & AUPR & AUROC & AUPR \\
\midrule
PathMNIST & 0.999 & 0.999 & 0.997 & 0.997 & 0.996 & 0.996 & 0.996 & 0.995 & 0.995 & \cellcolor{red!15}0.989 & 0.997 & 0.996 & \cellcolor{green!30}0.999 & \cellcolor{green!15}0.999 & \cellcolor{red!30}0.993 & 0.991 \\
DermaMNIST & \cellcolor{green!30}0.998 & \cellcolor{green!15}0.995 & 0.997 & 0.991 & 0.996 & 0.986 & 0.997 & 0.986 & 0.997 & 0.991 & \cellcolor{red!30}0.986 & \cellcolor{red!15}0.938 & 0.996 & 0.988 & 0.995 & 0.986 \\
OCTMNIST & 0.992 & 0.969 & \cellcolor{green!30}0.996 & \cellcolor{green!15}0.974 & 0.971 & 0.890 & 0.994 & 0.960 & 0.980 & 0.928 & 0.965 & 0.801 & 0.975 & 0.920 & \cellcolor{red!30}0.760 & \cellcolor{red!15}0.309 \\
PneumoniaMNIST & \cellcolor{green!30}0.999 & 0.984 & 0.998 & 0.986 & 0.988 & 0.912 & 0.992 & 0.894 & 0.990 & 0.937 & \cellcolor{red!30}0.985 & \cellcolor{red!15}0.835 & 0.999 & \cellcolor{green!15}0.990 & 0.989 & 0.877 \\
BreastMNIST & 0.987 & 0.811 & 0.955 & 0.768 & \cellcolor{red!30}0.690 & \cellcolor{red!15}0.184 & 0.869 & 0.571 & 0.749 & 0.258 & 0.970 & 0.796 & 0.965 & 0.751 & \cellcolor{green!30}0.995 & \cellcolor{green!15}0.851 \\
BloodMNIST & \cellcolor{green!30}0.999 & \cellcolor{green!15}0.999 & 0.999 & 0.999 & 0.999 & 0.999 & 0.999 & 0.999 & 0.999 & 0.999 & \cellcolor{red!30}0.999 & \cellcolor{red!15}0.996 & 0.999 & 0.999 & 0.999 & 0.999 \\
TissueMNIST & 0.999 & 0.999 & \cellcolor{green!30}0.999 & \cellcolor{green!15}0.999 & 0.999 & 0.999 & 0.998 & \cellcolor{red!15}0.998 & \cellcolor{red!30}0.998 & 0.999 & 0.998 & 0.999 & 0.999 & 0.999 & 0.998 & 0.999 \\
OrganAMNIST & 0.924 & \cellcolor{green!15}0.911 & \cellcolor{green!30}0.929 & 0.899 & 0.889 & \cellcolor{red!15}0.843 & \cellcolor{red!30}0.888 & 0.865 & 0.928 & 0.911 & 0.923 & 0.904 & 0.924 & 0.897 & 0.911 & 0.880 \\
OrganCMNIST & \cellcolor{green!30}0.870 & 0.762 & 0.869 & \cellcolor{green!15}0.773 & \cellcolor{red!30}0.796 & 0.682 & 0.821 & 0.677 & 0.850 & 0.753 & 0.820 & \cellcolor{red!15}0.668 & 0.850 & 0.754 & 0.868 & 0.753 \\
OrganSMNIST & 0.825 & 0.728 & 0.845 & 0.757 & 0.808 & 0.712 & 0.829 & 0.713 & \cellcolor{green!30}0.860 & \cellcolor{green!15}0.773 & \cellcolor{red!30}0.727 & \cellcolor{red!15}0.578 & 0.842 & 0.743 & 0.842 & 0.729 \\
\midrule
Average & \cellcolor{green!30}0.959 & \cellcolor{green!15}0.916 & 0.959 & 0.914 & \cellcolor{red!30}0.913 & \cellcolor{red!15}0.820 & 0.938 & 0.866 & 0.935 & 0.854 & 0.937 & 0.851 & 0.955 & 0.904 & 0.935 & 0.837 \\
\bottomrule
\end{tabular}%
}
\caption{Average OOD detection performance of ViT-Small models initialized with random weights evaluated across different datasets}
\label{table:ood_vit}
\end{table}

\begin{table}[h!]
\centering
\label{tab:performance_metrics}
\resizebox{1\textwidth}{!}{%
\begin{tabular}{l||cc|cc|cc|cc|cc|cc|cc|cc}
\toprule
Dataset & \multicolumn{2}{c|}{MoCo v3} & \multicolumn{2}{c|}{SimCLR} & \multicolumn{2}{c|}{DINO} & \multicolumn{2}{c|}{ReSSL} & \multicolumn{2}{c|}{BYOL} & \multicolumn{2}{c|}{VICReg} & \multicolumn{2}{c|}{NNCLR} & \multicolumn{2}{c}{Barlow Twins} \\
 & AUROC & AUPR & AUROC & AUPR & AUROC & AUPR & AUROC & AUPR & AUROC & AUPR & AUROC & AUPR & AUROC & AUPR & AUROC & AUPR \\
\midrule
PathMNIST & 0.987 & 0.965 & 0.974 & 0.956 & \cellcolor{green!30}0.996 & \cellcolor{green!15}0.994 & 0.975 & 0.959 & \cellcolor{red!30}0.807 & \cellcolor{red!15}0.722 & 0.949 & 0.901 & 0.987 & 0.972 & 0.967 & 0.934 \\
DermaMNIST & 0.993 & 0.966 & 0.994 & 0.972 & 0.995 & 0.971 & \cellcolor{green!30}0.997 & 0.983 & 0.996 & \cellcolor{green!15}0.988 & \cellcolor{red!30}0.859 & \cellcolor{red!15}0.563 & 0.994 & 0.964 & 0.965 & 0.815 \\
OCTMNIST & 0.976 & 0.852 & 0.977 & 0.892 & \cellcolor{green!30}0.999 & \cellcolor{green!15}0.998 & 0.981 & 0.863 & 0.852 & 0.743 & \cellcolor{red!30}0.810 & \cellcolor{red!15}0.292 & 0.898 & 0.645 & 0.932 & 0.664 \\
PneumoniaMNIST & \cellcolor{green!30}0.999 & 0.998 & 0.998 & 0.996 & 0.984 & 0.946 & 0.998 & 0.981 & 0.997 & 0.983 & \cellcolor{red!30}0.901 & \cellcolor{red!15}0.464 & 0.999 & \cellcolor{green!15}0.998 & 0.993 & 0.942 \\
BreastMNIST & \cellcolor{green!30}0.999 & \cellcolor{green!15}0.981 & 0.964 & 0.629 & 0.992 & 0.908 & 0.995 & 0.931 & 0.996 & 0.941 & \cellcolor{red!30}0.948 & \cellcolor{red!15}0.587 & 0.999 & 0.946 & 0.955 & 0.687 \\
BloodMNIST & 0.998 & 0.984 & 0.998 & 0.990 & \cellcolor{green!30}0.999 & 0.997 & 0.999 & \cellcolor{green!15}0.998 & 0.996 & 0.993 & 0.960 & 0.831 & 0.997 & 0.983 & \cellcolor{red!30}0.952 & \cellcolor{red!15}0.806 \\
TissueMNIST & 0.998 & 0.999 & 0.999 & 0.999 & \cellcolor{green!30}0.999 & 0.999 & 0.984 & 0.993 & 0.957 & 0.980 & \cellcolor{red!30}0.809 & \cellcolor{red!15}0.856 & 0.999 & \cellcolor{green!15}0.999 & 0.934 & 0.968 \\
OrganAMNIST & \cellcolor{green!30}0.931 & \cellcolor{green!15}0.915 & 0.793 & 0.756 & 0.907 & 0.903 & \cellcolor{red!30}0.782 & 0.750 & 0.902 & 0.872 & 0.808 & \cellcolor{red!15}0.743 & 0.849 & 0.824 & 0.802 & 0.781 \\
OrganCMNIST & 0.836 & 0.702 & 0.837 & 0.738 & 0.819 & \cellcolor{green!15}0.759 & \cellcolor{red!30}0.771 & 0.687 & 0.824 & 0.671 & 0.827 & \cellcolor{red!15}0.660 & 0.805 & 0.679 & \cellcolor{green!30}0.874 & 0.727 \\
OrganSMNIST & 0.843 & \cellcolor{green!15}0.738 & 0.774 & 0.644 & 0.802 & 0.727 & 0.800 & 0.715 & \cellcolor{red!30}0.724 & \cellcolor{red!15}0.600 & 0.816 & 0.638 & 0.811 & 0.704 & \cellcolor{green!30}0.850 & 0.713 \\
\midrule
Average & \cellcolor{green!30}0.956 & 0.910 & 0.931 & 0.857 & 0.949 & \cellcolor{green!15}0.920 & 0.928 & 0.886 & 0.905 & 0.849 & \cellcolor{red!30}0.869 & \cellcolor{red!15}0.654 & 0.934 & 0.872 & 0.922 & 0.804 \\
\bottomrule
\end{tabular}%
}
\caption{Average OOD detection performance of ViT-Small models initialized with \imagenet weights evaluated across different datasets.}
\label{table:ood_vit_in1k}
\end{table}

\begin{figure}[h]
    \centering
    \includegraphics[width=0.72\linewidth]{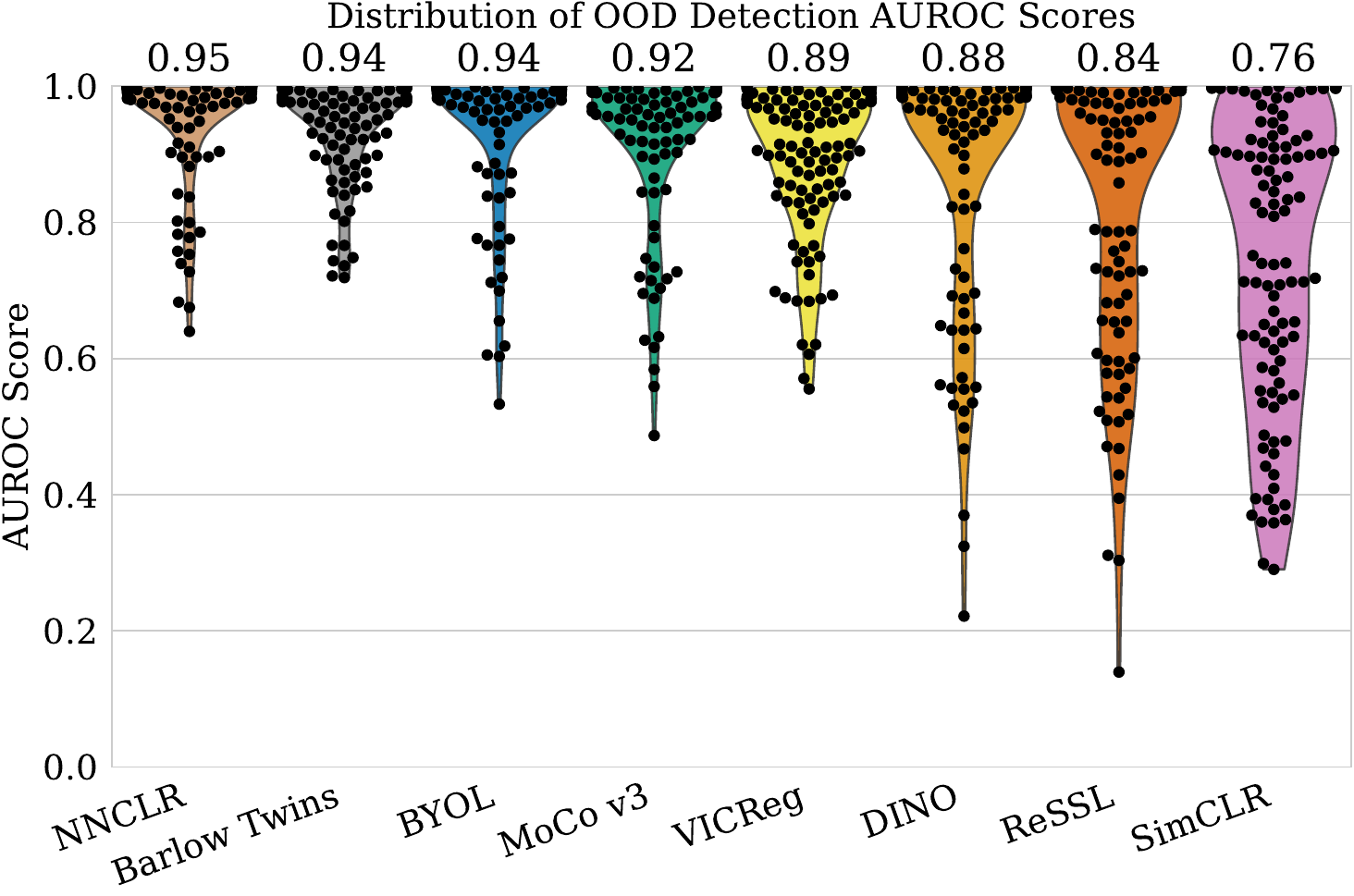}
    \caption{AUROC score distributions for various SSL methods on \imagenet initialized ResNet-50 backbone in OOD detection.}
    \label{fig:resnet_pretrained_ood_auroc}
\end{figure}

\begin{figure}[h]
    \centering
    \includegraphics[width=0.72\linewidth]{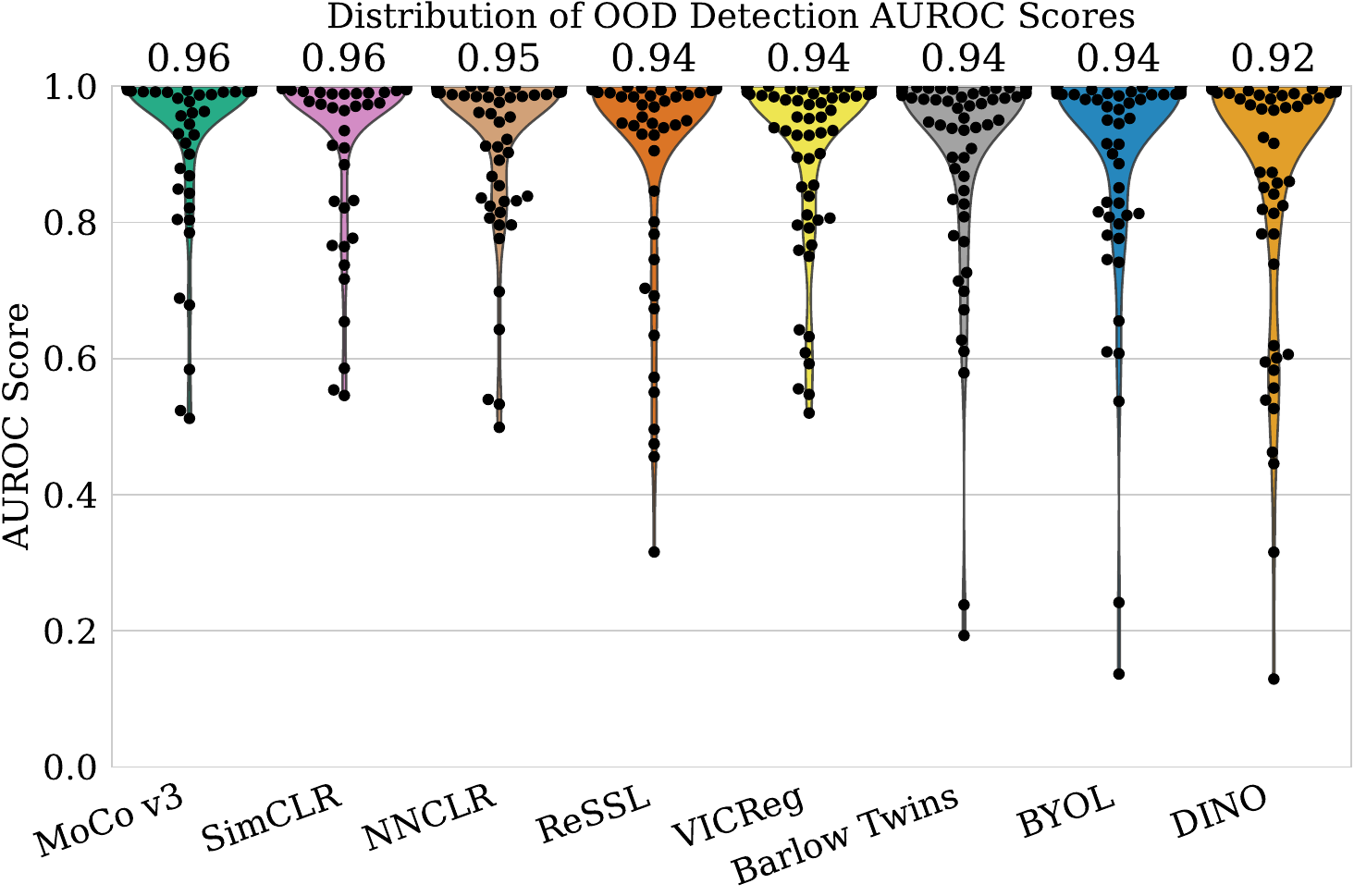}
    \caption{AUROC score distributions for various SSL methods on random initialized ViT-Small backbone in OOD detection.}
    \label{fig:vit_not_pretrained_ood_auroc}
\end{figure}

\begin{figure}[h]
    \centering
    \includegraphics[width=0.72\linewidth]{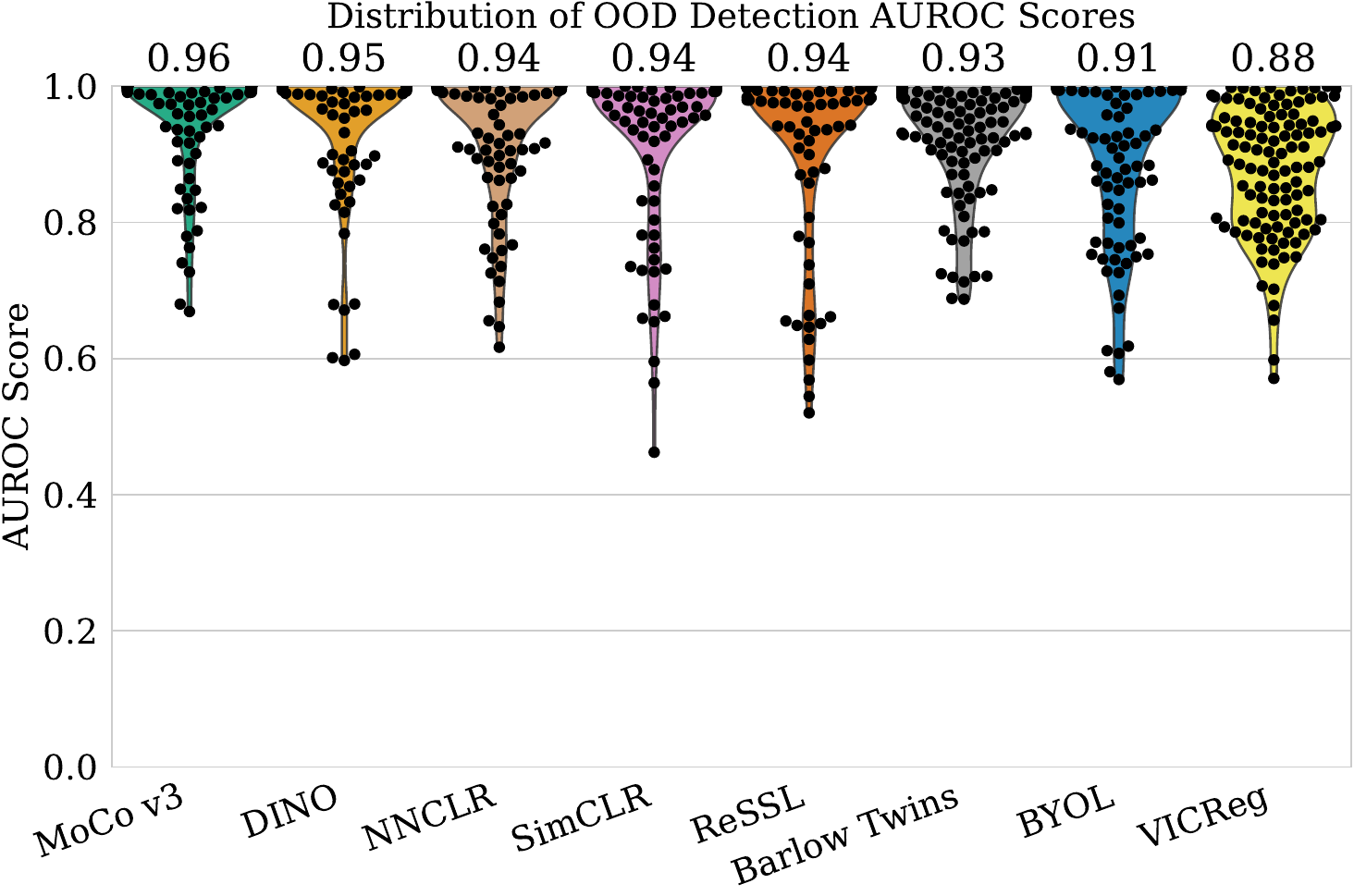}
    \caption{AUROC score distributions for various SSL methods on \imagenet initialized ViT-Small backbones in OOD detection.}
    \label{fig:vit_pretrained_ood_auroc}
\end{figure}

Figures~\ref{fig:resnet_pretrained_ood_auroc}, \ref{fig:vit_pretrained_ood_auroc}, and \ref{fig:vit_not_pretrained_ood_auroc} present the density distributions of AUROC scores for various SSL methods across different model backbones and initialization schemes. These visualizations provide insights into the performance of each method in OOD detection under distinct configurations. Notably, MoCo v3 consistently achieves the highest AUROC scores with ViT-Small backbones, both when randomly initialized and when initialized with \imagenet weights. In contrast, NNCLR demonstrates superior OOD detection performance when paired with a ResNet-50 backbone. Figure \ref{fig:resnet_pretrained_ood_auroc} suggests that for SimCLR models, \imagenet  initialization with a ResNet-50 backbone results in a distinctively low AUROC score for OOD detection. This variability in performance highlights the impact of both the SSL method and the model architecture on OOD detection capabilities, emphasizing the importance of selecting the right combination for optimal results. 

\paragraph{Effect of Backbone Architecture}
\label{sec:appendix_ap_ood_backbone}

\begin{figure*}[h]
    \centering
    \includegraphics[width=0.5\textwidth]{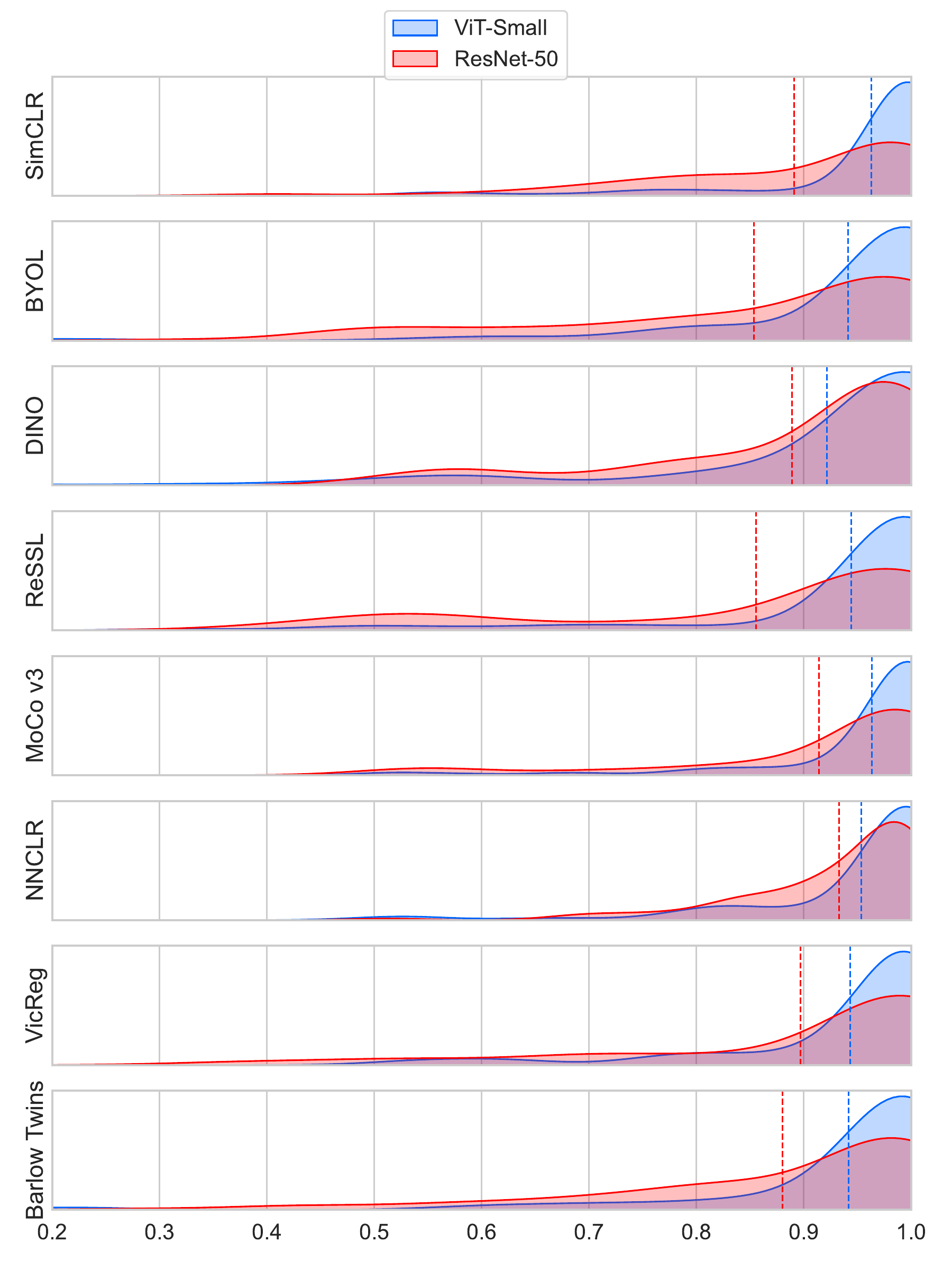}
    \captionsetup{justification=centering}
    \caption{
    Distribution of AUROC scores for different SSL methods.
        }
    \label{fig:multiple-density-plots}
\end{figure*}

The choice of backbone architecture is crucial for OOD detection performance. Figure~\ref{fig:multiple-density-plots} demonstrates how the choice of backbone affects the distribution of OOD AUROC scores for each method, strongly suggesting that ViT-Small is the superior option among all SSL methods in general. A comparison between Figures~\ref{fig:resnet50_method_auroc} and \ref{fig:vit_not_pretrained_ood_auroc} for random initialization setting reveals that ViT-Small architectures exhibit a lower variance and a higher density of AUROC scores in the upper range, indicating their effectiveness in OOD detection. Specifically, as depicted in Figure~\ref{fig:vit_not_pretrained_ood_auroc}, ViT-Small models, particularly those employing MoCo v3 and SimCLR, show pronounced peak densities near AUROC scores of $0.96$. This is significantly higher compared to the ResNet-50 models shown in Figure~\ref{fig:resnet50_method_auroc}, where the AUROC scores are more broadly distributed with a maximum around $0.93$. Furthermore, this observation holds true for the \imagenet initialization setting as well. Figures ~\ref{fig:resnet_pretrained_ood_auroc} and \ref{fig:vit_pretrained_ood_auroc}  demonstrate that the ViT-Small architecture consistently exhibits a higher density of large AUROC scores compared to the ResNet-50 architecture.

\begin{figure}[h]
    \centering
    \includegraphics[width=\linewidth]{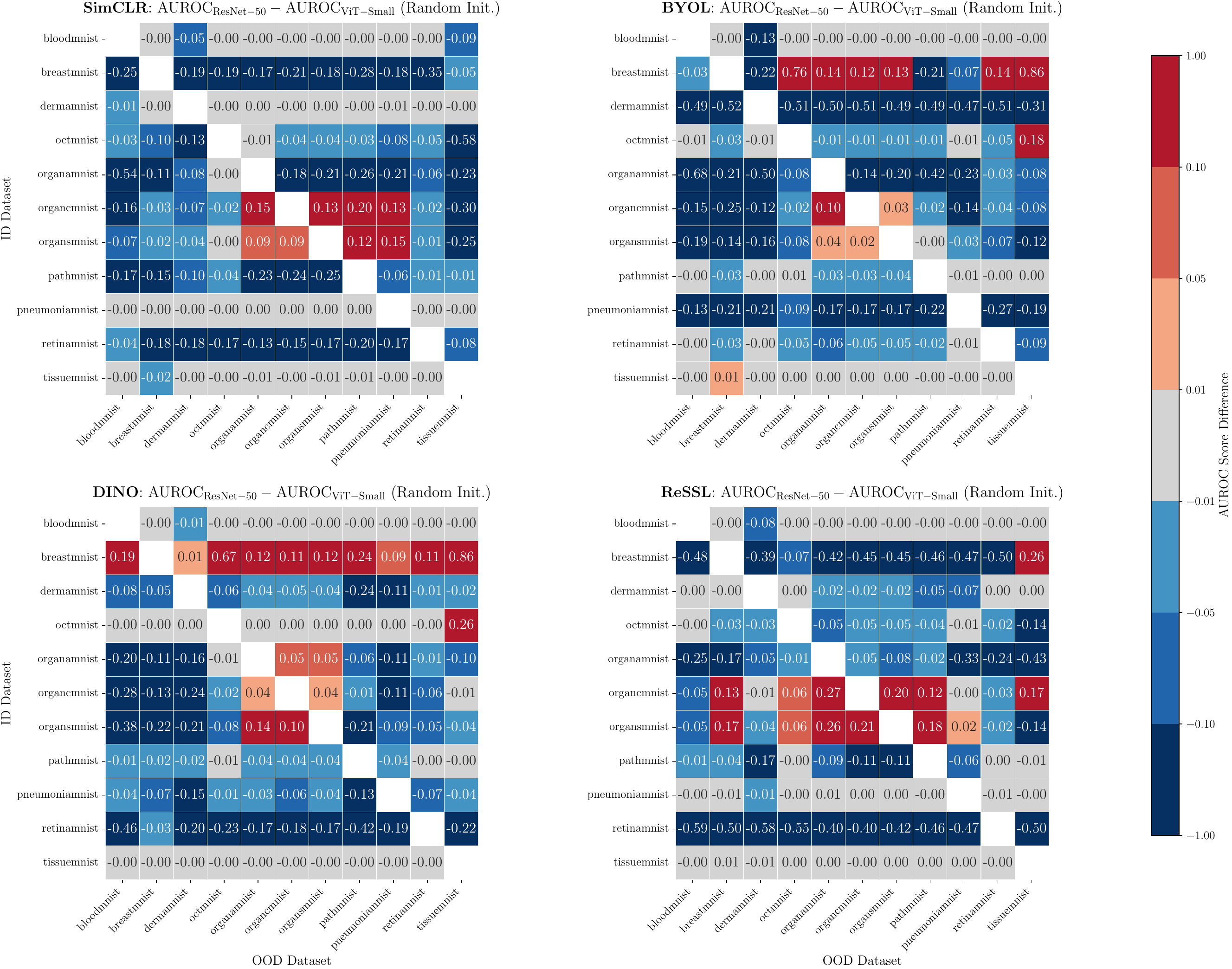}
    \caption{Effect of backbone choice (ResNet vs. ViT) on AUROC scores for OOD detection across various dataset combinations and SSL methods with random initialization. Negative values (blue) indicate better OOD detection performance with the ViT backbone, while positive values (red) favor ResNet-50.}
    \label{fig:resnet_vit_random_init_p1}
\end{figure}

\begin{figure}[h]
    \centering
    \includegraphics[width=\linewidth]{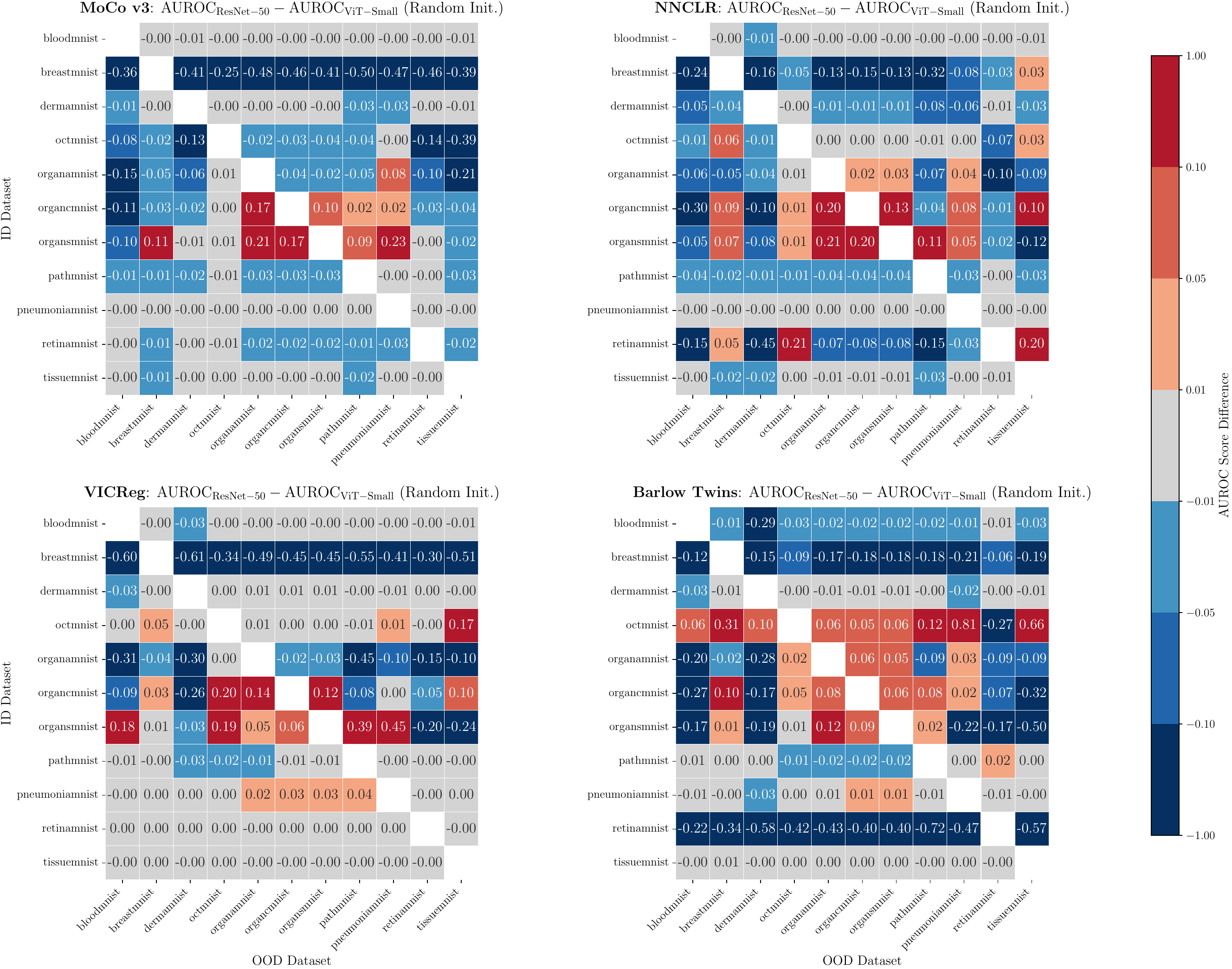}
    \caption{Effect of backbone choice (ResNet vs. ViT) on AUROC scores for OOD detection across various dataset combinations and SSL methods with random initialization. Negative values (blue) indicate better OOD detection performance with the ViT backbone, while positive values (red) favor ResNet-50.}
    \label{fig:resnet_vit_random_init_p2}
\end{figure}

\begin{figure}[h]
    \centering
    \includegraphics[width=\linewidth]{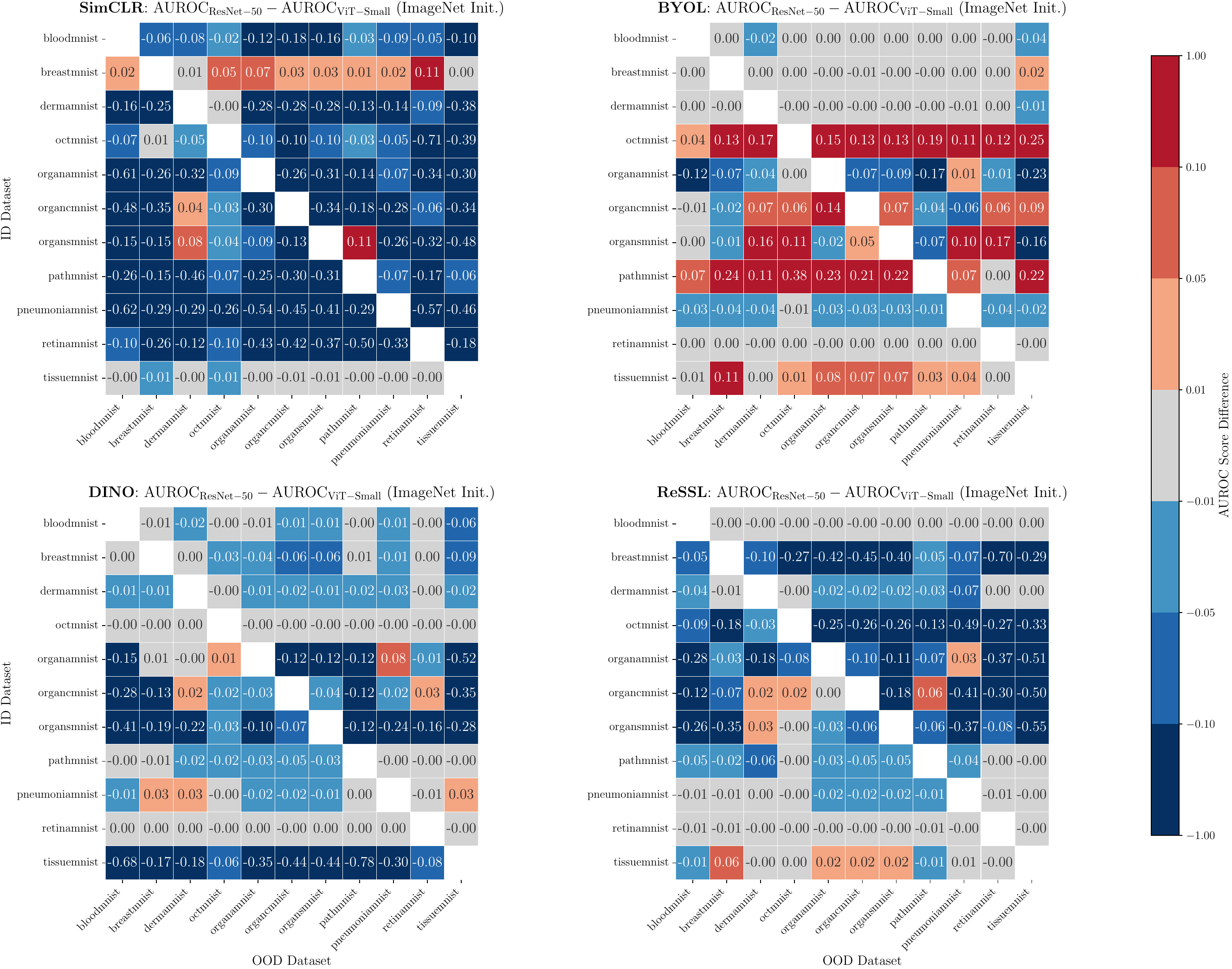}
    \caption{Effect of backbone choice (ResNet vs. ViT) on AUROC scores for OOD detection across various dataset combinations and SSL methods with \imagenet initialization. Negative values (blue) indicate better OOD detection performance with the ViT backbone, while positive values (red) favor ResNet-50.}
    \label{fig:resnet_vit_in1k_init_p1}
\end{figure}

\begin{figure}[h]
    \centering
    \includegraphics[width=\linewidth]{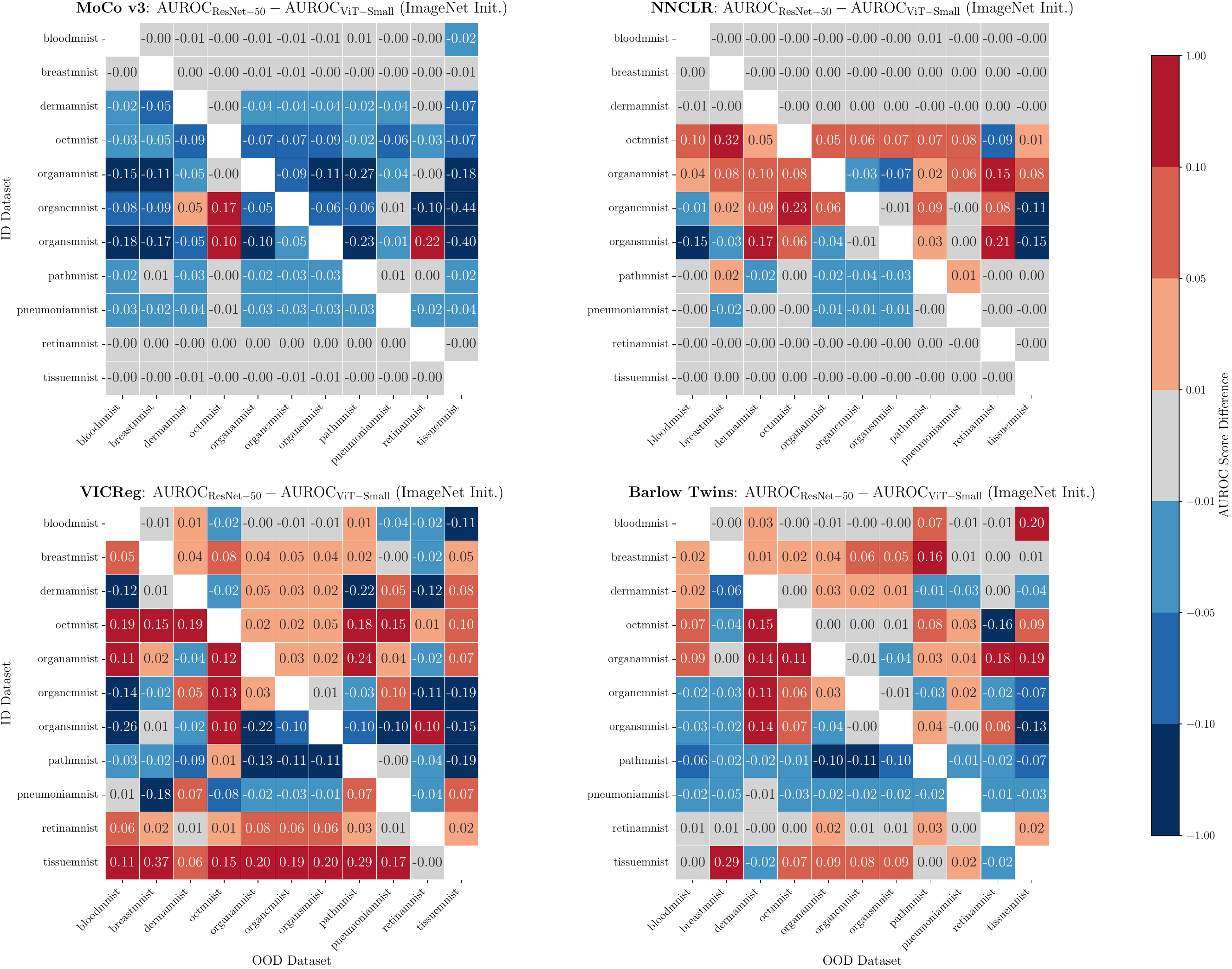}
    \caption{Effect of backbone choice (ResNet vs. ViT) on AUROC scores for OOD detection across various dataset combinations and SSL methods with \imagenet initialization. Negative values (blue) indicate better OOD detection performance with the ViT backbone, while positive values (red) favor ResNet-50.}
    \label{fig:resnet_vit_in1k_init_p2}
\end{figure}

To investigate the impact of backbone architecture in more detail, we analyze the effect of backbone among different $(\mathcal{P}_{\text{ID}}, \mathcal{P}_{\text{OOD}})$ pairs in Figures~\ref{fig:resnet_vit_random_init_p1} and \ref{fig:resnet_vit_random_init_p2}. For random initialization, we observe an improvement in OOD scores when ViT-Small is used across many of the $(\mathcal{P}_{\text{ID}}, \mathcal{P}_{\text{OOD}})$ pairs. In contrast, Figures~\ref{fig:resnet_vit_in1k_init_p1} and \ref{fig:resnet_vit_in1k_init_p2} reveal a shift in backbone preferences when models are initialized with \imagenet weights.
Notably, SimCLR, DINO, ReSSL, and MoCo v3 tend to favor ViT-Small for the majority of $(\mathcal{P}_{\text{ID}}, \mathcal{P}_{\text{OOD}})$ pairs, whereas BYOL, NNCLR, and VICReg demonstrate a preference for ResNet-50. For Barlow Twins, there is no clear preference for either backbone. 


\paragraph{Effect of Initialization}
\label{sec:appendix_ap_ood_initialization}

\begin{figure}[h]
    \centering
    \includegraphics[width=\linewidth]{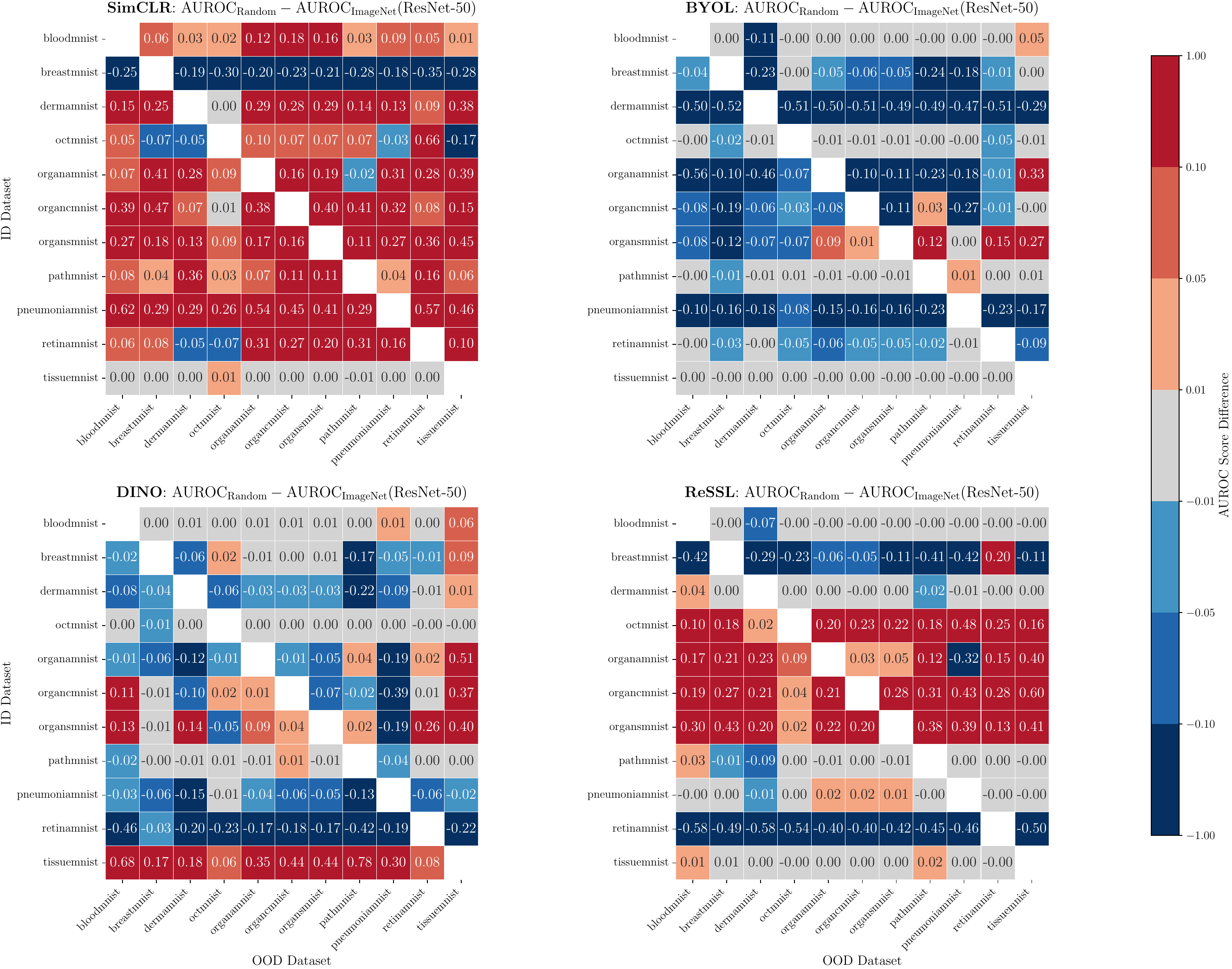}
    \caption{Effect of Initialization (Random vs. \imagenet) on AUROC scores for OOD detection across various dataset combinations and SSL methods with ResNet-50 backbone. Positive values (blue) indicate better OOD detection performance with the random initialization, while negative values (red) favor \imagenet initialization.}
    \label{fig:resnet_random_imagenet_p1}
\end{figure}

\begin{figure}[h]
    \centering
    \includegraphics[width=\linewidth]{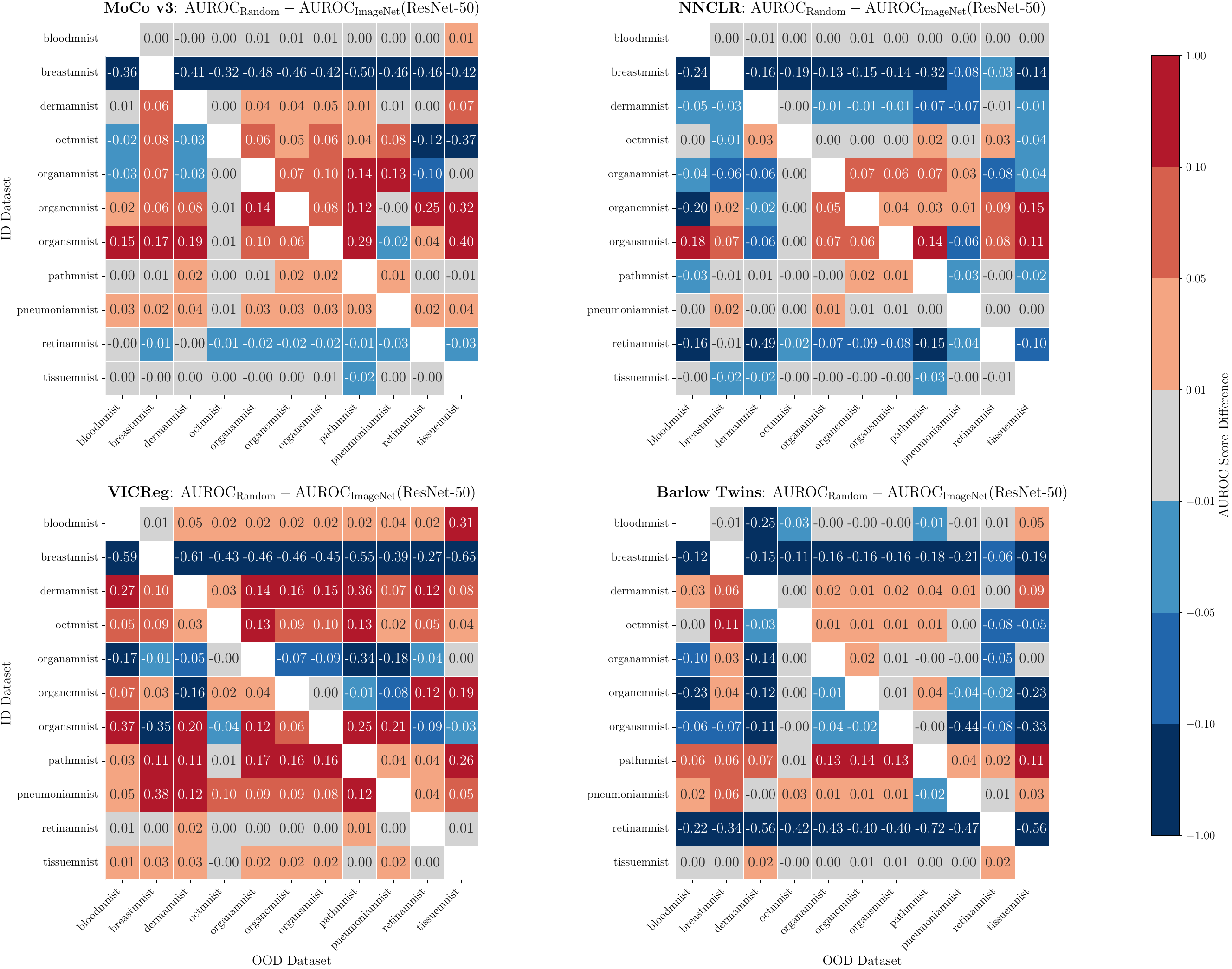}
    \caption{Effect of Initialization (Random vs. \imagenet) on AUROC scores for OOD detection across various dataset combinations and SSL methods with ResNet-50 backbone. Positive values (blue) indicate better OOD detection performance with the random initialization, while negative values (red) favor \imagenet initialization.}
    \label{fig:resnet_random_imagenet_p2}
\end{figure}

For the OOD detection task, we compare the importance of \imagenet-supervised weight initialization with that of random initialization. Figures \ref{fig:resnet_random_imagenet_p1} and \ref{fig:resnet_random_imagenet_p2} illustrate the effect of initialization across ResNet-50 based models and $(\mathcal{P}_{\text{ID}}, \mathcal{P}_{\text{OOD}})$ pairs. It is evident that the effect of initialization depends on both the choice of SSL model and the specific $(\mathcal{P}_{\text{ID}}, \mathcal{P}_{\text{OOD}})$ pair. For many datasets, methods such as SimCLR, ReSSL, MoCo v3, and VICReg tend to prefer random initialization over \imagenet initialization, whereas BYOL and DINO favor \imagenet initialization. Additionally, models trained on smaller datasets like BreastMNIST and RetinaMNIST specifically benefit from \imagenet initialization across all SSL methods for OOD detection. This observation indicates that, for the ResNet-50 backbone, smaller datasets gain significant advantages from supervised \imagenet weights.

\begin{figure}[h]
    \centering
    \includegraphics[width=\linewidth]{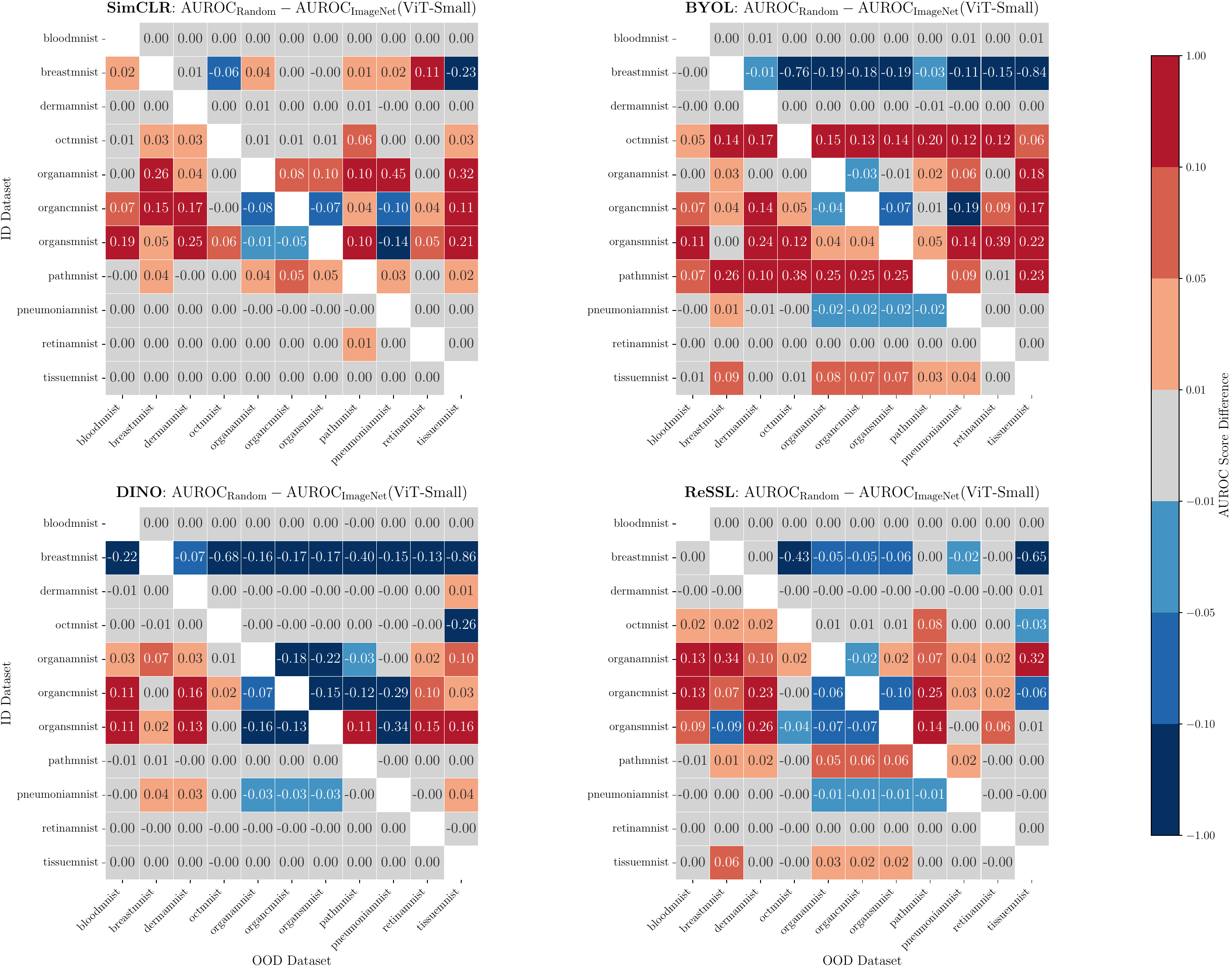}
    \caption{Effect of initialization (Random vs. \imagenet) on AUROC scores for OOD detection across various dataset combinations and SSL methods with ViT-Small backbone. Positive values (blue) indicate better OOD detection performance with the random initialization, while negative values (red) favor \imagenet initialization.}
    \label{fig:vit_random_imagenet_p1}
\end{figure}

\begin{figure}[h]
    \centering
    \includegraphics[width=\linewidth]{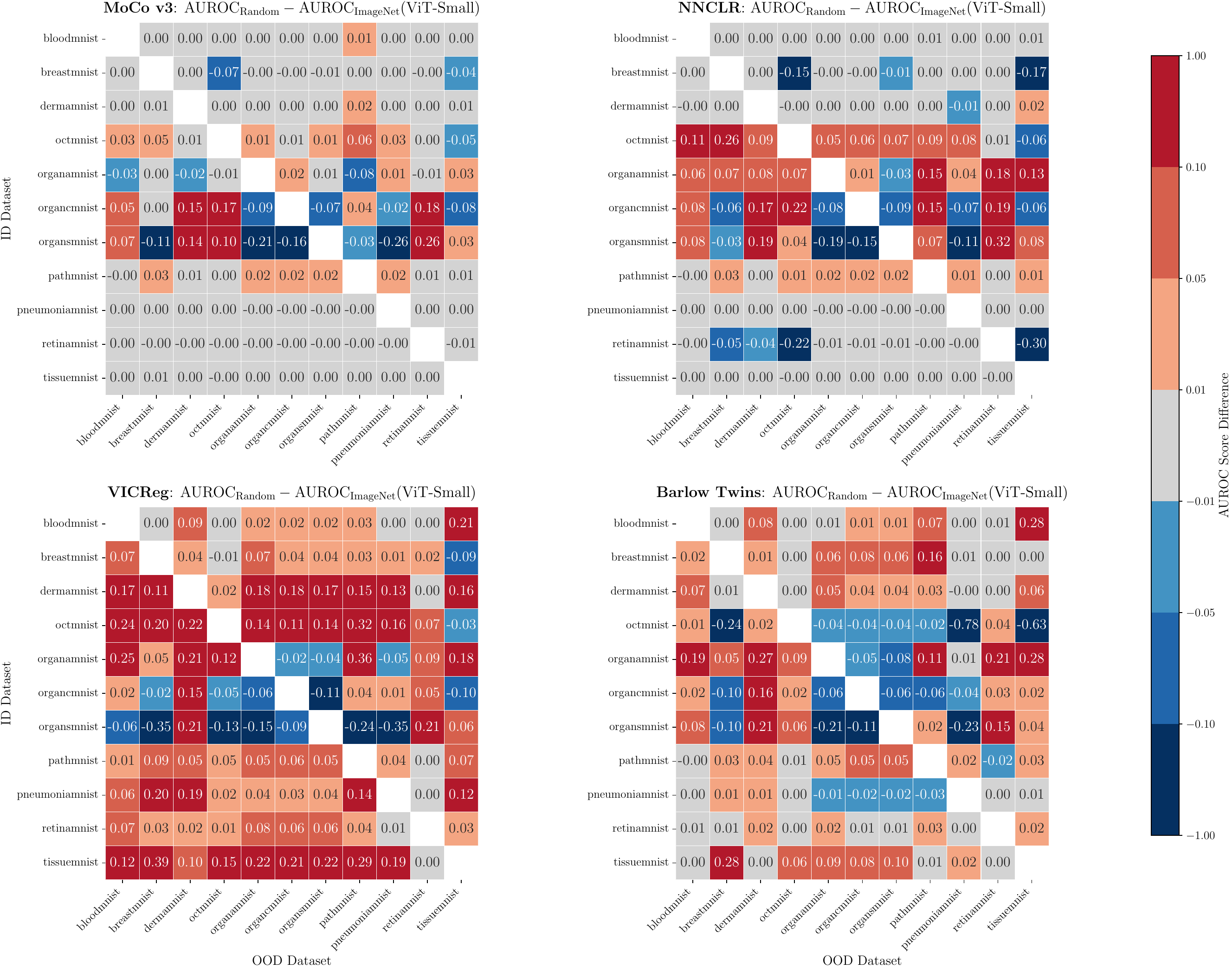}
    \caption{Effect of initialization (Random vs. \imagenet) on AUROC Scores for OOD detection across various dataset combinations and SSL methods with ViT-Small backbone. Positive values (blue) indicate better OOD detection performance with the random initialization, while negative values (red) favor \imagenet initialization.}
    \label{fig:vit_random_imagenet_p2}
\end{figure}

Next, we investigate whether this observation also holds for the ViT-Small backbone. Figures \ref{fig:vit_random_imagenet_p1} and \ref{fig:vit_random_imagenet_p2} illustrate the effect of initialization for ViT-Small architectures. Compared to ResNet-50, it can be observed that the effect of initialization is diminished, as the differences between AUROC scores are closer to zero for many $(\mathcal{P}_{\text{ID}}, \mathcal{P}_{\text{OOD}})$ pairs. Despite this, several models still favor random initialization over \imagenet initialization for when trained with datasets such as OCTMNIST, OrganAMNIST, OrganCMNIST, and OrganSMNIST. BYOL, DINO, and ReSSL methods trained with BreastMNIST prefers \imagenet initialization for better OOD detection. However, unlike the ResNet-50 backbone, this preference does not extend consistently across other SSL methods. These findings suggest that while ViT-Small may be less sensitive to initialization strategies overall, certain models and datasets still benefit from appropriate weight initialization.

\begin{table}[ht]
    \centering
    \caption{AUROC Differences (self-supervised \imagenet minus supervised \imagenet) for ResNet-50 models. Values are reported as $\Delta$AUC $\pm$ 95\% CI. Significant differences ($p < 0.05$) are highlighted in \colorbox{green!45}{green} if self-supervised weights performs better and \colorbox{red!45}{red} if supervised weights performs better. If the absolute difference is smaller than $0.05$, they are highlighted in \colorbox{green!15}{light green} and \colorbox{red!15}{light red} instead.}
    \label{tab:auc_diff_table_ood}
    \begin{tabular}{l|c|c|c|c}
\toprule
        Dataset        & MoCo v3                                        & DINO                                         & BYOL                                         & SimCLR                                       \\
        \midrule
        BloodMNIST     & \cellcolor{red!15}$-0.003 \pm 0.004$           & \cellcolor{green!15}$0.005 \pm 0.005$          & $0.008 \pm 0.016$                            & \cellcolor{red!30}$-0.057 \pm 0.048$           \\
        BreastMNIST    & \cellcolor{red!15}$-0.033 \pm 0.043$           & \cellcolor{red!45}$-0.398 \pm 0.166$           & $0.001 \pm 0.003$                            & \cellcolor{red!45}$-0.066 \pm 0.068$           \\
        Dermamnist     & \cellcolor{red!45}$-0.095 \pm 0.066$           & \cellcolor{green!15}$0.010 \pm 0.007$          & $-0.008 \pm 0.013$                           & \cellcolor{green!30}$0.091 \pm 0.113$           \\
        OctMNIST       & \cellcolor{green!15}$0.047 \pm 0.035$           & \cellcolor{red!15}$-0.048 \pm 0.059$           & \cellcolor{red!15}$-0.004 \pm 0.005$           & $0.099 \pm 0.184$                            \\
        OrganAMNIST    & $0.047 \pm 0.075$                              & $0.040 \pm 0.091$                              & \cellcolor{green!45}$-0.072 \pm 0.098$          & $0.073 \pm 0.124$                            \\
        OrganCMNIST    & $-0.026 \pm 0.096$                             & \cellcolor{green!45}$0.157 \pm 0.115$           & \cellcolor{red!45}$-0.147 \pm 0.112$           & $-0.078 \pm 0.149$                           \\
        OrganSMNIST    & $-0.062 \pm 0.174$                             & $0.131 \pm 0.158$                              & $-0.009 \pm 0.072$                           & $-0.104 \pm 0.203$                           \\
        PathMNIST      & \cellcolor{red!15}$-0.022 \pm 0.020$           & \cellcolor{red!45}$-0.270 \pm 0.080$           & \cellcolor{red!15}$-0.027 \pm 0.015$           & $-0.032 \pm 0.141$                           \\
        PneumoniaMNIST & \cellcolor{red!15}$-0.025 \pm 0.020$           & \cellcolor{green!15}$0.011 \pm 0.005$          & \cellcolor{green!15}$0.029 \pm 0.013$          & \cellcolor{green!45}$0.302 \pm 0.142$           \\
        RetinaMNIST    & \cellcolor{red!15}$-0.002 \pm 0.002$           & \cellcolor{red!45}$-0.069 \pm 0.054$           & \cellcolor{red!15}$-0.031 \pm 0.040$           & \cellcolor{red!15}$-0.036 \pm 0.164$           \\
        TissueMNIST    & \cellcolor{red!15}$-0.018 \pm 0.021$           & \cellcolor{green!45}$0.347 \pm 0.241$           & $-0.001 \pm 0.002$                           & $-0.007 \pm 0.014$                           \\
        \bottomrule
    \end{tabular}
\end{table}

Table~\ref{tab:auc_diff_table_ood} summarizes the AUROC differences—computed as the AUROC of models initialized with self-supervised \imagenet weights minus that of models initialized with supervised \imagenet weights—across various datasets for four SSL methods, using an $\alpha$ level of 0.05. As detailed in the table, the magnitude and direction of the differences vary considerably with the SSL method. For instance, DINO shows a strong preference for initializing with supervised \imagenet weights for BreastMNIST, while it favors self-supervised \imagenet initialization for TissueMNIST. Similarly, SimCLR indicates a marked preference for self-supervised \imagenet initialization in PneumoniaMNIST, but a preference for supervised \imagenet weights in BloodMNIST and BreastMNIST. In contrast, MoCo v3 and BYOL tend to yield more modest differences overall, suggesting less pronounced preferences. These results highlight that the choice of self-supervised initialization significantly influences OOD detection performance in a dataset-dependent manner.

\paragraph{Effect of Multi-domain Datasets}

\begin{figure}[h]
    \centering
    \includegraphics[width=0.8\linewidth]{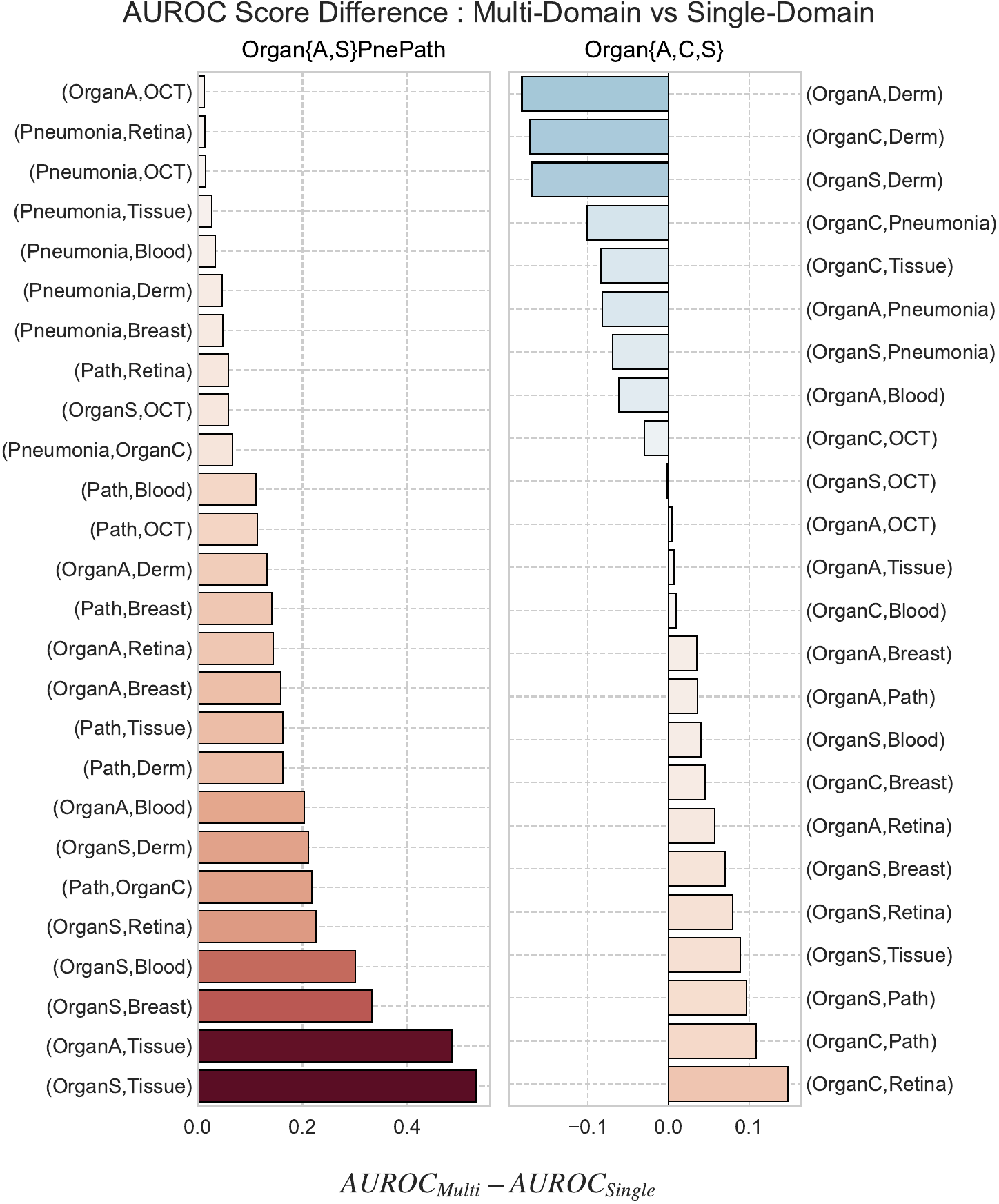}
    \caption{Differences in OOD AUROC scores between models trained on single-domain datasets and those trained on multi-domain datasets, averaged across different models. The y-axis lists the $(\mathcal{P}_{\text{ID}}, \mathcal{P}_{\text{OOD}})$ pairs, where $\mathcal{P}_{\text{ID}}$ represents the in-domain dataset used to train single-domain models and to calculate Gaussian class centers for both single-domain and multi-domain models, while $\mathcal{P}_{\text{OOD}}$ represents the out-of-distribution (OOD) dataset. Positive values on the x-axis indicate higher average AUROC scores for single-domain models, whereas negative values favor multi-domain models. Left: Comparison between single-domain models and models trained on the multi-domain dataset Organ\{A,C,S\}. Right: Comparison between single-domain models and models trained on Organ\{A,S\}PnePath.
    }
    \label{fig:multimodality_full_comparison}
\end{figure}

Figure~\ref{fig:multimodality_full_comparison}  compares the OOD AUROC scores of models trained on single-domain datasets versus multi-domain datasets. On average, multi-domain datasets containing more diverse samples, such as Organ\{A,S\}PnePath, outperform less diverse datasets like Organ\{A,C,S\} across all $(\mathcal{P}_{\text{ID}}, \mathcal{P}_{\text{OOD}})$ pairs.

The left panel of Figure~\ref{fig:multimodality_full_comparison} highlights that models trained on single-domain datasets generally achieve higher AUROC scores compared to those trained on the Organ\{A,C,S\} dataset. Conversely, the right panel shows that models trained on Organ\{A,S\}PnePath consistently achieve higher AUROC scores compared to single-domain trained models, leading to superior overall OOD detection performance.

These results emphasize the critical role of dataset diversity in improving OOD detection performance. Incorporating more diverse samples during training significantly enhances a model's ability to generalize and detect OOD examples effectively. This underscores the importance of dataset design and diversity in the development of robust OOD detection systems.   

\newpage
\phantom{.}  
\newpage
\phantom{still need a new page}
\newpage
\phantom{please}
\newpage
\newpage
\phantom{.}  
\newpage
\phantom{still need a new page}
\newpage
\phantom{please}
\newpage
\phantom{.}  
\newpage
 \phantom{please}
 \newpage

\subsubsection{Generalizability}
\begin{figure}[h]
\centering
    \hspace*{-0.5cm}
\includegraphics[width=1\textwidth]{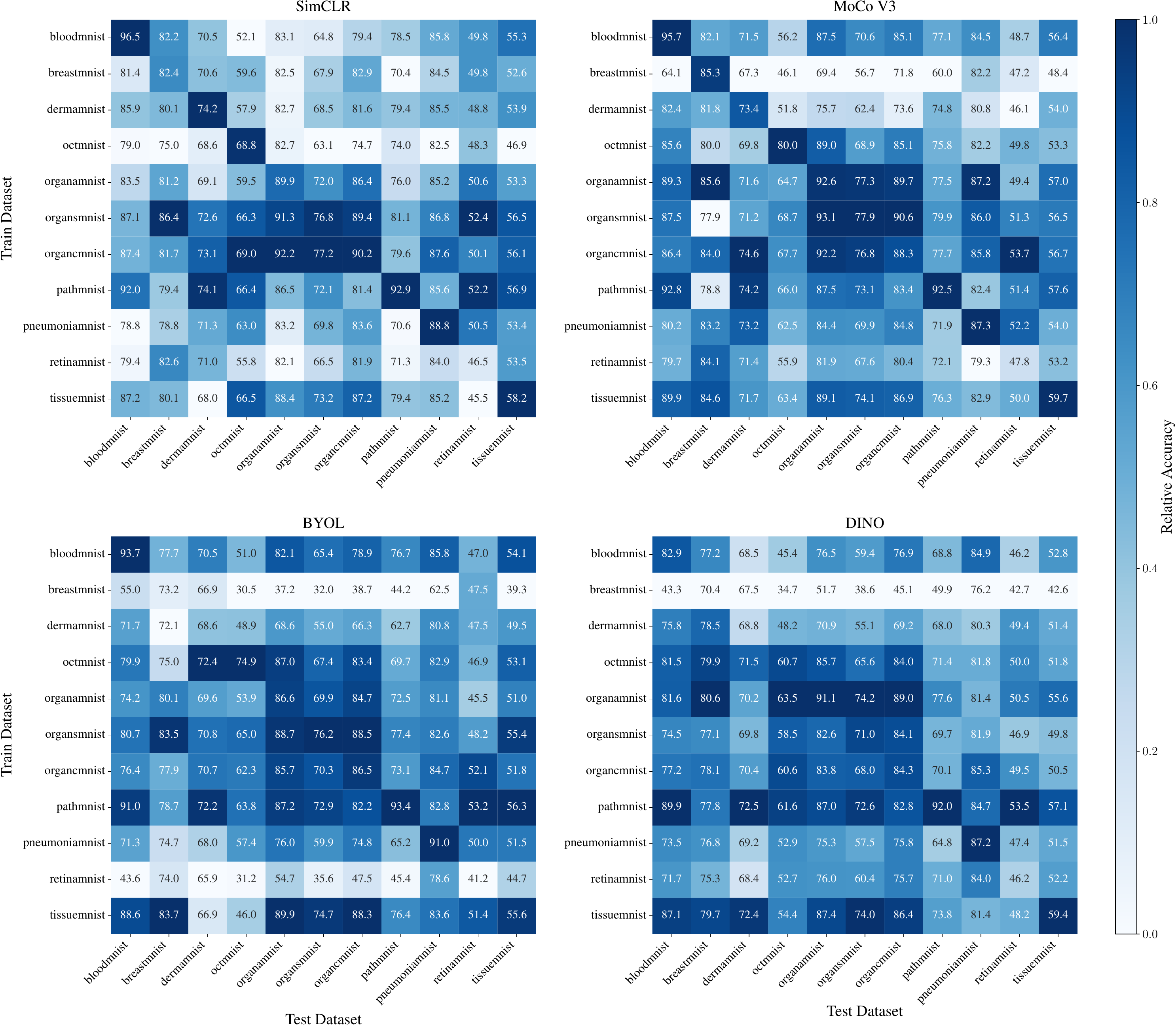} \\
\captionsetup{justification=centering}
\caption{
Cross-dataset transfer performance of four SSL methods (SimCLR, MoCo v3, BYOL, and DINO) using ResNet-50 with random initialization. Each heatmap shows the linear evaluation accuracy (\%) when the model is pre-trained on one dataset (y-axis) and evaluated on another (x-axis), with column-wise normalized color intensities where darker blue indicates higher relative performance.  
}
\label{fig:transfer_heatmap}
\end{figure}

\begin{figure}[h] 
\centering
    \hspace*{-0.5cm}
\includegraphics[width=1\textwidth]{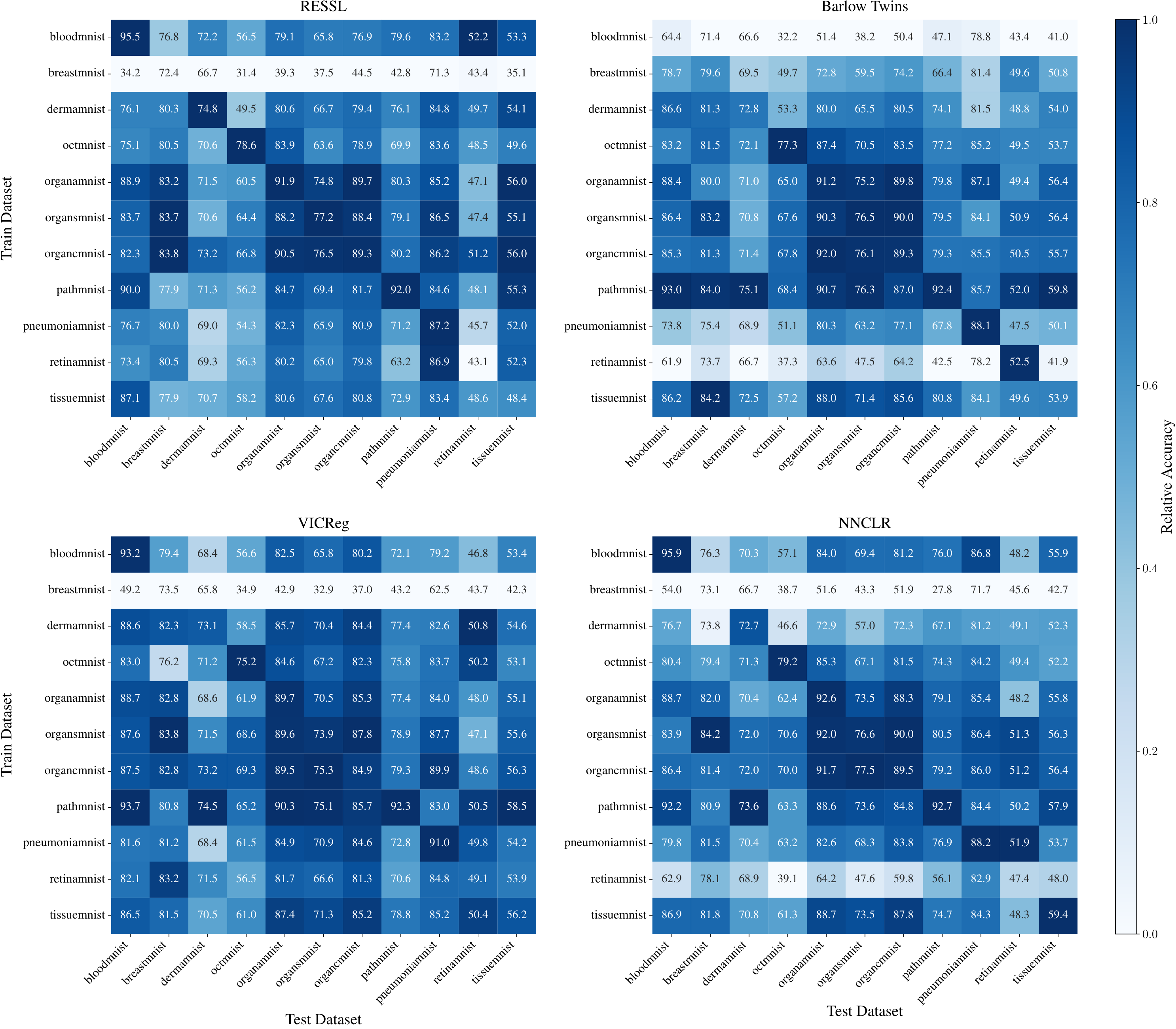} \\
\captionsetup{justification=centering}
\caption{
Cross-dataset transfer performance of four SSL methods (RESSL, Barlow Twins, VICReg, and NNCLR) using ResNet-50 with random initialization. Each heatmap shows the linear evaluation accuracy (\%) when the model is pre-trained on one dataset (y-axis) and evaluated on another (x-axis), with column-wise normalized color intensities where darker blue indicates higher relative performance.  
}
\label{fig:transfer_heatmap_2}
\end{figure}

In Figures ~\ref{fig:transfer_heatmap} and ~\ref{fig:transfer_heatmap_2}, we provide a detailed analysis of cross-dataset transfer performance across the MedMNIST collection for various self-supervised learning methods.

The heatmaps illustrate the performance matrix, where each row corresponds to a source (training) dataset and each column corresponds to a target (test) dataset. The diagonal elements represent in-domain performance, indicating the results when the source and target datasets are identical. Off-diagonal elements reveal the transferability of learned representations, showcasing how well models trained on one dataset generalize to others.

This comprehensive analysis offers insights into the strengths and limitations of different SSL methods when applied to diverse medical datasets, highlighting their adaptability in cross-domain scenarios.


\label{sec:appendix_ap_g}

\begin{figure}[h]
\centering
\begin{minipage}{1\textwidth}
\begin{flushleft}
    \setlength{\parindent}{10pt}
    \textbf{Mean Cross-Dataset Performance:} In Figure~\ref{fig:average_transfer_heatmap}, we report the mean accuracies across five SSL methods (SimCLR, MoCo v3, ReSSL, DINO, and BYOL) for 13 distinct medical imaging datasets. The performance matrix, visualized with column-wise normalized colors, includes both single-domain and multi-domain datasets (Organ\{A,C,S\} and Organ\{A,S\}PnePath). Matrix elements, where the train and test datasets are identical, represent the in-domain performance. Notably, the Organ\{A,C,S\} datasets exhibit stronger transfer performance among themselves, forming a distinct cluster that suggests these datasets share similar underlying features, which is due to their shared medical imaging modality with different viewing perspectives. \newline
    
    \indent Furthermore, the empirical results demonstrate the superiority of multi-domain training, with Organ\{A,C,S\} exhibiting superior performance in 6 out of 11 target datasets compared to its constituent domains (as shown in Figure~\ref{fig:average_transfer_heatmap}). This multi-domain approach achieves a mean accuracy of 76.11\% across all target datasets, significantly outperforming the individual constituent domains which achieve mean accuracies of 74.14\%, 75.06\%, and 75.07\% respectively. The analysis suggests a positive correlation between dataset size and source robustness, with the smallest datasets (BreastMNIST: 780 samples, RetinaMNIST: 1,600 samples) showing the poorest generalization performance. However, this correlation is not absolute, as evidenced by PathMNIST achieving the second-highest robustness despite some larger datasets performing worse. Interestingly, while Organ\{A,S\}PnePath does not consistently outperform its constituent datasets on most individual targets (as shown in Figure~\ref{fig:average_transfer_diff_heatmap}), it emerges as the most robust source dataset overall, 
    achieving the highest average test accuracy (76.33\%) across all target domains. 
    This suggests that while domain-specific pre-training can provide better performance on particular targets, combining diverse domains leads to better overall generalization.
\end{flushleft}
\end{minipage}

\vspace{0.4cm}
\hspace*{-0.8cm}
\includegraphics[width=0.7\textwidth]{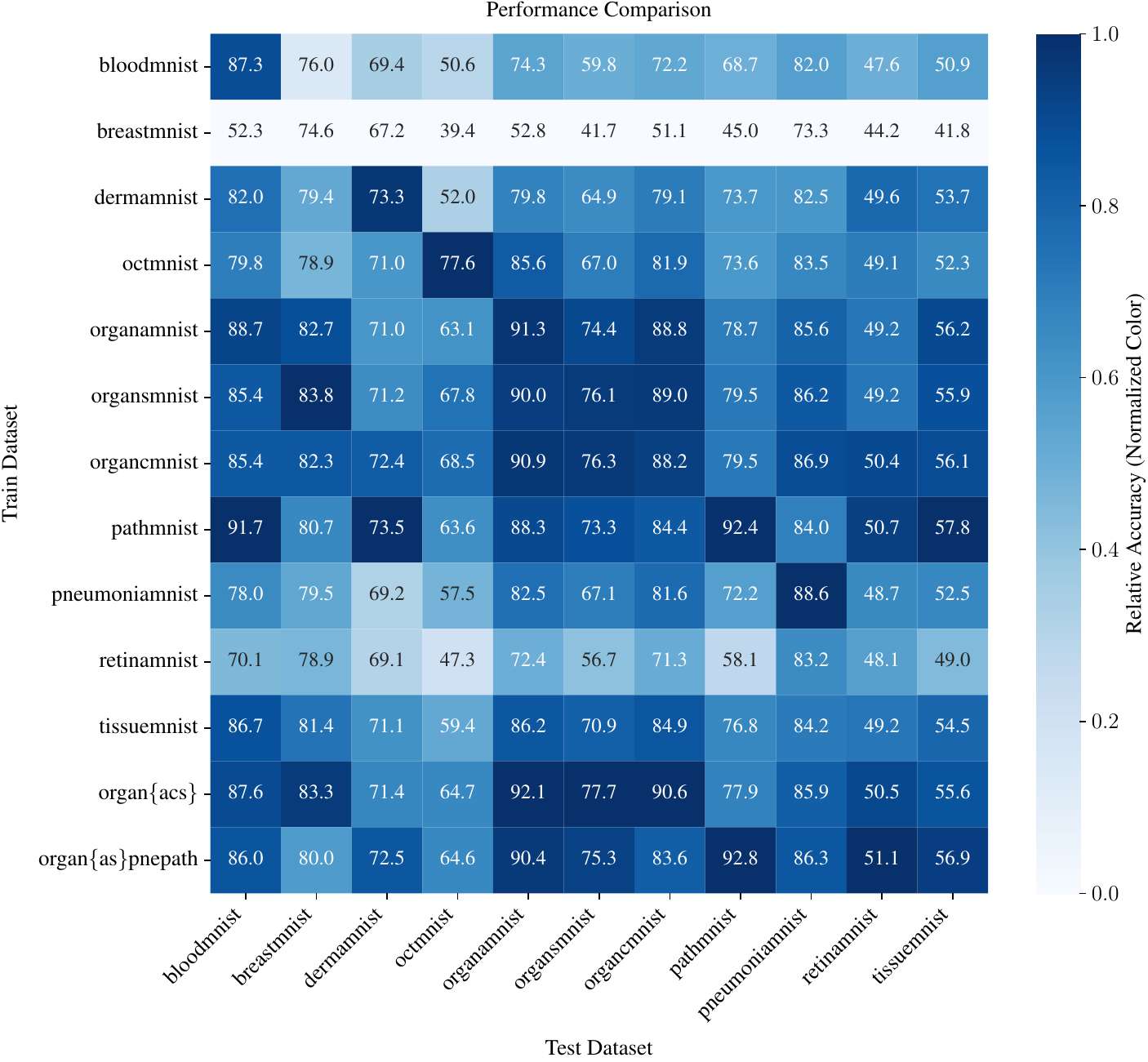} \\ 
\captionsetup{justification=centering}
\caption{
Performance matrix showing cross-dataset accuracy results averaged across five self-supervised methods (SimCLR, MoCo v3, ReSSL, DINO, BYOL). Values indicate mean accuracy percentages, with colors normalized per column to better visualize relative performance within each target domain.
}
\label{fig:average_transfer_heatmap}
\end{figure}

\begin{figure}[h]
\centering

\hspace*{-0.9cm}

\includegraphics[width=0.7\textwidth]{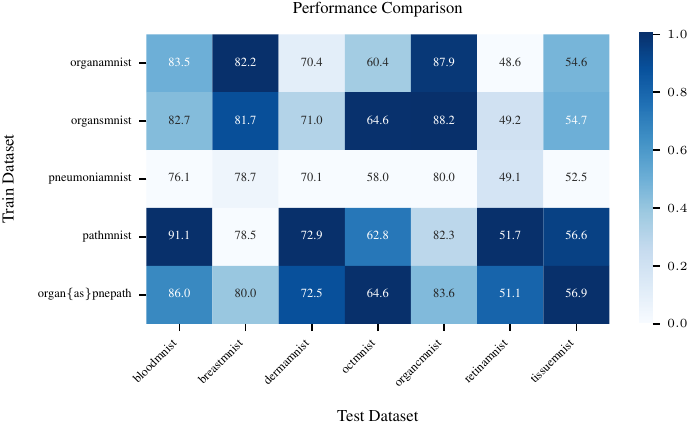} \\ 
\captionsetup{justification=centering}
\caption{
Performance matrix showing cross-domain accuracy results for Organ\{A,S\}PnePath and its constituent single-domain datasets across other target datasets, averaged over five self-supervised methods (SimCLR, MoCo v3, ReSSL, DINO, BYOL). Each row represents a source dataset used for training, while columns show the target datasets for evaluation. Values indicate mean accuracy percentages, with colors normalized per column to better visualize relative performance within each target domain. 
}
\label{fig:average_transfer_diff_heatmap}
\end{figure}

\begin{figure}[h]
\centering
\includegraphics[width=0.7\textwidth]{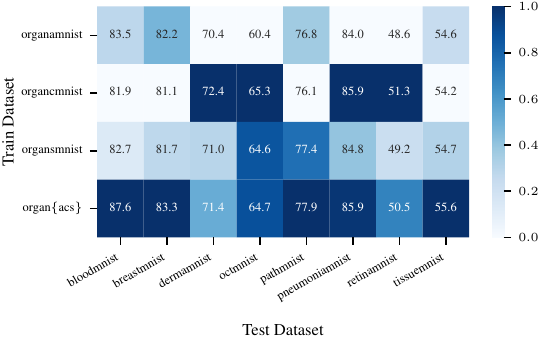} \\ 
\captionsetup{justification=centering}
\caption{
Performance matrix showing cross-domain accuracy results for Organ\{A,C,S\} and its constituent single-domain datasets across other target datasets, averaged over five self-supervised methods (SimCLR, MoCo v3, ReSSL, DINO, BYOL). Each row represents a source dataset used for training, while columns show the target datasets for evaluation. Values indicate mean accuracy percentages, with colors normalized per column to better visualize relative performance within each target domain.  
}
\label{fig:organs_heatmap}
\end{figure}

\begin{table}[h]
\begin{flushleft}
   \setlength{\parindent}{10pt}
   \textbf{Comparison of \imagenet vs. Random Initialization for In-Domain and Cross-Dataset Performance:} In Table~\ref{tab:init_comparison_generalizability}, we provide the results of paired $t$-tests to analyze the effect of \imagenet initialization on the in-domain performance and generalizability of the learned representations. To evaluate the impact of \imagenet initialization compared to random initialization, 
    we conducted paired $t$-tests for each test dataset across all cross-dataset training combinations including the in-domain setting where test and train splits come from the same dataset.
   The results demonstrate statistically significant improvements $(p < 0.05)$ across all datasets, with varying degrees of benefit. OCTMNIST showed the largest relative improvement ($22.08 \pm 31.16\%$), followed by PathMNIST ($16.95 \pm 32.10\%$) and OrganSMNIST ($13.55 \pm 29.62\%$). Even datasets with higher baseline accuracies benefited from \imagenet initialization, with BloodMNIST improving from 80.23 to 88.41 ($13.52 \pm 27.58\%$ relative gain). The smallest improvements were observed in BreastMNIST ($3.47 \pm 6.47\%$), DermaMNIST ($3.76 \pm 4.48\%$), PneumoniaMNIST ($3.93 \pm 7.28\%$) and RetinaMNIST ($4.00 \pm 7.56\%$), though these gains remained statistically significant. Notably, all improvements were achieved with strong statistical significance $(p < 0.05)$, providing robust evidence for the benefit of \imagenet initialization in self-supervised pretraining across diverse medical imaging domains.
\end{flushleft}
\vspace{0.4cm}
\centering
\begin{tabular}{c|c|c|c|c}
\toprule
Dataset & Random Init. & \imagenet Init. & Improvement & $p$-value \\
\midrule
BloodMNIST & $80.52$ & $88.41$ & $13.52 \pm 27.58\%$ & $0.0000$ \\
BreastMNIST & $79.68$ & $82.26$ & $3.47 \pm 6.47\%$ & $0.0000$ \\
DermaMNIST & $70.67$ & $73.26$ & $3.76 \pm 4.48\%$ & $0.0000$ \\
OctMNIST & $58.33$ & $68.24$ & $22.08 \pm 31.16\%$ & $0.0000$ \\
OrganAMNIST & $81.54$ & $88.94$ & $12.79 \pm 26.96\%$ & $0.0000$ \\
OrganCMNIST & $79.57$ & $85.15$ & $10.86 \pm 27.45\%$ & $0.0002$ \\
OrganCMNIST & $66.31$ & $72.30$ & $13.55 \pm 29.62\%$ & $0.0000$ \\
PathMNIST & $72.90$ & $81.86$ & $16.95 \pm 32.10\%$ & $0.0000$ \\
PneumoniaMNIST & $83.50$ & $86.52$ & $3.93 \pm 7.28\%$ & $0.0000$ \\
RetinaMNIST & $48.82$ & $50.63$ & $4.00 \pm 7.56\%$ & $0.0000$ \\
TissueMNIST & $52.99$ & $57.46$ & $9.50 \pm 13.30\%$ & $0.0000$ \\
\bottomrule
\end{tabular}
\caption{Comparison of random initialization versus \imagenet initialization across different medical datasets. Results show mean accuracy (\%) for both initialization methods, relative improvement, and statistical significance.}
\label{tab:init_comparison_generalizability}
\end{table}

\newpage
\phantom{.}
\newpage
\phantom{.}
\newpage
\phantom{.}
\newpage
\phantom{.}
\newpage

\subsubsection{KNN Evaluation}

Here, we evaluate the effectiveness of a simpler, computationally efficient classifier on top of self-supervised features, compared to a standard linear classifier. We train a KNN classifier on features extracted from the training dataset using selected backbones (ResNet-50 and ViT-Small) with both \imagenet and random initialization. To ensure optimal KNN performance, we conduct a grid search over several hyperparameters, including the number of neighbors, temperature scaling, and distance functions (Euclidean and cosine). This approach allows us to systematically assess how well KNN classifiers leverage the representations across various datasets and initialization strategies.
        
\begin{figure*}[t]
    \centering
    \begin{minipage}{1\textwidth}
        \centering
        \includegraphics[width=\textwidth]{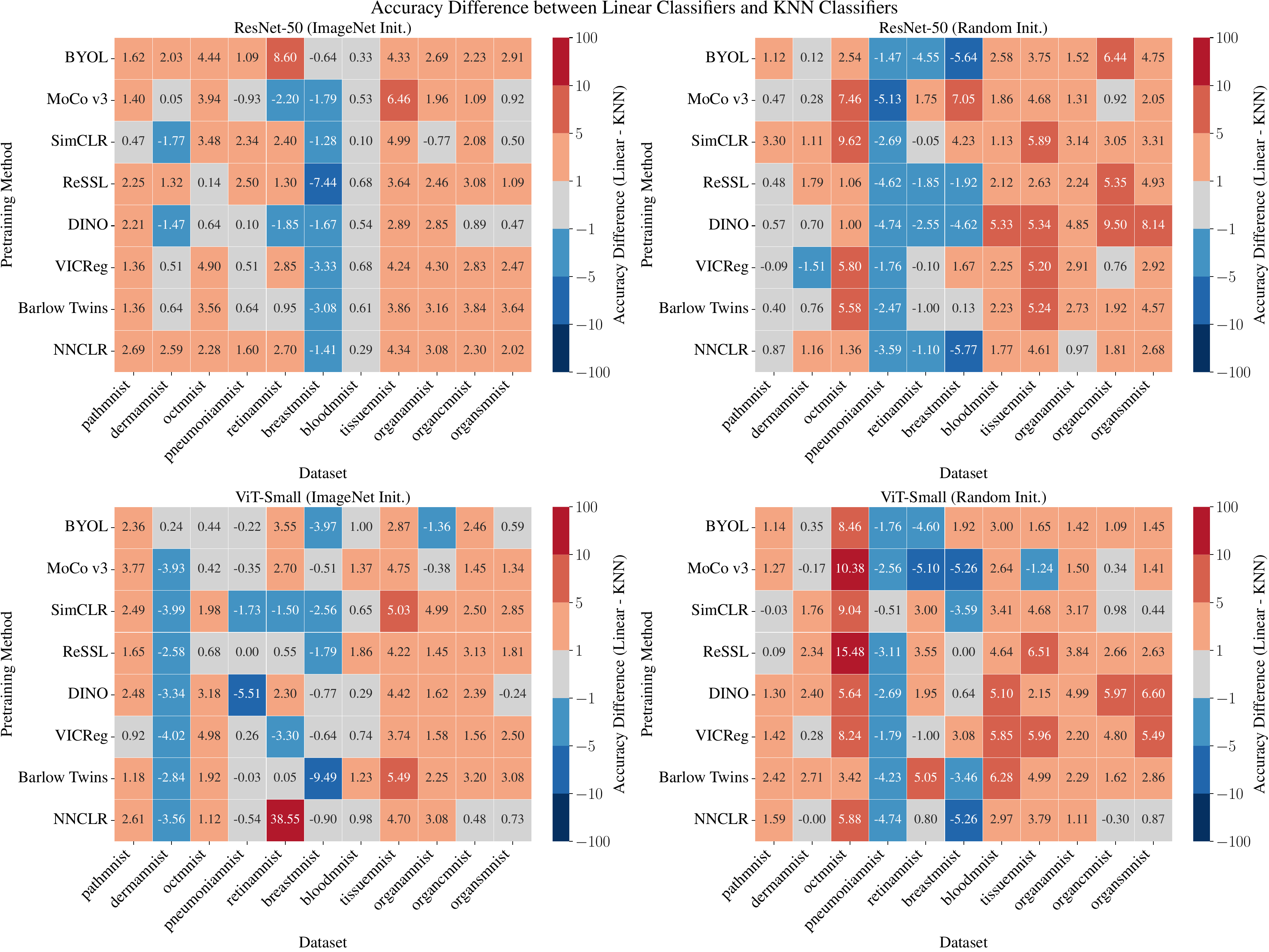}
        \captionsetup{justification=centering}
        \caption{
            Heatmaps showing the accuracy differences between KNN and linear classifiers for various methods on different datasets, evaluated for both ResNet-50 and ViT-Small backbones as well as \imagenet and random initializations. Positive values (shades of red) indicate higher accuracy with a linear classifier while negative values (shades of blue) indicate higher accuracy with a KNN classifier. Gray values show where the accuracy differences are less significant, less than 1.
        }
        \label{fig:pretrained_vs_random_init_vs_knn_heatmaps}
    \end{minipage}
\end{figure*}

Figure \ref{fig:pretrained_vs_random_init_vs_knn_heatmaps} presents a detailed comparison of the performance differences between KNN and linear classifiers across various self-supervised learning methods on multiple datasets. Positive values (shaded in red) indicate higher accuracy with the linear classifier, while negative values (blue) highlight cases where KNN performs better. Gray values represent minimal accuracy differences, suggesting similar performances between the classifiers in those instances. This analysis reveals that the choice of classifier can significantly impact performance, with notable variation across datasets and initialization strategies (\imagenet vs. random) on ResNet-50 and ViT-Small backbones.
        
Interestingly, there is a greater disparity in performance with random initialization compared to \imagenet initialization. Specifically, there are 36 absolute differences exceeding 5 (indicated by a darker shade) with random initialization, whereas only 7 such differences occur with \imagenet initialization. Another notable trend is that the initialization strategy and backbone choice primarily drive these performance differences, with consistent patterns across methods for the same initialization and backbone combination. This is evident from the vertical strips of similar colors or values, indicating that the trends are largely method-independent. However, exceptions like BreastMNIST and RetinaMNIST display different colors in their respective strips, likely due to their smaller dataset sizes, where greater fluctuations are expected.
        
In conclusion, \imagenet-initialized models generally favor linear classifiers, while random-initialized models show more instability in performance. These findings emphasize the importance of carefully selecting both the initialization strategy and classifier type based on dataset characteristics, as these choices can meaningfully impact model performance. In particular, there are cases where using a KNN classifier, which is computationally less demanding, can yield better results than training a linear classifier.

\newpage
\subsection{Conclusion}

We present a comprehensive empirical study evaluating the robustness, generalizability, and cross-domain efficacy of self-supervised learning methods in medical imaging. Our analysis encompasses 11 distinct medical datasets from MedMNIST, examining 8 popular SSL approaches across both convolutional (ResNet-50) and transformer-based (ViT-Small) architectures. To ensure rigorous comparison, we conduct hyperparameter optimization through grid search for both pre-training and linear evaluation phases, and report the mean accuracy with confidence interval over five different runs.
Our experimental framework investigates five distinct pre-training paradigms: (1) supervised learning with random initialization, (2) supervised learning with supervised \imagenet initialization, (3) self-supervised learning with random initialization, (4) self-supervised learning with supervised \imagenet initialization, and (5) self-supervised learning with self-supervised \imagenet initialization. We evaluate these approaches across multiple dimensions, including in-domain classification performance, out-of-distribution (OOD) detection capabilities, transfer learning efficacy, and cross-dataset generalization.

\paragraph*{Novelty and Significance}
Our work presents the most comprehensive analysis of self-supervised learning methods in medical imaging that systematically evaluates multiple critical aspects often overlooked, including OOD detection, cross-dataset generalizability, multi-domain pretraining, transfer learning and performance under label scarcity. Our finding that MoCo v3 consistently outperforms other methods—achieving superior performance in 5 out of 11 datasets while maintaining minimal accuracy degradation in cross-domain tasks and showing superior performance in OOD detection- provides a clear direction for method selection in medical applications. Equally significant is our discovery that ViT architectures, despite lower in-domain performance, demonstrate superior cross-dataset generalization and OOD detection capabilities—a critical consideration for real-world clinical deployment. Additionally, our observation that multi-modality training improves OOD detection while single-modality multi-domain training enhances accuracy offers practical guidance for designing robust medical imaging systems. These insights advance the understanding of how SSL techniques transfer across the unique challenges of medical domains and provide evidence-based recommendations for practitioners developing generalizable medical imaging AI.

\paragraph*{}
\subsection{Key Findings and Empirical Observations}
\label{sec:key_findings}

Our comprehensive analysis of self-supervised learning (SSL) methods in medical imaging revealed several important findings across different dimensions:

\subsubsection{In-Domain Performance}
\begin{itemize}
    \item \textbf{Method Performance:} MoCo v3 emerged as the most effective SSL method, achieving superior performance in 5 out of 11 datasets while maintaining minimal accuracy degradation (in 4 out of 11 datasets) for cross-domain generalization tasks.
    
    \item \textbf{Initialization Impact:} \imagenet initialization generally improved in-domain performance across methods compared to random initialization. However, no clear superior approach emerged when comparing supervised \imagenet and self-supervised \imagenet weights for initializing self-supervised pretraining on medical images.
    \item \textbf{Architecture Comparison:} ResNet architecture outperformed Vision Transformers (ViT) in the majority of in-domain evaluations. When trained from scratch, ResNet achieved higher accuracy in 8 out of 11 datasets. With pretrained weights, this advantage extended to 9 out of 11 datasets.
    
    \item \textbf{Initialization Effect on Architecture Gap:} The performance gap between ResNet and ViT architectures narrowed when moving from random to \imagenet initialization, though the relative ranking remained largely unchanged (except for the Retina dataset).
    
    \item \textbf{Method-Specific Benefits:} BYOL and DINO demonstrated remarkable improvement with \imagenet initialization, transitioning from among the worst performers to among the best performers.
        
    \item \textbf{Label Scarcity Effects:} Under limited label availability, ViT exhibited larger accuracy drops compared to ResNet, a counterintuitive finding given its lower initial accuracy (where smaller drops would typically be expected due to diminishing returns).
\end{itemize}

\subsubsection{Out-of-Distribution (OOD) Detection}
\begin{itemize}
    \item \textbf{Method Effectiveness:} MoCo demonstrated superior OOD detection capabilities when trained from scratch.
    
    \item \textbf{Architecture Advantage:} ViT consistently outperformed ResNet in OOD detection tasks across different methods and datasets.
    
    \item \textbf{Initialization Impact:} The effect of \imagenet initialization on OOD detection performance was inconsistent, with significant variation across methods and datasets.
    
    \item \textbf{Architectural Trends:} The ViT superiority trend generally persisted with \imagenet initialization, though with method-specific variations in relative performance.
\end{itemize}

\subsubsection{Generalizability}
\begin{itemize}
    \item \textbf{Initialization Effect:} \imagenet initialization improved cross-dataset generalizability, producing both better in-domain performance and more transferable representations.
    
   \item \textbf{Architecture Comparison:} While ResNet-50 achieved higher in-domain performance than ViT-Small in most datasets (8/11 when trained from scratch and 9/11 when starting from ImageNet weights), ViT-Small showed different behavior in cross-dataset generalization. When pre-trained with \imagenet weights and evaluated on generalization tasks, ViT-Small reversed the in-domain trend by outperforming ResNet-50 in 8 out of 11 datasets, as shown in Figure~\ref{fig:generalizability_arch.pdf}. This suggests that ViT-Small benefits more from transfer learning in cross-dataset scenarios, despite its weaker in-domain performance.

    \item \textbf{Convergence with Pretraining:} With pretrained weights, performance differences between architectures consistently approached zero, suggesting that pretraining reduces architecture-specific advantages.
\end{itemize}

\subsubsection{Multi-Domain Effects}
\begin{itemize}
    \item \textbf{Single-Modality Impact:} Single-modality multi-domain training (Organ\{ACS\}) generally improved in-domain accuracy but reduced OOD detection performance.
    
    \item \textbf{Multi-Modality Impact:} Multi-modality training (Organ\{AS\}PnePath) typically decreased in-domain accuracy but consistently improved OOD detection capabilities.
\end{itemize}

These findings provide valuable insights into the complex interplay between self-supervised learning methods, architectural choices, initialization strategies, and dataset characteristics in medical imaging applications.


\end{document}